\documentclass[a4paper,fleqn]{./style/cas-dc}

\usepackage[numbers, sort&compress]{natbib}

\usepackage{cleveref}
\usepackage{subcaption}
\usepackage{bm}
\usepackage{multirow}
\usepackage{rotating}
\usepackage{arydshln}
\usepackage{algorithm}
\usepackage{algpseudocode}
\usepackage{mathtools}
\usepackage{siunitx}

\graphicspath{{images-src/}} 

\DeclareMathOperator*{\argmax}{arg\,max}
\crefname{figure}{Fig.}{Figs.}
\crefname{equation}{Eq.}{Eqs.}

\begin{document}
\let\WriteBookmarks\relax
\def\floatpagepagefraction{1}
\def\textpagefraction{.001}
\shorttitle{Context-adaptive policy framework}
\shortauthors{T. R. Winter et~al.}
\let\printorcid\relax

\title [mode = title]{A context-adaptive policy framework for robust and reactive robotic manipulation via uncertainty-aware imitation learning}

\author[1]{Tim R. Winter}
\cormark[1]
\credit{Conceptualization,
	Data curation,
	Formal analysis,
	Investigation,
	Methodology,
	Resources,
	Software,
	Validation,
	Visualization,
	Writing - original draft}

\cortext[cor1]{Corresponding author; E-mail address: tim.winter@dlr.de.}

\author[1]{Leonard Kl\"upfel}
\credit{Data curation, Software}

\author[1]{Ashok M. Sundaram}
\credit{Project administration, Resources, Writing - review and editing}

\author[1]{Werner Friedl}
\credit{Resources}

\author[1]{ Maximo A. Roa}
\credit{Funding acquisition, Project administration, Resources}

\author[1]{Freek Stulp}
\credit{Funding acquisition, Resources, Writing - review and editing}

\author[1]{Jo{\~{a}}o Silv{\'{e}}rio}
\credit{Conceptualization, Methodology, Resources, Supervision, Writing - review and editing}

\affiliation[1]{organization={German Aerospace Center (DLR), Robotics and Mechatronics Center (RMC)},
	addressline={M\"unchener Str. 20},
	city={We\ss ling},
	postcode={82234},
	country={Germany},
	email={,\\  E-mail: name.surname@dlr.de}}

\begin{abstract}
	Generating robust and reactive manipulation strategies that can adapt to changing context information is a challenging task in robotics.
	Over the years, Learning from Demonstration (LfD) has emerged as an intuitive and effective solution for generating reactive policies, particularly by following dynamical-system(DS)-based approaches.
	However, most state-of-the-art DS-based approaches focus on addressing the robustness limitations, overlooking the modulation of policies in response to the environment.
	As a result, they tend to be inflexible with respect to parameterization by task-dependent variables.
	In this work, we build on existing work on policy fusion and uncertainty quantification to propose a context-adaptive policy framework that combines task-parameterized, robust and reactive manipulation.
	For this, we use LfD to acquire a policy that is conditioned on the robot state and low-dimensional task-dependent parameters reflecting the environment.
	We combine the learned policy with additional uncertainty-aware policies using a Mixture of Experts (MoE) formulation to improve its out-of-distribution (OOD) robustness and convergence behavior.
	The approach is evaluated on the LASA handwriting dataset and on a real 7-DoF robot in three scenarios: force-conditioned grasping, manipulation of deformable food items and object-centric grasping.
\end{abstract}



\maketitle

\section{Introduction}\label{sec:introduction}
Context-adaptive manipulation is crucial in tasks in which the robot needs to adapt its motion to its state and task-dependent parameters.
This becomes particularly relevant in the field of deformable object manipulation, as the manipulation strategy is often influenced by the objects' changeable shape,
or when contact forces need to be incorporated.
Due to the complexity inherent in modeling this context\footnote{In this work, we refer to context as the combination of robot state and task-dependent parameters that reflect the environment.}
information, the application of conventional planning and control algorithms is restricted.
Consequently, recent approaches focus on solving this problem using data-driven machine learning techniques.
Especially, the application of LfD approaches in the domain of robotic manipulation has gained popularity in recent years,
due to the typically reduced data requirements and the adaptability to new tasks~\cite{Ravichandar2020}.
However, classical LfD approaches have focused either on time-variant or DS-based policies, both exhibiting limited reactivity properties to changing environments.
An in-depth review of related work can be found in \Cref{sec:related_work}.

\begin{figure}
	\centering
	\begin{subfigure}[b]{1.0\columnwidth}
		\centering
		\includegraphics[trim={4.5cm 1cm 4.5cm 0},clip,width=0.32\columnwidth]{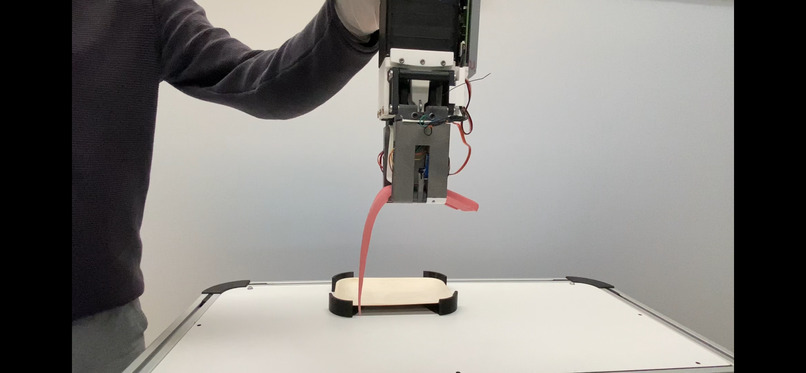}
		\includegraphics[trim={4.5cm 1cm 4.5cm 0},clip,width=0.32\columnwidth]{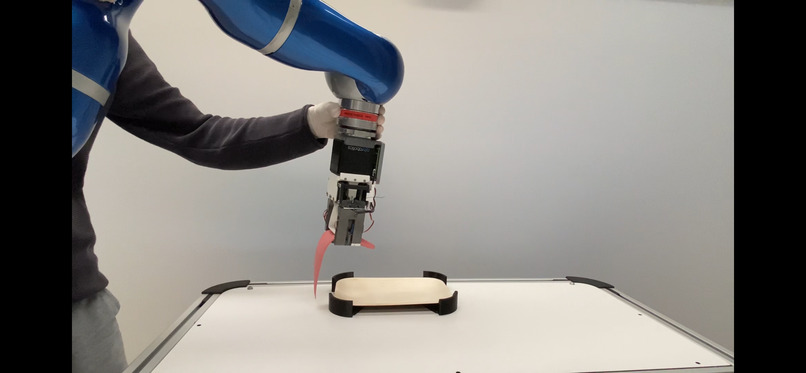}
		\includegraphics[trim={4.5cm 1cm 4.5cm 0},clip,width=0.32\columnwidth]{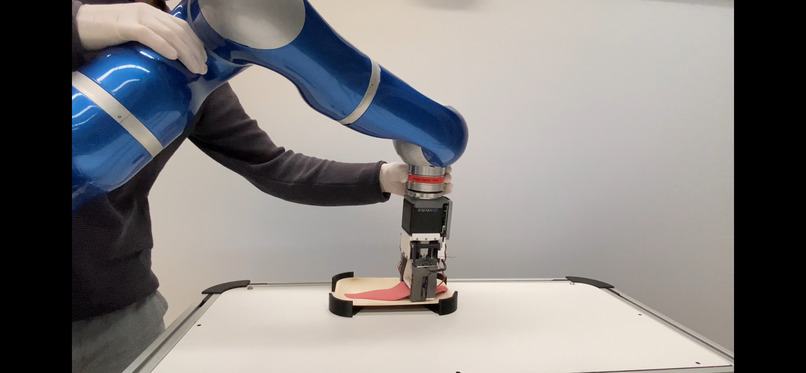}
	\end{subfigure}
	\\[0.1cm]
	\begin{subfigure}[b]{1.0\columnwidth}
		\centering
		\includegraphics[trim={4.5cm 1cm 4.5cm 0},clip,width=0.32\columnwidth]{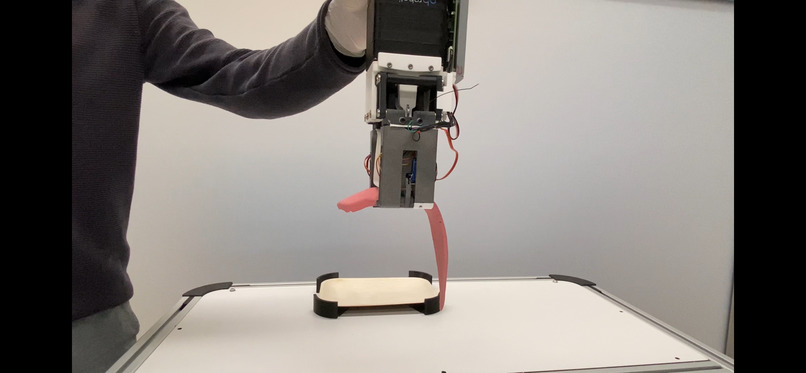}
		\includegraphics[trim={4.5cm 1cm 4.5cm 0},clip,width=0.32\columnwidth]{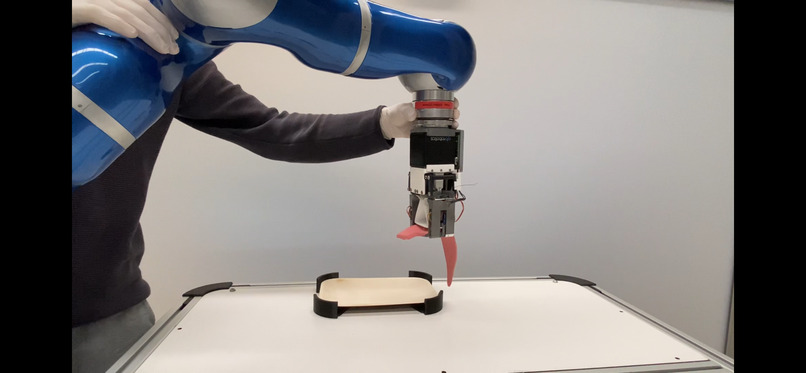}
		\includegraphics[trim={4.5cm 1cm 4.5cm 0},clip,width=0.32\columnwidth]{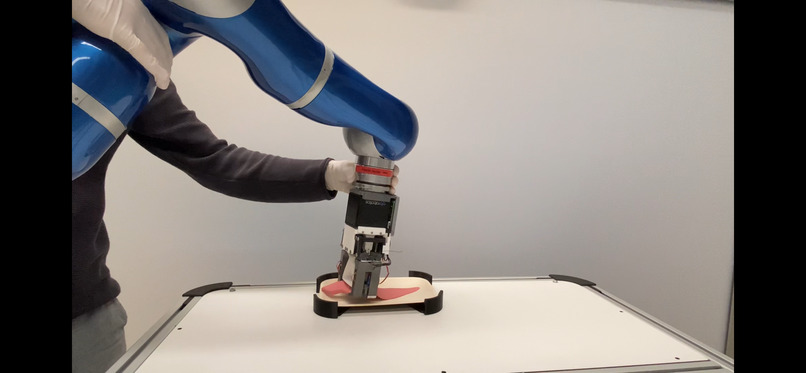}
	\end{subfigure}
	\caption{\label{fig:kinesthetic_teaching}In the above images we see how placing a deformable piece of silicone fish on a tray requires different approach and place strategies depending on how it has been grasped.
		Instead of accurately modeling the different possibilities, we propose to learn them from human demonstrations.}
\end{figure}

\begingroup
\renewcommand{\thefootnote}{}
\footnotetext[0]{%
	\fontsize{9}{11}\selectfont
	\begin{tabular}{@{\hspace{0.0em}}l}
		Digital Object Identifier (DOI): \href{https://doi.org/10.1016/j.robot.2026.105649}{10.1016/j.robot.2026.105649} \\
		Code available at: \href{https://github.com/DLR-RM/CAPOF}{github.com/DLR-RM/CAPOF}                               \\
		Video available at: \href{https://www.youtube.com/watch?v=vxQcqhXqHA8}{youtube.com/watch?v=vxQcqhXqHA8}
	\end{tabular}
}
\endgroup

In this work, we take inspiration from prior achievements in policy fusion and uncertainty quantification to design a robust DS-based policy framework that can be modulated in response to the environment.
For this, we propose an MoE formulation with uncertainty-aware experts and mixing coefficients, where individual expert policies are active in distinct regions of the state space.
We employ Gaussian Process Regression (GPR)~\cite{Rasmussen2006} (\Cref{sec:background}) due its data efficiency to learn an \textit{LfD policy} conditioned on the robot state and task-dependent parameters.
Moreover, we make use of the analytical uncertainty formulation provided by GPR to derive a \textit{stabilizing policy} that guides the robot towards regions with low epistemic uncertainty, preventing undesired behavior,
and design a kernel-based \textit{goal attractor policy} that empirically enhances the convergence behavior of the framework.
\Cref{sec:policy_fusion} introduces our approach.
Concretely, the main contributions of this work can be summarized as:
\begin{itemize}
	\item Incorporating the task-parameters explicitly into the learned policy (\ref{sec:lfd_policy})
	      --- \cite{KhansariZadeh2011,Pignat2019,Franzese2021,Meszaros2022} consider only the robot state and are thus not able to adapt to the environment.
	\item Formulating the problem as an MoE framework with varying, uncertainty-aware mixing coefficients (\ref{sec:coefficients_design})
	      --- works like~\cite{Franzese2021,Meszaros2022} assume constant coefficients.
	\item Extending the stabilizing policy of~\cite{Franzese2021,Meszaros2022} to include orientation as part of the state (\ref{sec:stab_policy}),
	      and introducing a novel kernel-based goal attractor policy (\ref{sec:goal_policy}),
	      demonstrating experimentally that both increase robustness.
\end{itemize}
Using the LASA handwriting dataset~\cite{LASA} and three different real-robot manipulation tasks, in \Cref{sec:evaluation} we show that our approach is able to robustly reproduce the demonstrated trajectories
while adapting to changing context information, maintaining robustness when confronted with disturbances, different initial conditions or distribution shifts and reaching the demonstrated goal states.
Finally, we discuss our approach in \Cref{sec:discussion} and draw a conclusion in \Cref{sec:conclusion}.

\section{Related Work}\label{sec:related_work}
In literature, there is a variety of possibilities to categorize LfD approaches in robotic manipulation.
In this work, we focus on approaches that learn a probabilistic policy $p_{\bm{\theta}}(\bm{\xi}\vert \bm{s})$ which maps states $\bm{s}$ to actions $\bm{\xi}$ based on a demonstration dataset,
where $\bm{\theta}$ are the policy parameters.
The learned policy can then be executed to reproduce the demonstrated behavior at test input $\bm{s}_*$, i.e.,~$\bm{\xi}_* = f_{\bm{\theta}}(\bm{s}_*)$.

Early works concentrate mostly on time-variant policies (e.g.,~$p_{\bm{\theta}}(\bm{\xi}\vert t)$).
This includes classical regression approaches applied to the domain of robotic motion generation,
such as Gaussian Mixture Regression (GMR)~\cite{Calinon2007,Calinon2015}, Gaussian Process Regression (GPR)~\cite{NguyenTuong2008}, and Locally Weighted Regression (LWR)~\cite{Atkeson1991}.
In addition, the concept of movement primitives (MP) was exploited to encode complex motor behaviors through the superposition of parameterized basis functions.
Notable examples include Dynamic MP (DMP)~\cite{Ijspeert2013}, Probabilistic MP (ProMP)~\cite{Paraschos2013}, and Kernelized MP (KMP)~\cite{Huang2019}.
In order to increase generalization and adaptability, these approaches have been extended into task-parameterized or context-aware variants,
which modulate the generated policies in response to environmental or task-dependent parameters.
Among these, Task-Parameterized Gaussian Mixture Models (TP-GMM)~\cite{Calinon2015,Silverio2015}, Task-Parameterized DMP (TP-DMP)~\cite{Stulp2013,Pervez2017} and
Task-Parameterized ProMP (TP-ProMP)~\cite{Kulak2021,Ewerton2016} have gained particular prominence, with several proposed enhancements targeting more flexible adaptation.
However, these approaches impose strong assumptions on the task-parameters and, due to their time-dependent nature, exhibit limited spatial robustness and restricted capacity to adapt to dynamically changing environments.

DS-based policies, which learn a function from robot state $\bm{x}$ to its derivative $\dot{\bm{x}}$, i.e.,~$p_{\bm{\theta}}(\dot{\bm{x}}\vert \bm{x})$, are a popular approach to address the aforementioned limitations,
as they offer the capacity for real-time trajectory adaptation due to the continuous incorporation of the robot state.
However, these methods often struggle to maintain robust performance under disturbances, unseen initial conditions or distribution shifts,
which can lead to significant trajectory deviations or unpredictable behavior.
Rather than explicitly addressing these robustness limitations, recent approaches such as diffusion policies~\cite{Chi2024} or flow-based policies~\cite{Zhang2025}
aim to achieve robust behavior by increasing the diversity and coverage of the training distribution.
Consequently, these approaches typically require large datasets to achieve robust behavior, and inference is often computationally expensive.
Nevertheless, reliable performance is generally limited to states that remain sufficiently close to the training distribution.

In contrast, several works have focused on learning stable dynamical systems with formal guarantees, including Lyapunov-constrained Gaussian mixture models~\cite{KhansariZadeh2011}
and stochastically stable Gaussian process state-space models~\cite{Umlauft2020}.
More recent approaches combine expressive neural models with stability guarantees,
including energy-based formulations~\cite{Jin2023a},
contrastive learning~\cite{PerezDattari2023},
stability-constrained deep dynamics~\cite{Manek2019},
and Neural ODEs with Lyapunov and barrier constraints~\cite{Nawaz2024}.
While these methods effectively learn stable dynamics from data, they typically restrict the permissible input and output variables to the robot state and its derivatives,
which limits their adaptability to changing environments.
Therefore, recent advances have integrated stability guarantees into task-parameterized frameworks,
such as Elastic-DS~\cite{Li2023} and SE(3) LPV-DS~\cite{Sun2024}, improving adaptability.
However, these methods --- in line with the broader task-parameterized literature~\cite{Calinon2015} ---
build on geometric or Cartesian-frame-based task parameterization, such as object poses and spatial transformations, rather than generic task-parameters,
e.g.,~object shape descriptors, measured contact forces or task phase variables.
A comprehensive overview of DS-based imitation learning, including stability-aware methods, is provided by~\cite{Hu2024}.

Another line of DS-based imitation learning works tackle the generalization limitations without offering formal control-theoretic guarantees;
instead they achieve robustness based on policy fusion~\cite{Pignat2019,Franzese2021,Meszaros2022}.
Although these approaches do not impose restrictions on state and action variables and would therefore, in principle, allow modulation by arbitrary parameters,
they routinely overlook the incorporation of variables different from the robot state and its derivatives, limiting their applicability in changing environments.

Thus, to the best of the authors' knowledge, a unified approach combining real-time adaptation, OOD robustness, and modulation by arbitrary external parameters remains largely underexplored.
In this work, we propose to address this gap using a Mixture of Experts formulation~\cite{Jacobs1991}
with a GPR-based policy backbone that provides uncertainty estimates.
We leverage these estimates for two purposes: to design a policy that enhances OOD robustness, and to determine expert activation weights.

\section{Background}\label{sec:background}
In this section, we present the methodological background for the development of our approach (\Cref{sec:policy_fusion}).

\subsection{Gaussian Process Regression}\label{sec:gpr}
Gaussian Process Regression (GPR)~\cite{Rasmussen2006} is a probabilistic, non-parametric approach to regression, wherein the latent function is assumed to follow a Gaussian process prior.
Given a set of $N$ training inputs and the corresponding noisy outputs $\{\bm{s}_{n}, \bm{\xi}_{n}\}^N_{n=1}$,
GPR leverages a joint Gaussian prior over the latent function values.
Here, $\bm{s}_{n} \in \mathbb{R}^{\mathcal{I}}$ specifies the input and $\bm{\xi}_{n} \in \mathbb{R}^{\mathcal{O}}$ denotes the corresponding output
and $\mathcal{I}$ and $\mathcal{O}$ are the dimensions of input and output space.
For a multidimensional output GP with independent output variables and shared hyperparameters across the dimensions,
a posterior predictive distribution for a single test input $\bm{s}_*$ can be derived by conditioning the joint prior on the observed data,
providing both, the mean prediction $\bm{\mu}_* \in \mathbb{R}^{\mathcal{O}}$ and the associated predictive latent-function variance $v_* \in [0,1]$:
\begin{gather}
	\bm{\mu}_* = \bm{k}_*^\top(\bm{K} + \sigma^2 \bm{I}_{N})^{-1}\bm{\Xi}, \label{eq:mu_pred} \\
	v_* = k_{**} - \bm{k}_*^\top(\bm{K} + \sigma^2 \bm{I}_{N})^{-1}\bm{k}_*. \label{eq:var_pred}
\end{gather}
Here, $k_{**} = k(\bm{s}_*, \bm{s}_*)$ denotes the kernel matrix computed at test input $\bm{s}_*$,
$\bm{k}_* = \begin{bmatrix} k(\bm{s}_*, \bm{s}_{1}) & \cdots & k(\bm{s}_*, \bm{s}_{N}) \end{bmatrix}^\top$ is a kernel matrix combining test input and training inputs and
\begin{equation}
	\bm{K} = \begin{bmatrix}
		k(\bm{s}_{1}, \bm{s}_{1}) & \cdots & k(\bm{s}_{1}, \bm{s}_{N}) \\
		\vdots                    & \ddots & \vdots                    \\
		k(\bm{s}_{N}, \bm{s}_{1}) & \cdots & k(\bm{s}_{N}, \bm{s}_{N})
	\end{bmatrix}
\end{equation}
refers to the kernel matrix evaluating the similarity of the demonstrated trajectory points.
Moreover, $\bm{I}_{N}$ is an $N \times N$ identity matrix and $\bm{\Xi}$ is defined as $\bm{\Xi} = \begin{bmatrix} \bm{\xi}_{1} & \cdots & \bm{\xi}_{N} \end{bmatrix}^\top$.

The kernel function $k(\bm{s}_{i}, \bm{s}_{j})$ is chosen according to the characteristics of the data.
In this work, a unit-amplitude radial basis function (RBF) kernel with $k_{**}=1$, given by
\begin{equation}
	k(\bm{s}_{i}, \bm{s}_{j}) = \text{exp}\left(-\frac{1}{2}(\bm{s}_{i} - \bm{s}_{j})^\top \bm{L} (\bm{s}_{i} - \bm{s}_{j})\right), \label{eq:rbf}
\end{equation}
is used as we assume smooth, continuous trajectories.
The length scales $\bm{L} = \text{diag}(\bm{l})^{-2}$, with $\bm{l} = \begin{bmatrix} l_1 \cdots l_{\mathcal{I}} \end{bmatrix} \ \in \ \mathbb{R}^{\mathcal{I}}$,
and the noise variance $\sigma^2$ denote the hyperparameters of GPR.

The predictive variance of GPR (\Cref{eq:var_pred}) provides a closed-form epistemic uncertainty\footnote{Epistemic uncertainty arises from limited knowledge of the model and can be reduced with more data, whereas aleatoric uncertainty reflects inherent stochasticity in the data; see~\cite{Huellermeier2021} for an
	introduction.}
quantification, that increases with the distance to the training data, and therefore helps to identify out of distribution states.

\begin{figure*}
	\centering
	\includegraphics[width=1.0\textwidth]{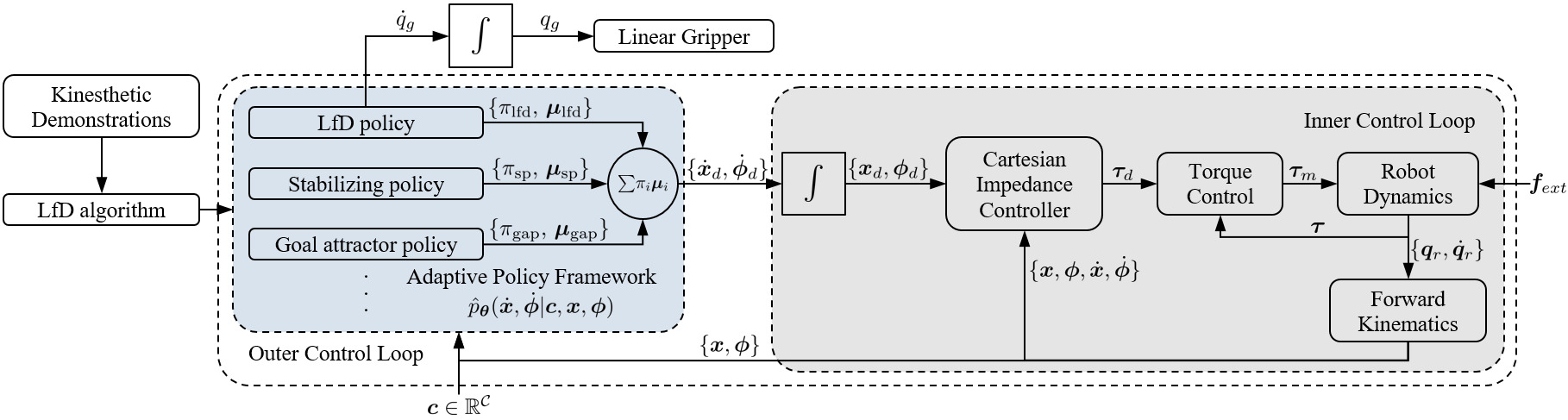}
	\caption{\label{fig:control_loop}Schematic overview of the two-layered control loop structure with the context-adaptive policy framework (highlighted in blue) that generates desired end-effector velocities in the outer control loop
		and the inner control loop (highlighted in gray) that makes the robot follow the desired velocities.}
\end{figure*}

\subsection{Probabilistic Mixture of Experts}\label{sec:moe}
The combination of expert models is a common machine learning technique for modeling high-dimensional data that adheres to multiple low-dimensional constraints.
It facilitates learning or formulating several simpler probability distributions that cover each a low-dimensional constraint and combining their outputs.
The two most common methods are the Mixture of Experts (MoE)~\cite{Jacobs1991} and the Product of Experts (PoE)~\cite{Hinton1999}.
With PoE, all conditions must be approximately satisfied to obtain the desired output, and it thus corresponds to an ``and'' operation.
MoE can be interpreted as an ``or'' operation that returns a weighted arithmetic mean of the individual expert distributions.

Unlike prior work that uses PoE to jointly satisfy multiple simultaneously active objectives within a shared task execution space~\cite{Pignat2021}
or employs gating mechanisms to select a single conditional expert model for representing multimodal movement primitives~\cite{Yildirim2024},
our approach focuses on smoothly composing locally specialized policies that dominate in distinct regions of the task or parameter space.
Concretely, we model the probabilistic control actions as a Gaussian distribution, i.e.,~$\hat{\bm{\xi}}_* \sim \mathcal{N}\left(\hat{\bm{\mu}}, \hat{\bm{\Sigma}}\right)$,
whose mean and covariance are obtained via moment-matching from an MoE.
The MoE is defined as $\hat{p}_{\bm{\theta}}(\bm{\xi}\vert \bm{s}) = \sum_{i=1}^{P} \pi_{i} p_{\bm{\theta}, i}(\bm{\xi}\vert \bm{s})$,
where $P$ is the number of experts, $p_{\bm{\theta}, i}(\bm{\xi}\vert \bm{s}) = \mathcal{N}\left(\bm{\mu}_{i},\bm{\Sigma}_{i}\right)$ represents the probability density function of expert $i$ and $\pi_{i}$ is its mixing coefficient, with $\sum_{i=1}^{P} \pi_{i} = 1$.
The resulting mixture moments are given by
\begin{equation}
	\hat{\bm{\mu}} = \sum_{i=1}^{P} \!\pi_{i} \bm{\mu}_{i}, \label{eq:MoE}
	\quad \hat{\bm{\Sigma}} = \sum_{i=1}^{P}\!\pi_{i}\! \left(\bm{\Sigma}_i + \bm{\mu}_{i}\bm{\mu}_{i}^\top\right) - \hat{\bm{\mu}}\hat{\bm{\mu}}^\top.
\end{equation}

\section{Context-adaptive policy framework}\label{sec:policy_fusion}
The aim of the context-adaptive policy framework is to provide a general framework for robust and reactive robotic manipulation within dynamically changing environments.
In order to achieve the reactive behavior of the robotic system, we apply a two-layer control loop structure, displayed in \Cref{fig:control_loop}, as the basis of our approach.
Here, the inner control loop (highlighted in gray), consisting of a low-level controller and the robot's dynamics, makes the robot follow desired linear and angular end-effector velocities $\dot{\bm{x}}_d$, $\dot{\bm{\phi}}_d$.\footnote{In this manuscript, we assume task-space control, which enables embodiment-independent motion specification and improves interpretability of the supplementary policies. However, a joint-space implementation would also be possible in principle.}
These velocities are obtained in the outer control loop\footnote{Note that the control rate of the outer control loop can be lower than the rate of the inner control loop, which is 1~kHz in this work.}
by querying a context-adaptive policy $\hat{p}_{\bm{\theta}}(\dot{\bm{x}}, \dot{\bm{\phi}}\vert \bm{c}, \bm{x}, \bm{\phi})$ at each time step.
This DS-based policy is conditioned on the end-effector pose, defined by position $\bm{x}$ and orientation $\bm{\phi}$,
and unlike state-of-the-art stable DS-based learning frameworks, also on generic low-dimensional task-dependent parameters $\bm{c} \in \mathbb{R}^{\mathcal{C}}$,
capturing task-relevant environmental features such as contact forces (\Cref{sec:regrasp}), object shape descriptors (\Cref{sec:fish}) or phase variables (\Cref{sec:dynamic_grasping}).
The outer control loop is exited if the robot has remained close to the goal pose for a predefined number of steps, a maximum number of iterations is reached, or an external stop is triggered.

Since, in the case of a purely data-driven policy $\hat{p}_{\bm{\theta}}$, the described control structure would lead to unpredictable robot behavior when $\bm{x}$, $\bm{\phi}$ or $\bm{c}$ deviate from the trained distribution,
we propose to formulate the context-adaptive policy as an MoE that combines an LfD policy with supplementary policies that keep the robot in low epistemic uncertainty regions and empirically enhance the convergence behavior.
While DS-based approaches that enforce constraints (e.g.,~stability~\cite{KhansariZadeh2011}) on the LfD policy typically restrict inputs and outputs to the robot state and its derivatives,
the combination of multiple policies enables the incorporation of additional, non-robot-related inputs into the LfD policy.
The proposed MoE formulation thus facilitates the modulation by arbitrary external low-dimensional task-dependent parameters while enabling robust and reactive manipulation.
In addition, the modular structure of our MoE formulation allows policies to be added and removed ad-hoc, contributing to better explainability and interpretability.

As backbone of the context-adaptive policy framework (blue part in \Cref{fig:control_loop}), we propose three distinct policies,\footnote{Additional policies can, in principle, be included.}
which we treat as Gaussian-distributed experts with means $\bm{\mu}_{i}$ and mixing coefficients $\pi_{i}$ (\Cref{eq:MoE}):
\begin{itemize}
	\item \textbf{LfD policy (}$\pi_{\mathrm{lfd}}$, $\bm{\mu}_{\mathrm{lfd}}$\textbf{):} A data-driven policy that is modulated in response to the current state of the robot and task-dependent parameters, which imitates the demonstrated motions.
	\item \textbf{Stabilizing policy (}$\pi_{\mathrm{sp}}$, $\bm{\mu}_{\mathrm{sp}}$\textbf{):} An uncertainty-aware policy that guides the robot towards regions with low epistemic uncertainty, overcoming the covariate shift and thus maintaining reliable performance.
	\item \textbf{Goal attractor policy (}$\pi_{\mathrm{gap}}$, $\bm{\mu}_{\mathrm{gap}}$\textbf{):} A kernel-based policy that empirically improves the convergence behavior of the framework.
\end{itemize}
We define the control action $\hat{\bm{\xi}}_* = \left[\dot{\bm{x}}^\top_d \> \dot{\bm{\phi}}^\top_d\right]^\top$ at test input $\bm{s}_*$ using the first moment of the MoE (\Cref{eq:MoE}), resulting in:
\begin{equation}
	\hat{\bm{\xi}}_* = \hat{\bm{\mu}} = \pi_{\mathrm{lfd}}  \bm{\mu}_{\mathrm{lfd}} + \pi_{\mathrm{sp}}  \bm{\mu}_{\mathrm{sp}} + \pi_{\mathrm{gap}}  \bm{\mu}_{\mathrm{gap}}. \label{eq:MoE_all_policies}
\end{equation}
In the following sections we describe the formulation of the means and mixing coefficients of the individual experts.

\subsection{LfD policy}\label{sec:lfd_policy}
In this work, we use GPR (see \Cref{sec:background}) to generate the LfD policy due to the small amount of data required and the explicit uncertainty formulation provided.
This makes it an ideal base policy for demonstrating the benefits of our uncertainty-aware framework.
Since GPs are computationally limited in the number of training data they can handle efficiently,
we subsample $N$ points from a set of $H$ demonstrated trajectories $\{\{\bm{s}_{m,h}, \bm{\xi}_{m,h}\}^{M_h}_{m=1}\}^H_{h=1}$, with $M_h$ samples.
The subsampled training inputs $\bm{s}_n=\left[\bm{c}^\top_n \> \bm{x}^\top_n \> \bm{\phi}^\top_n\right]^\top$ consist of the end-effector pose, along with additional low-dimensional task-dependent parameters (see \Cref{sec:policy_fusion}),
while the corresponding actions $\bm{\xi}_n=\left[\dot{\bm{x}}^\top_n \> \dot{\bm{\phi}}^\top_n\right]^\top$ contain the linear and angular end-effector velocities, as typical in DS-based learning frameworks (e.g.,~\cite{KhansariZadeh2011}).
Since the posterior predictive distribution exhibited by GPR already fulfills the requirement to be a Gaussian-distributed expert,
we define $\bm{\mu}_{\mathrm{lfd}} = \left[\bm{\mu}_{\dot{\bm{x}}}^\top \> \bm{\mu}_{\dot{\bm{\phi}}}^\top \right]^\top$ as the mean end-effector velocity computed by \Cref{eq:mu_pred}.\footnote{In the experiments, we extend the output $\bm{\xi}$ of the LfD policy by the joint angle velocity of the gripper $\dot{q}_g$ in order to infer the grasp behavior as well.
The mean computed according to \Cref{eq:mu_pred} therefore contains three distinct terms $\bm{\mu}_* = \left[\bm{\mu}_{\dot{\bm{x}}}^\top \> \bm{\mu}_{\dot{\bm{\phi}}}^\top \> \mu_{\dot{q}_g}^\top \right]^\top$.
Within the MoE formulation, we still consider only the mean end-effector velocity, while the mean joint angle velocity $\mu_{\dot{q}_g}$ is used separately to control the gripper as can be seen in \Cref{fig:control_loop}.}

\begin{figure*}
	\centering
	\begin{subfigure}[t]{0.2455\textwidth}
		\centering
		\includegraphics[trim={0 0 0 1cm},clip, width=\linewidth]{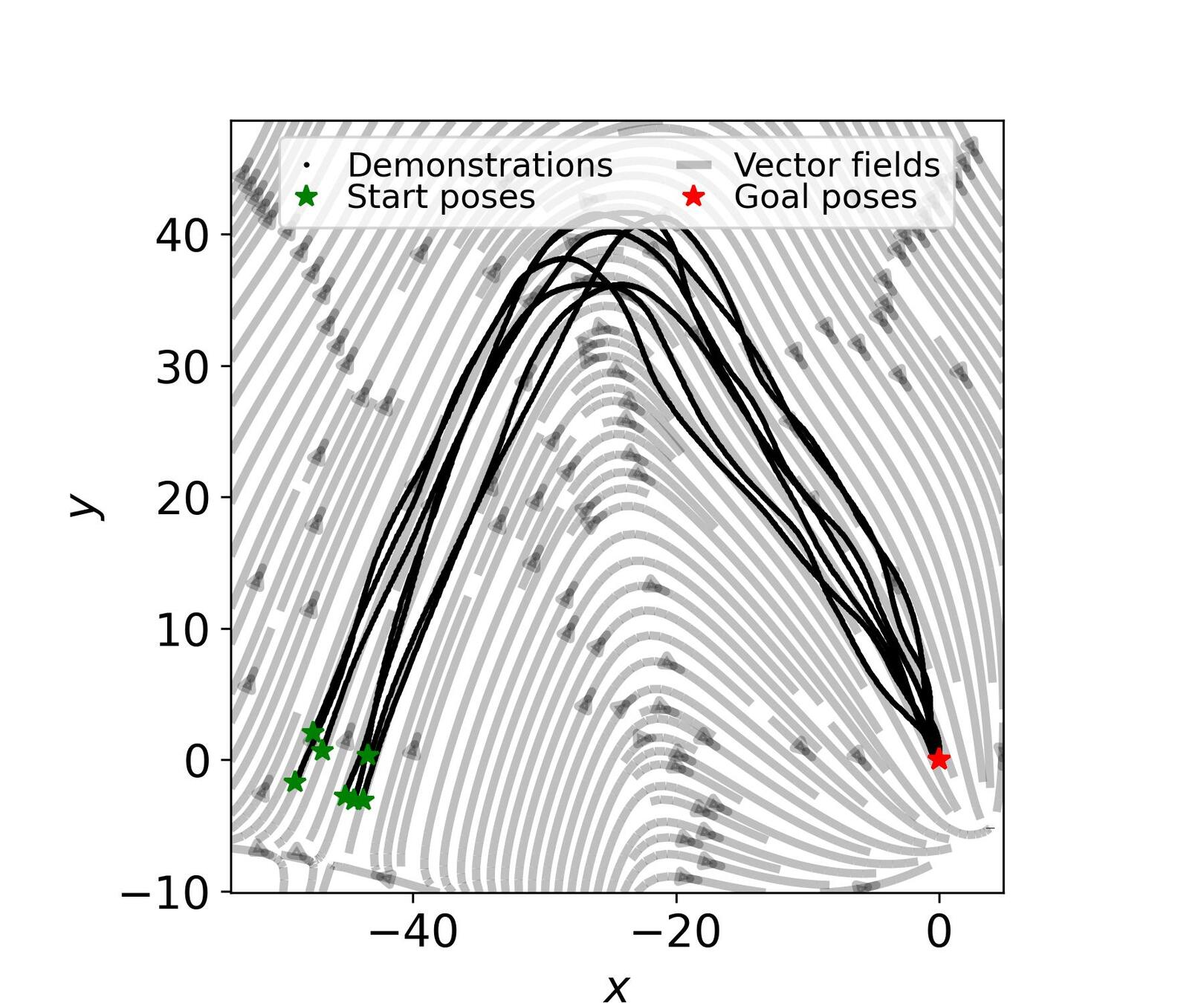}
		\caption{LfD policy}
		\label{fig:MoE_lfd}
	\end{subfigure}
	\hfill
	\begin{subfigure}[t]{0.2455\textwidth}
		\centering
		\includegraphics[trim={0 0 0 1cm},clip, width=\linewidth]{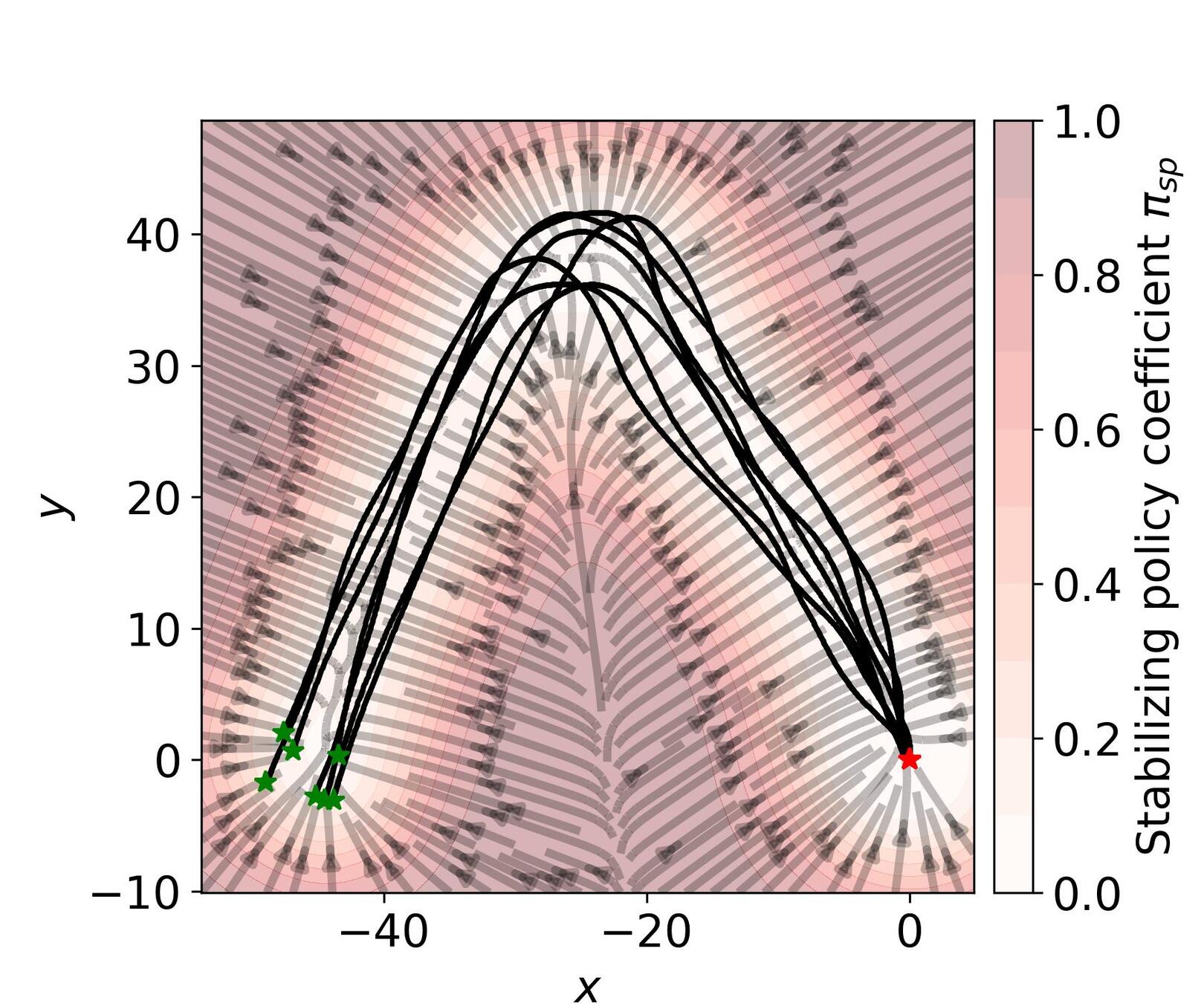}
		\caption{Stabilizing policy}
		\label{fig:MoE_stab}
	\end{subfigure}
	\hfill
	\begin{subfigure}[t]{0.2455\textwidth}
		\centering
		\includegraphics[trim={0 0 0 1cm},clip, width=\linewidth]{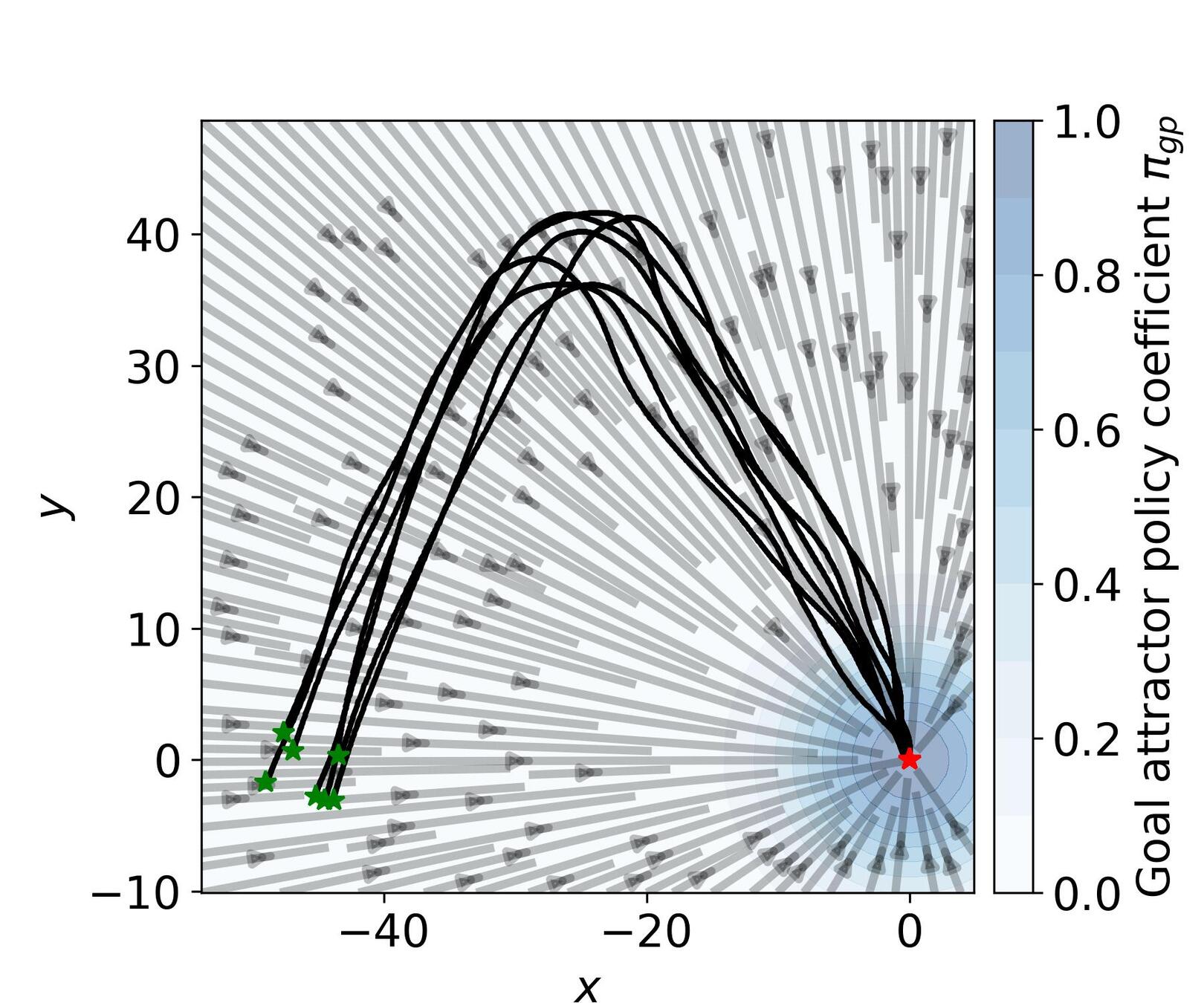}
		\caption{Goal attractor policy}
		\label{fig:MoE_conv}
	\end{subfigure}
	\hfill
	\begin{subfigure}[t]{0.2455\textwidth}
		\centering
		\includegraphics[trim={0 0 0 1cm},clip, width=\linewidth]{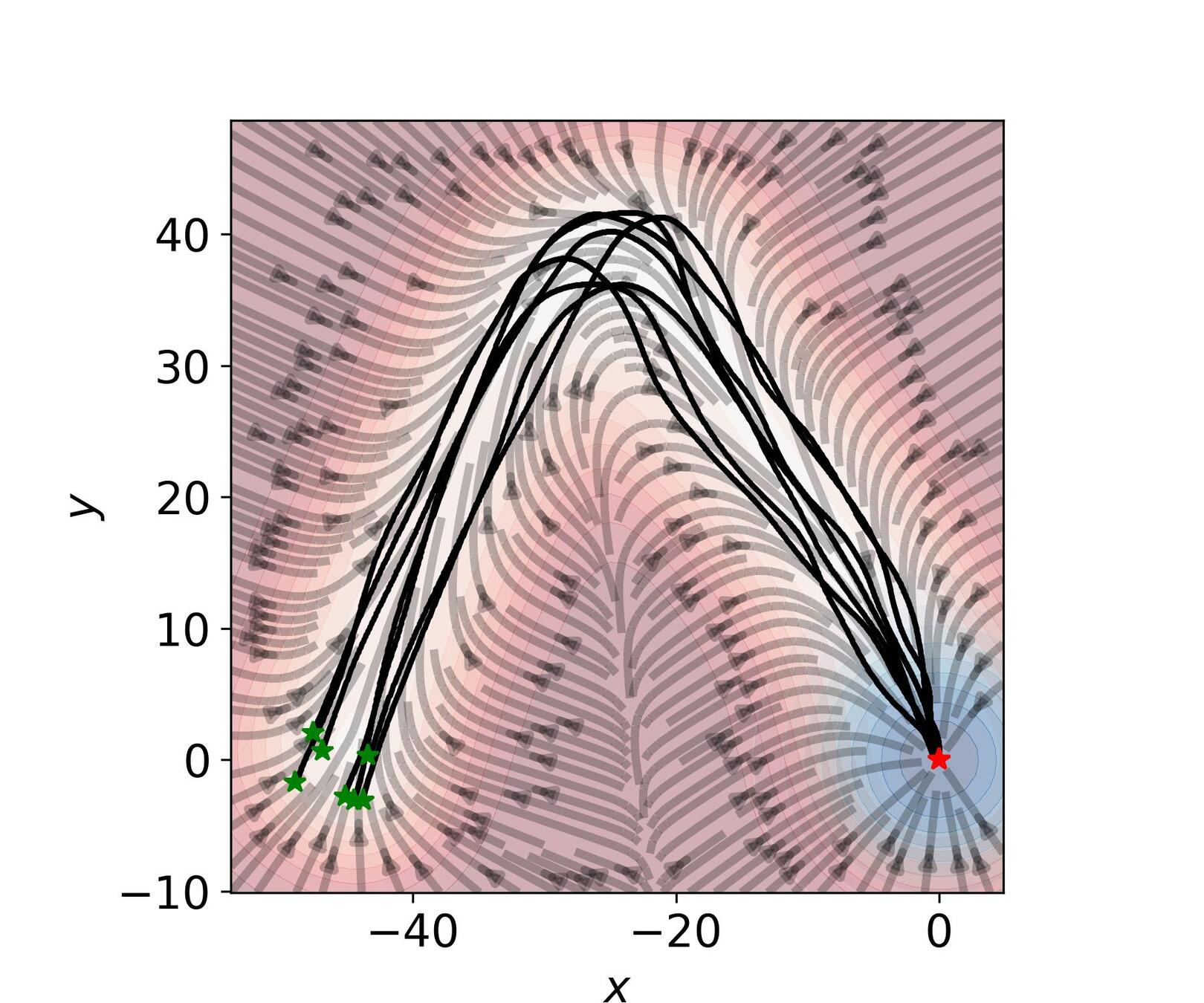}
		\caption{MoE}
		\label{fig:MoE_comb}
	\end{subfigure}
	\caption{\label{fig:MoE}Two-dimensional example showing the vector fields and the mixing coefficient activation of the MoE and the individual policies when trained on the first letter of the LASA handwriting dataset ('Angle').
		(a) shows the LfD policy, (b) illustrates the stabilizing policy, (c) displays the goal attractor policy and (d) demonstrates the combination of the individual policies using the MoE formulation.}
\end{figure*}

The vector fields of the LfD policy are exemplarily illustrated in \Cref{fig:MoE_lfd} for the first letter of the LASA handwriting dataset~\cite{LASA}, where we only consider $\bm{s}=\bm{x} \in \mathbb{R}^2$, i.e.,~no task-dependent parameter
$\bm{c}$.
The false attractors displayed in the illustration (bottom-left and bottom-right regions) highlight the generalization issues of purely data-driven machine learning approaches applied to control, which lead to undesired behavior in underrepresented regions.
This reaffirms the need for supplementary policies that empirically improve the OOD robustness and convergence behavior.

\subsection{Stabilizing policy}\label{sec:stab_policy}
Due to the aforementioned OOD generalization limitations, LfD policies typically fail to generalize beyond well-covered regions,
effectively restricting operation to the vicinity of the demonstrated trajectories.
To improve robustness in OOD states --- i.e.,~states that lie beyond the support of the observed demonstrations ---
we introduce a stabilizing policy that actively guides the robot toward regions of low epistemic uncertainty,
which typically coincide with the regions covered by the demonstration data.
For the formulation of this policy, we draw upon prior work~\cite{Franzese2021,Meszaros2022} and leverage the predictive variance of GPR (\Cref{eq:var_pred}),
which provides a closed-form epistemic uncertainty quantification that increases smoothly with distance from the training data.
Specifically, the approach generates actions along the gradient of the predictive variance, thereby directing the system toward the nearest low-uncertainty region.
Additionally, we make use of the combined kernel for position and orientation,
and incorporate, contrary to~\cite{Franzese2021,Meszaros2022}, orientation in the stabilizing policy.
Since the state space $\bm{s}_n=\left[\bm{c}^\top_n \> \bm{x}^\top_n \> \bm{\phi}^\top_n\right]^\top$ jointly encodes both Cartesian position and end-effector orientation,
the RBF kernel naturally operates on this full pose representation.
The predictive variance and its gradient therefore inherently capture uncertainty contributions from both translation and rotation,
enabling the stabilizing policy to simultaneously steer the robot in position and orientation.

For this, we derive the gradient of $v_*$ with respect to the current state $\bm{s}_*$,\footnote{For simplicity and improved readability, we write $\nabla$ instead of $\nabla_{\!\bm{s}_*}$.}
assuming the RBF kernel \Cref{eq:rbf}:
\begin{equation}
	\nabla v_* = - 2 \nabla \bm{k}_*^\top (\bm{K} + \sigma^2 \bm{I}_{N})^{-1} \bm{k}_*,
	\label{eq:sig_grad}
\end{equation}
where \ \ $\nabla \bm{k}_* = \left[\nabla k(\bm{s}_*, \bm{s}_1)^\top \> \cdots  \> \nabla k(\bm{s}_*, \bm{s}_N)^\top\right]^\top$ \ \ and $\nabla k(\bm{s}_*, \bm{s}_n) = - k(\bm{s}_*, \bm{s}_n) \bm{L} (\bm{s}_* - \bm{s}_n)$ is the gradient of the RBF kernel.
In this work we assume that, in contrast to the robot state, the task-dependent parameter $\bm{c}$ can only be controlled indirectly and thus focus on the robot state dependent terms $\nabla_{\!\bm{x}_*} v_*$, $\nabla_{\!\bm{\phi}_*} v_*$ to design the stabilizing policy.\footnote{Note that for an input of the form of $\bm{s}$, the gradient contains three distinct terms,
	$\nabla v_* = \left[(\nabla_{\!\bm{c}_*} v_*)^\top \> (\nabla_{\!\bm{x}_*} v_*)^\top \> (\nabla_{\!\bm{\phi}_*} v_*)^\top\right]^\top$.}
We normalize the gradients for position and orientation independently to unit length and
introduce the hyperparameter $\bm{K}_{\mathrm{sp}} = \{K_{{\mathrm{sp}, \bm{x}}}, K_{{\mathrm{sp}, \bm{\phi}}}\}$ to map the normalized gradients to the velocity domain.
Moreover, to regulate the magnitude of the velocities, we scale them by the respective predictive variance magnitudes $v_{\bm{c}, \bm{x}}$, $v_{\bm{c}, \bm{\phi}}$,
which are computed using \Cref{eq:var_pred} with $\bm{s}=\left[\bm{c}^\top \> \bm{x}^\top\right]^\top$ and $\bm{s}=\left[\bm{c}^\top \> \bm{\phi}^\top\right]^\top$, respectively.
This leads to vanishing velocities towards the associated demonstrated trajectories.
Based on this, we define the mean of the stabilizing policy distribution as:
\begin{equation}
	\bm{\mu}_{\mathrm{sp}} =
	\begin{bmatrix}
		- K_{{\mathrm{sp}, \bm{x}}} \frac{\nabla_{\!\bm{x}_*} v_*}{\max{(\|  \nabla_{\!\bm{x}_*} v_* \| , \varepsilon)}} v_{\bm{c}, \bm{x}} \\
		- K_{{\mathrm{sp}, \bm{\phi}}} \frac{\nabla_{\!\bm{\phi}_*} v_*}{\max{(\|  \nabla_{\!\bm{\phi}_*} v_* \| ,\varepsilon)}} v_{\bm{c}, \bm{\phi}}
	\end{bmatrix}
	,
	\label{eq:stabilizing_policy}
\end{equation}
where $\varepsilon$ is a small positive constant to prevent numerical issues.\footnote{In practice, we set it to the smallest positive floating-point number available in Python, $2.2 \times 10^{-308}$.}
Equation (\ref{eq:stabilizing_policy}) guides the robot towards regions with low epistemic uncertainty.
This effect can be observed on the vector fields in \Cref{fig:MoE_stab}.

\subsection{Goal attractor policy}\label{sec:goal_policy}
GPR alone is not guaranteed to converge to the demonstrated goal poses $\bm{s}_{M_h, h}$, as can be seen in \Cref{fig:MoE_lfd}.
While the stabilizing policy improves OOD robustness and safety, it does not explicitly enforce the robot to reach and maintain the goal pose.
To address this, we further introduce a goal-attractor policy that steers the robot toward the associated goal pose $\bm{s}_{g} = \argmax{\{k(\bm{s}_*, \bm{s}_{M_1, 1}),\ldots, k(\bm{s}_*, \bm{s}_{M_H, H})\}}$,
which we define as the demonstrated goal pose with the highest kernel activation at current state $\bm{s}_*$.
As the RBF kernel activation quantifies the similarity of two given states,
$\bm{s}_g$ corresponds to the demonstrated goal pose with the highest similarity to the current state in the combined feature space.
To guide the robot toward $\bm{s}_g$, we adopt the natural choice of following the gradient w.r.t. the robot state of this activation,
given by $\nabla k(\bm{s}_*, \bm{s}_{g})$ with $\bm{s}_{g} = \left[\bm{c}^\top_g \> \bm{x}^\top_g \> \bm{\phi}^\top_g\right]^\top$,
i.e.,~to move in the direction that increases the kernel value.

As with the stabilizing policy, we normalize the gradients for position and orientation independently and introduce the hyperparameter $\bm{K}_{\mathrm{gap}} = \{K_{{\mathrm{gap}, \bm{x}}}, K_{{\mathrm{gap}, \bm{\phi}}}\}$ to map the normalized gradients to the velocity domain.
But, in contrast, we now use the kernels $k(\bm{x}_*,\bm{x}_{g})$ and $k(\bm{\phi}_*,\bm{\phi}_{g})$ to obtain vanishing velocities towards the goal pose, and retain the positive gradient as we wish to increase the kernel activation.
Consequently, we define the mean of the goal attractor policy distribution as:
\begin{equation}
	\bm{\mu}_{\mathrm{gap}} =
	\begin{bmatrix}
		K_{{\mathrm{gap}, \bm{x}}} \frac{\nabla_{\!\bm{x}_*} k(\bm{s}_*, \bm{s}_{g})}{\max{(\|  \nabla_{\!\bm{x}_*} k(\bm{s}_*, \bm{s}_{g}) \| ,\varepsilon)}} (1-k(\bm{x}_*,\bm{x}_{g})) \\
		K_{{\mathrm{gap}, \bm{\phi}}} \frac{\nabla_{\!\bm{\phi}_*} k(\bm{s}_*, \bm{s}_{g})}{\max{(\|  \nabla_{\!\bm{\phi}_*} k(\bm{s}_*, \bm{s}_{g}) \| ,\varepsilon)}} (1-k(\bm{\phi}_*,\bm{\phi}_{g}))
	\end{bmatrix}
	.
	\label{eq:goal_attractor}
\end{equation}
The vector fields resulting from \Cref{eq:goal_attractor} are visualized in \Cref{fig:MoE_conv} for the considered illustrative example.

\subsection{MoE mixing coefficient design}\label{sec:coefficients_design}
We here outline the design choices for the mixing coefficients of \Cref{eq:MoE_all_policies}
that are automatically computed based on the uncertainty inherent in the provided demonstrations.

\subsubsection{Goal attractor policy coefficient $\pi_{\mathrm{gap}}$}
Close to the respective goal, the robot should approach it and maintain its pose there.
Consequently, the activation is designed to increase as the distance to the goal pose decreases.
To facilitate this local activation and smooth transitions between the policies, we define
\begin{equation}
	\label{eq:pi_gap}
	\pi_{\mathrm{gap}} = k(\bm{s}_*, \bm{s}_{g})
\end{equation}
as the goal attractor policy coefficient,
where $\bm{s}_{g}$ is recomputed at each iteration based on the current state $\bm{s}_*$.

\subsubsection{Stabilizing policy coefficient $\pi_{\mathrm{sp}}$}
Furthermore, we assign a high priority to staying close to demonstrations, avoiding dangerous, unseen behaviors.
Therefore, we use the variance prediction provided by GPR (\Cref{eq:var_pred}), which increases smoothly when deviating from the data, as activation.
Unlike approaches with constant coefficients~\cite{Franzese2021,Meszaros2022}, this yields a state-dependent activation that adapts according to the predictive uncertainty.

We reduce the mixing coefficient by the activation of the goal attractor policy, as we assign a higher priority to it close to the goal.
We thus define
\begin{equation}
	\label{eq:pi_sp}
	\pi_{\mathrm{sp}} = (1 - \pi_{\mathrm{gap}}) v_* = (1 - k(\bm{s}_*, \bm{s}_{g})) v_*.
\end{equation}

\subsubsection{LfD policy coefficient $\pi_{\mathrm{lfd}}$}
Taking into account $\sum_{i=1}^{P} \pi_{i} = 1$ (\Cref{sec:moe}) results in
\begin{equation}
	\label{eq:pi_lfd}
	\pi_{\mathrm{lfd}} = 1 - \pi_{\mathrm{gap}} - \pi_{\mathrm{sp}} = (1 - k(\bm{s}_*, \bm{s}_{g})) (1 - v_*)
\end{equation}
as the mixing coefficient for the LfD policy.
Hence, the LfD policy is smoothly activated near the demonstrations, while the goal attractor policy takes over close to the goal.

The mixing coefficients for each policy as well as the combined MoE output $\hat{\bm{\mu}}$, applied to the 2D example, are displayed in \Cref{fig:MoE} using contour lines.
Here, each policy acts as an expert in its respective area, while the combined policy unifies the advantages of the individual policies.

\Cref{alg:cap} summarizes this section in pseudo-code.\footnote{A definition of \textit{goalReached} is given in \Cref{sec:hyperparameter_opt}.}

\begin{algorithm}
	\caption{Skill learning and execution}\label{alg:cap}
	\begin{algorithmic}
		\State \textbf{Input:} $\{\{\bm{s}_{m,h}, \bm{\xi}_{m,h}\}^{M_h}_{m=1}\}^H_{h=1}$, $\{\bm{l}, \sigma^2, \bm{K}_\mathrm{sp}, \bm{K}_\mathrm{gap}\}$, $\bm{s}_0$
		\State $\{\bm{s}_{n}, \bm{\xi}_{n}\}^N_{n=1} \subset \{\{\bm{s}_{m,h}, \bm{\xi}_{m,h}\}^{M_h}_{m=1}\}^H_{h=1}$ \Comment{Subsample data}
		\State 	$\parbox[c]{0.4\linewidth}{
				$\bm{R} \gets \mathrm{chol}(\bm{K} + \sigma_n^2 I)$ \\
				$\bm{\alpha} \gets \bm{R}^\top \backslash (\bm{R} \backslash \bm{\Xi})$
			}$ \Comment{Train GP (cf. \cite{Rasmussen2006}, Alg. 2.1)}
		\State $\bm{s}_* \gets \bm{s}_0$ \Comment{Assign initial conditions}
		\State $i \gets 0$ \Comment{Reset iteration counter}
		\State goalReached $\gets$ false
		\State stopTriggered $\gets$ false
		\While{$\neg \ \text{goalReached} \land i < I_\mathrm{max} \land \neg \ \text{stopTriggered}$}
		\State $\bm{\mu}_\mathrm{lfd} \gets \text{\Cref{eq:mu_pred}}$ \Comment{Compute GP mean (\cite{Rasmussen2006}, Alg. 2.1)}
		\State $\bm{\mu}_\mathrm{sp} \gets \text{\Cref{eq:stabilizing_policy}}$ \Comment{Calculate stab. policy mean}
		\State $\bm{\mu}_\mathrm{gap} \gets \text{\Cref{eq:goal_attractor}}$ \Comment{Determine goal attr. policy mean}

		\State $\pi_\mathrm{gap} \gets \text{\Cref{eq:pi_gap}}$ \Comment{Calculate goal attr. mixing coeff.}
		\State $\pi_\mathrm{sp} \gets \text{\Cref{eq:pi_sp}}$ \Comment{Determine stab. mixing coeff.}
		\State $\pi_\mathrm{lfd} \gets \text{\Cref{eq:pi_lfd}}$ \Comment{Derive LfD mixing coeff.}
		\State $\hat{\bm{\xi}}_* \gets \text{\Cref{eq:MoE_all_policies}}$ \Comment{Compute control action (MoE mean)}
		\State $\bm{s}_* \gets \text{InnerControlLoop}(\hat{\bm{\xi}}_*)$ \Comment{Apply $\hat{\bm{\xi}}_*$ to robot}
		\State $\text{goalReached} \gets \text{checkGoalReached}(\bm{s}_*)$
		\State $\text{stopTriggered} \gets \text{checkExternalStop}()$
		\State $i \gets i + 1$ \Comment{Increment iteration counter}
		\EndWhile
	\end{algorithmic}
\end{algorithm}

\section{Evaluation}\label{sec:evaluation}
We evaluate our approach in four experiments of increasing complexity.
First, we investigate its performance on the LASA handwriting dataset~\cite{LASA}, omitting the task-dependent parameter $\bm{c}$,
and focusing on its OOD robustness and convergence behavior with only the robot state as input (\Cref{sec:lasa}).
We then use a real 7-DoF robot in three manipulation scenarios, namely force-conditioned grasping (\Cref{sec:regrasp}), manipulation of deformable food items (\Cref{sec:fish}) and object-centric grasping (\Cref{sec:dynamic_grasping}).\footnote{Videos of all experiments can be found as supplementary material.}
For these scenarios, task selection was guided by the aim to consider diverse contextual information, realized through the incorporation of both continuous and discrete
task-dependent parameters such as contact forces (\Cref{sec:regrasp}), object shape descriptors (\Cref{sec:fish}) and phase variables (\Cref{sec:dynamic_grasping}).
The hyperparameter values used are obtained through black box optimization as described in \Cref{sec:hyperparameter_opt} and are listed in \Cref{table:hyperparams}.

\begin{table*}
	\caption{\label{table:baseline_experiments}Quantitative evaluation (average success $S$, iterations $I$, distance $D$) on the LASA handwriting dataset.
		Results are averaged over all letters and reported as the mean and standard deviation across 20 random seeds.
		Each seed includes multiple trials per letter: 10 with random initialization (``$\bm{x}_{0,k}$ random'') and 7 with on-track initialization (first point in each of the 7 demonstrations, ``$\bm{x}_{0,k}$ on demos'').
		\textbf{Note:} Since actions are generated from the first moment of the MoE, our policy is deterministic for a fixed set of hyperparameters.
		Thus, variation across evaluation seeds stems solely from the initial poses.
		When starting on demonstrations, the initial poses for each letter are the same across all seeds, resulting in standard deviations of 0.0.}
	\begin{center}
		\begin{tabular}{clccc}
			\toprule
			Exp. & Method                                                      & $S$ [\%]           & $I$                & $D$                \\
			\midrule
			\multirow{7}{*}{\rotatebox{90}{$\bm{x}_{0,k}$ on demos}}
			     & GPR (LfD policy \ref{sec:lfd_policy}, \cite{Rasmussen2006}) & $11.0\pm0.0$       & $454.9\pm0.0$      & $2.11\pm0.00$      \\
			     & GPR \& stab. (\ref{sec:stab_policy})                        & $86.7\pm0.0$       & $165.8\pm0.0$      & $0.23\pm0.00$      \\
			     & GPR \& goal attr. (\ref{sec:goal_policy})                   & $99.5\pm0.0$       & $\bm{113.3\pm0.0}$ & $0.36\pm0.00$      \\
			     & \textbf{Proposed framework}                                 & $\bm{100.0\pm0.0}$ & $114.3\pm0.0$      & $\bm{0.20\pm0.00}$ \\[1pt]
			\cdashline{2-5}\noalign{\vskip 2pt}
			     & Diffusion Policy (\hspace{1sp}\cite{Chi2024})               & $9.4\pm1.0$        & $477.5\pm2.0$      & $43.61\pm1.96$     \\
			     & SEDS (likelihood, \cite{KhansariZadeh2011})                 & $99.8\pm0.3$       & $125.8\pm3.9$      & $2.99\pm1.29$      \\
			\midrule
			\multirow{7}{*}{\rotatebox{90}{$\bm{x}_{0,k}$ random}}
			     & GPR (LfD policy \ref{sec:lfd_policy}, \cite{Rasmussen2006}) & $5.1\pm1.0$        & $480.5\pm3.8$      & $8.07\pm0.29$      \\
			     & GPR \& stab. (\ref{sec:stab_policy})                        & $86.7\pm0.0$       & $127.5\pm1.7$      & $\bm{0.65\pm0.03}$ \\
			     & GPR \& goal attr. (\ref{sec:goal_policy})                   & $58.6\pm2.2$       & $270.1\pm9.0$      & $7.40\pm0.30$      \\
			     & \textbf{Proposed framework}                                 & $\bm{100.0\pm0.0}$ & $\bm{72.0\pm1.7}$  & $\bm{0.65\pm0.03}$ \\[1pt]
			\cdashline{2-5}\noalign{\vskip 2pt}
			     & Diffusion Policy (\hspace{1sp}\cite{Chi2024})               & $10.9\pm1.1$       & $470.7\pm4.2$      & $60.31\pm4.38$     \\
			     & SEDS (likelihood, \cite{KhansariZadeh2011})                 & $99.5\pm0.7$       & $100.9\pm5.6$      & $3.52\pm0.99$      \\
			\bottomrule
		\end{tabular}
	\end{center}
\end{table*}

\subsection{Hyperparameter optimization}\label{sec:hyperparameter_opt}
Given the influence of GPR hyperparameters on the supplementary policies (e.g.,~through the shared kernel length over all policies),
the optimization of the combined hyperparameters $\bm{\theta} = \{\bm{l}, \sigma^2, \bm{K}_{\mathrm{sp}}, \bm{K}_{\mathrm{gap}}\}$ is performed jointly
using the CMA-ES algorithm~\cite{Hansen2016}, which is used due to its derivative-free nature and its ability to efficiently search complex parameter spaces.
To this end, the skill learning and execution procedure, as summarized in \Cref{alg:cap}, is executed ten times for the provided demonstrations,
while the robot behavior (\textit{InnerControlLoop}) is simulated by computing the next pose at each time step using $\bm{x}_{t+1, i} = \bm{x}_{t,i} + \Delta t \dot{\bm{x}}_{t,i}$ and $\bm{\phi}_{t+1, i} = \bm{\phi}_{t,i} + \Delta t \dot{\bm{\phi}}_{t,i}$.
To optimize performance both on and off the demonstrated trajectories, the starting poses $\{\bm{x}_{0,k}, \bm{\phi}_{0,k}\}^{10}_{k=1}$ are sampled randomly within the demonstrated input space,
naturally accounting for sensitivity by evaluating performance across diverse starting poses.
In cases where task-dependent parameters are employed, the recorded values $\{\bm{c}_{m,h}\}^{M_h}_{m=1}$ from a random demonstration $h$ are re-injected into the policy.

Each of the ten simulated trials is terminated when the robot stayed in the goal region for ten iterations (\textit{goalReached}) or $I_{\mathrm{max}} = 500$ iterations are reached.
Defining $\bm{p} = \left[\bm{x}^\top \> \bm{\phi}^\top\right]^\top \in \mathbb{R}^{\mathcal{P}}$ as the end-effector pose,
the robot has reached the goal region if each component $p_{*,r}$ of the current end-effector pose $\bm{p}_*$ lies within the distance $d_r = 0.01 \cdot d_{\mathrm{max}, r}$
to one of the demonstrated goal poses $\{\bm{p}_{M_h,h}\}^{H}_{h=1}$,
with $d_{\mathrm{max}, r} = \max\limits_{m,h}{p_{m,h,r}} - \min\limits_{m,h}{p_{m,h,r}}$ describing the max-min range across all demonstrations for the specific entry and $r \in [0,\mathcal{P}]$.
We classify trials in which the robot successfully remained in the goal region as successful $S_k = 1$; otherwise, they are classified as unsuccessful $S_k = 0$.
Further, we measure the number of iterations $i_k \in [0,I_{\mathrm{max}}]$ and the mean distance between the executed poses and the closest demonstrated poses across all time steps $D_k = \frac{1}{i_k} \sum_{j=0}^{i_k} \min\limits_{m,h}{\left\|  \bm{p}_{m,h} - \bm{p}_{j} \right\| }$.
We define the multi-objective cost function as an equally weighted sum:
\begin{equation}
	\min_{\bm{\theta}}{J} = (1 - S) + \frac{I}{I_{\mathrm{max}}} + \frac{D}{\sqrt{\sum_{r=1}^{\mathcal{P}} d_{\mathrm{max}, r}^2}},
\end{equation}
where, $S$, $I$ and $D$ are the average success, iterations and distance over ten trials, thus all objectives lie \mbox{within $[0,1]$}.
We further constrain the search space of the hyperparameters in $\bm{\theta}$ to 1\% - 100\% of their respective max-min ranges.

The hyperparameters for the LASA handwriting dataset experiment (\Cref{sec:lasa}) are obtained within 500 optimization iterations using a combined objective function and a single set of hyperparameters for all letters.
As the robot state and action spaces remain the same, we optimize the hyperparameters for the real-robot experiments (\Cref{sec:regrasp,sec:fish,sec:dynamic_grasping}) over 500 optimization iterations using the training data of the experiment described in \Cref{sec:dynamic_grasping}
and then manually refined the task-parameter-dependent kernel length for the experiments described in \Cref{sec:regrasp,sec:fish}, as only the domain of $\bm{c}$ varies across tasks.
Starting from the CMA-ES-derived initial value (\Cref{sec:dynamic_grasping}), $l_{\bm{c}}$ was tuned via a unit-step search within the bounds of the demonstrated $\bm{c}$ range.
Each candidate value was evaluated over five independent trials initialized at the demonstrated start poses.
The parameter was incremented until entering, and subsequently leaving, the region of 100\% success rate,
thereby identifying the contiguous interval of tested values achieving perfect success.
The final value of $l_{\bm{c}}$ was chosen as the median of this interval.
This criterion emphasizes operational robustness and is consistent with our sensitivity analysis (\Cref{sec:c-perturbations}),
which indicates approximately symmetric performance degradation around the unperturbed value, confirming robustness against $\bm{c}$ perturbations up to $\pm 2.5\,\text{N}$.
We selected this strategy as it provides a fast and practical solution,
while a direct optimization of the hyperparameters, as for \ref{sec:dynamic_grasping}, would also be possible.

\begin{figure*}
	\centering
	\includegraphics[width=1.0\textwidth]{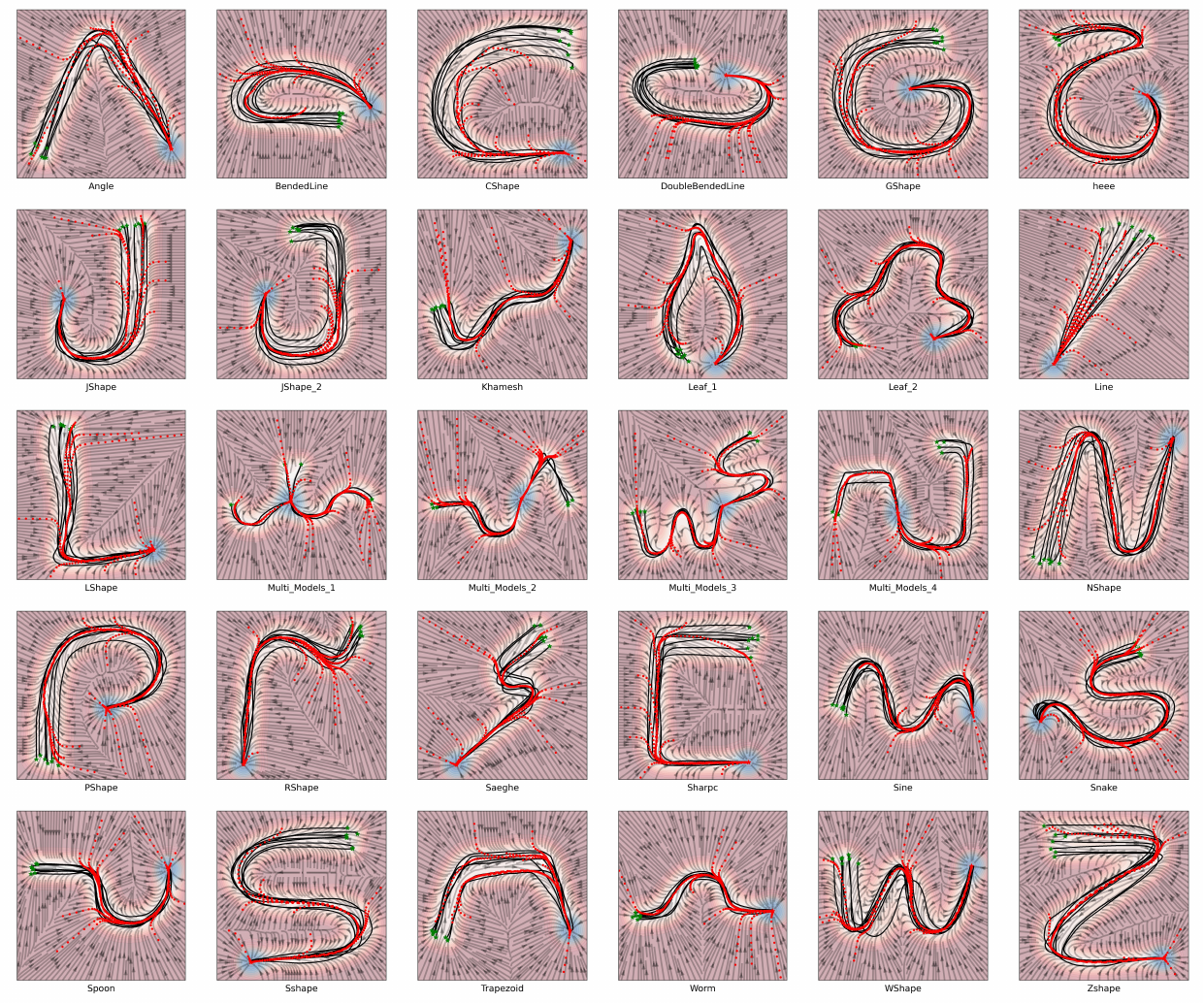}
	\caption{\label{fig:LASA_random}Reproductions for all letters of the LASA dataset when $\bm{x}_{0,k}$ is placed ten times randomly within the demonstrated input ranges
		showing the executed paths, vector fields and mixing coefficients of the MoE (refer to \Cref{fig:MoE} for clarification).}
\end{figure*}

\subsection{{LASA} handwriting dataset}\label{sec:lasa}
First, we quantitatively evaluate our approach on the LASA handwriting dataset~\cite{LASA}
and compare it with different combinations of the individual policies\footnote{Notably, this evaluation explicitly includes GPR as a standalone LfD policy, which serves as a small-data behavioral cloning baseline.}  and two established baseline approaches: SEDS~\cite{KhansariZadeh2011} and Diffusion Policy~\cite{Chi2024}.
In this experiment, we focus on the reproduction of the demonstrated handwriting motions with different initial conditions, showing the approach's OOD robustness and empirical convergence behavior.
Thus, we define $\bm{s}=\bm{x} \in \mathbb{R}^2$, $\bm{\xi}=\dot{\bm{x}} \in \mathbb{R}^2$, omitting the task-dependent parameter $\bm{c}$.
We perform two sets of simulations, similar to those described in \Cref{sec:hyperparameter_opt}, for all letters:
one in which $\bm{x}_{0,k}$ is placed at the initial pose of each of the seven demonstrations and one in which $\bm{x}_{0,k}$ is placed ten times randomly within the demonstrated input ranges.
As evaluation metrics, we use $S$, $I$ and $D$ (see \Cref{sec:hyperparameter_opt}).
For Diffusion Policy, the \textit{state-based environment}\footnote{Available at \url{https://github.com/real-stanford/diffusion\_policy}.}  is used, and the \textit{likelihood} criterion is applied for
SEDS~\cite{LASA}.
The provided source code is adapted to the dataset and experimental setup, while the hyperparameters are kept at their default values ---
SEDS hyperparameters are already tuned for this task, and Diffusion Policy hyperparameters follow commonly employed configurations for low-dimensional control tasks.
Comprehensive Diffusion Policy tuning is omitted due to the substantially higher computational effort required for both training and inference compared to our method.
This disparity is reflected in the significantly higher inference time of Diffusion Policy (see \Cref{table:inference_time}).

\Cref{table:baseline_experiments} shows the averaged results over 20 seeds, each comprising multiple trials (ten for random starting and seven for on-track starting) per letter,
and \Cref{fig:LASA_random} illustrates the executed trajectories, vector fields, and mixing coefficients of the MoE for a single seed across all letters of the LASA handwriting dataset,
where $\bm{x}_{0,k}$ is randomly initialized ten times within the demonstrated input ranges.
For completeness, the corresponding results for the same seed, with $\bm{x}_{0,k}$ initialized at the starting pose of each of the seven demonstrations,
are shown in \Cref{fig:LASA_on_demo}.

\subsection{Force-conditioned grasping}\label{sec:regrasp}
\begin{figure*}
	\centering
	\begin{subfigure}[b]{0.2455\textwidth}
		\centering
		\includegraphics[trim={8.25cm 0 4.5cm 0},clip, width=0.375\linewidth]{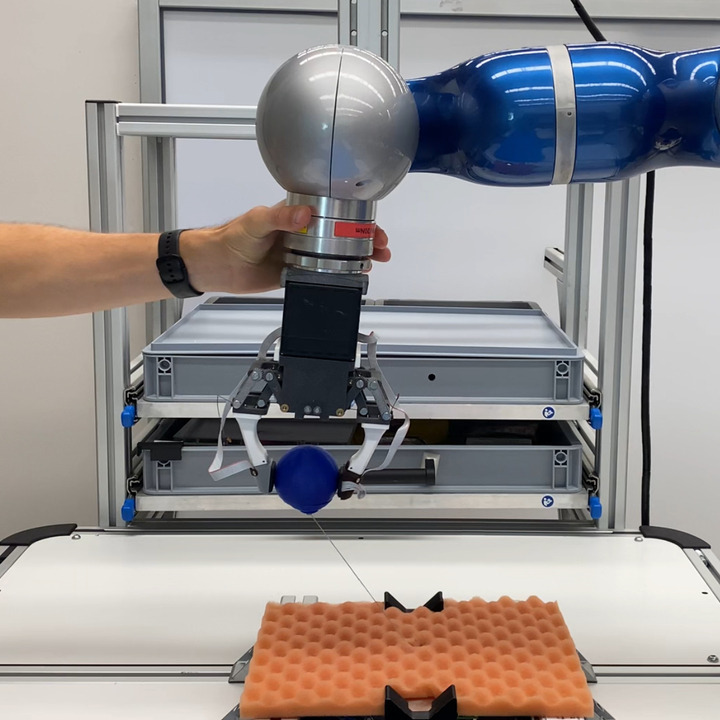}
		\includegraphics[trim={8.25cm 0 4.5cm 0},clip, width=0.375\linewidth]{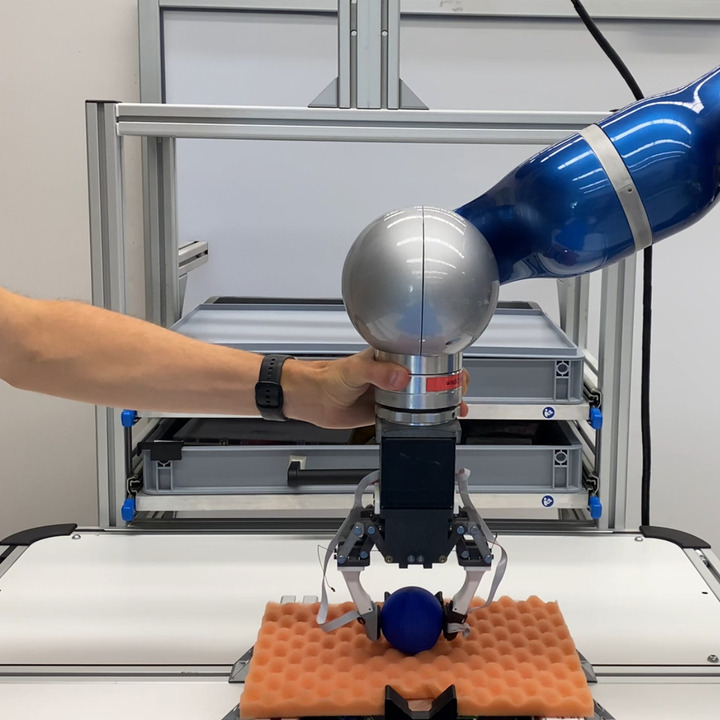}
		\caption{Goal and grasp pose}
		\label{fig:regrasp_goal_grasp_pose}
	\end{subfigure}
	\begin{subfigure}[b]{0.2455\textwidth}
		\centering
		\includegraphics[trim={0 0.5cm 0 0.25cm},clip, width=\linewidth]{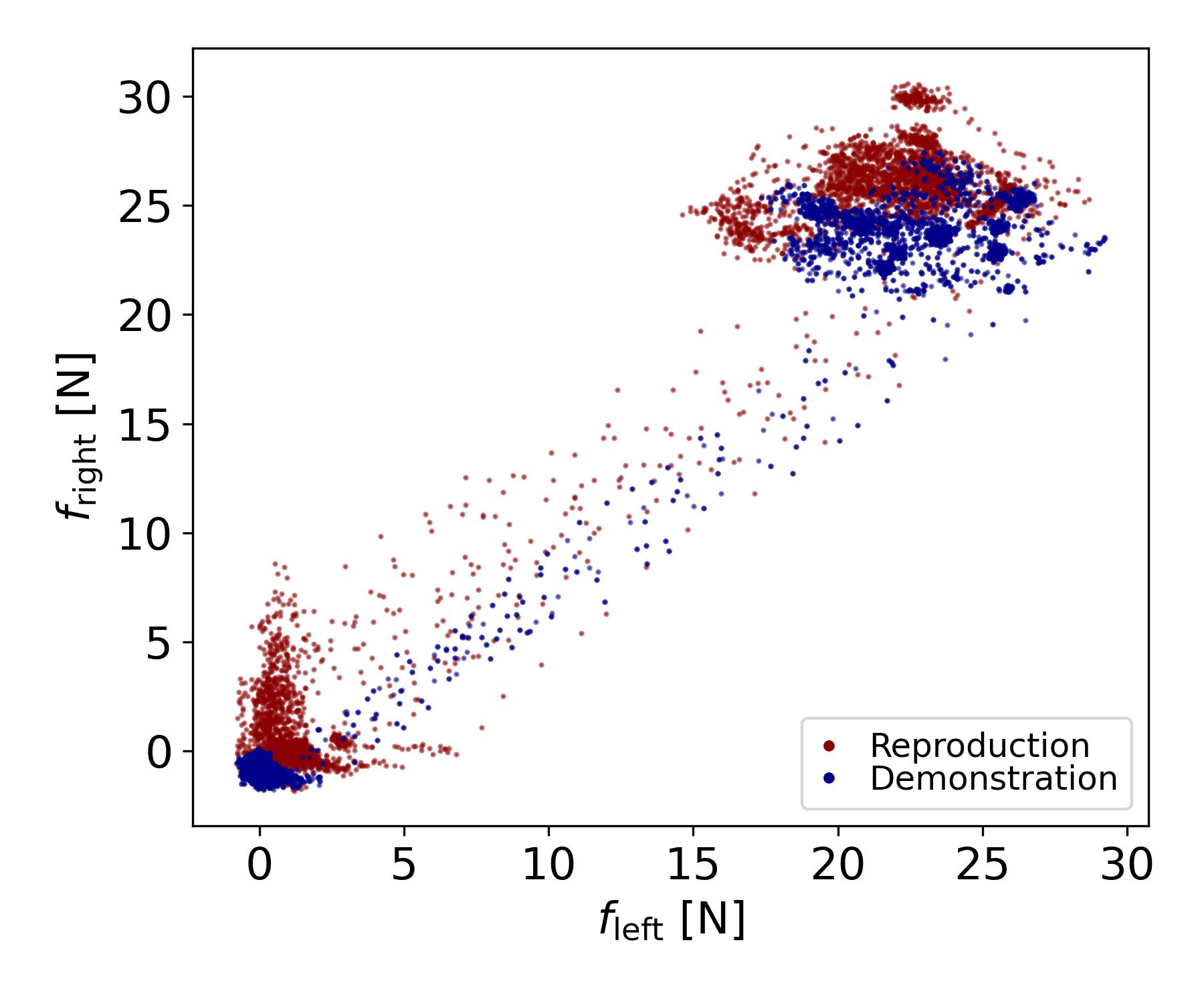}
		\caption{Task-parameters}
		\label{fig:regrasp_forces}
	\end{subfigure}
	\begin{subfigure}[b]{0.2455\textwidth}
		\centering
		\includegraphics[trim={0 0.5cm 0 0.25cm},clip, width=\linewidth]{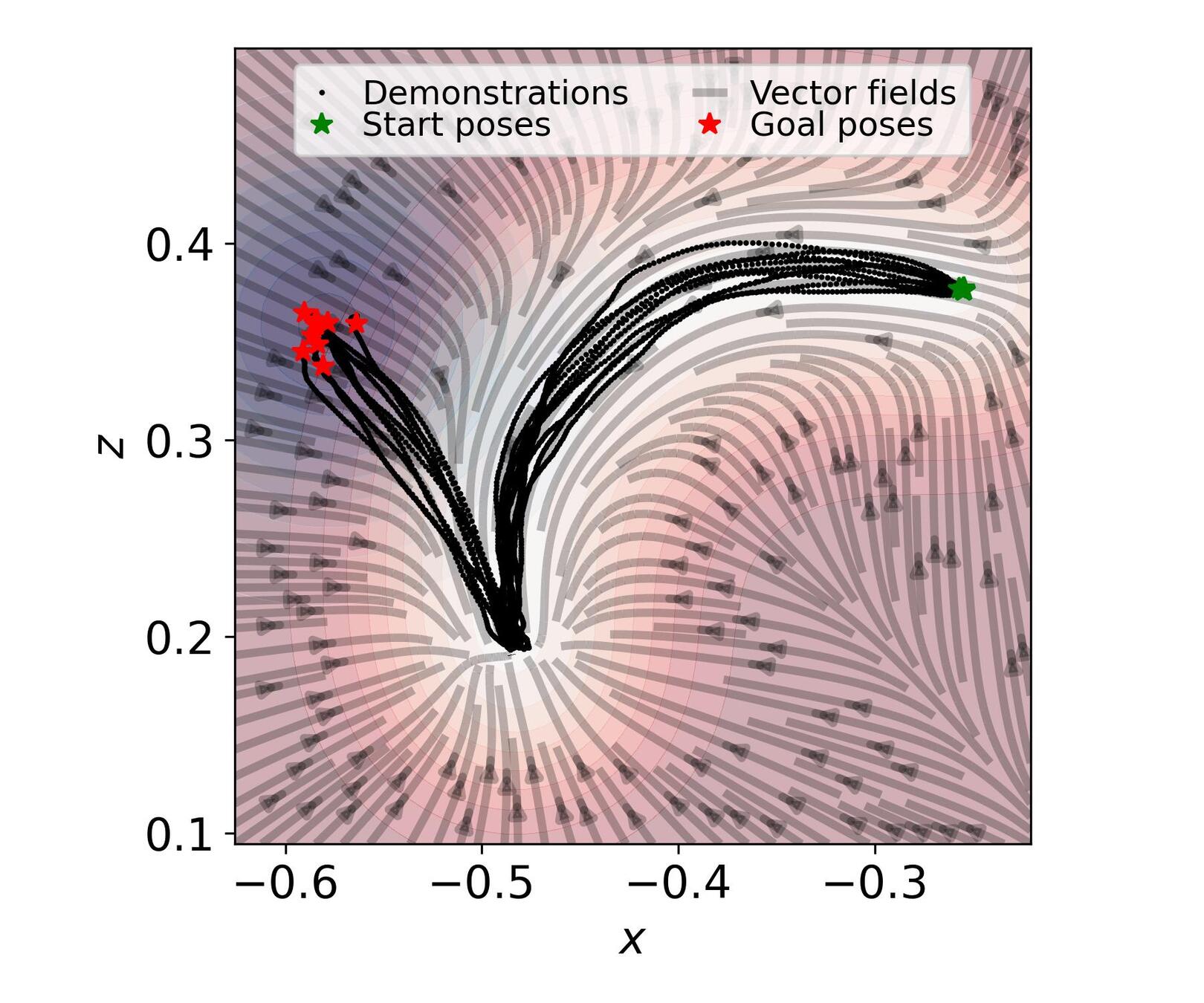}
		\caption{Pre-grasp vector fields}
		\label{fig:regrasp_reproduced_grasp}
	\end{subfigure}
	\begin{subfigure}[b]{0.2455\textwidth}
		\centering
		\includegraphics[trim={0 0.5cm 0 0.25cm},clip, width=\linewidth]{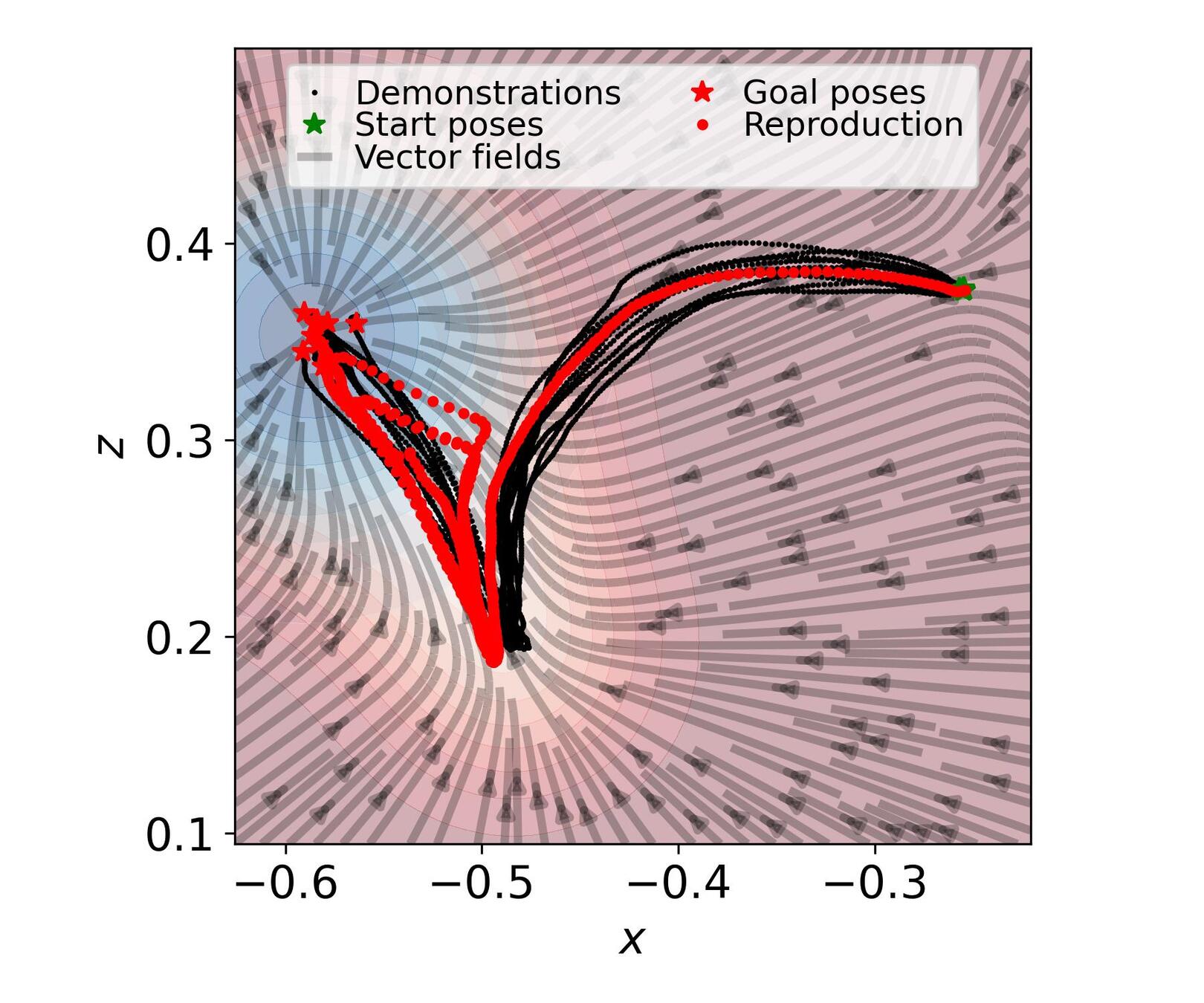}
		\caption{Post-grasp vector fields}
		\label{fig:regrasp_reproduced_goal}
	\end{subfigure}
	\begin{subfigure}[b]{0.2455\textwidth}
		\centering
		\includegraphics[trim={0 0.5cm 0 0.25cm},clip, width=\linewidth]{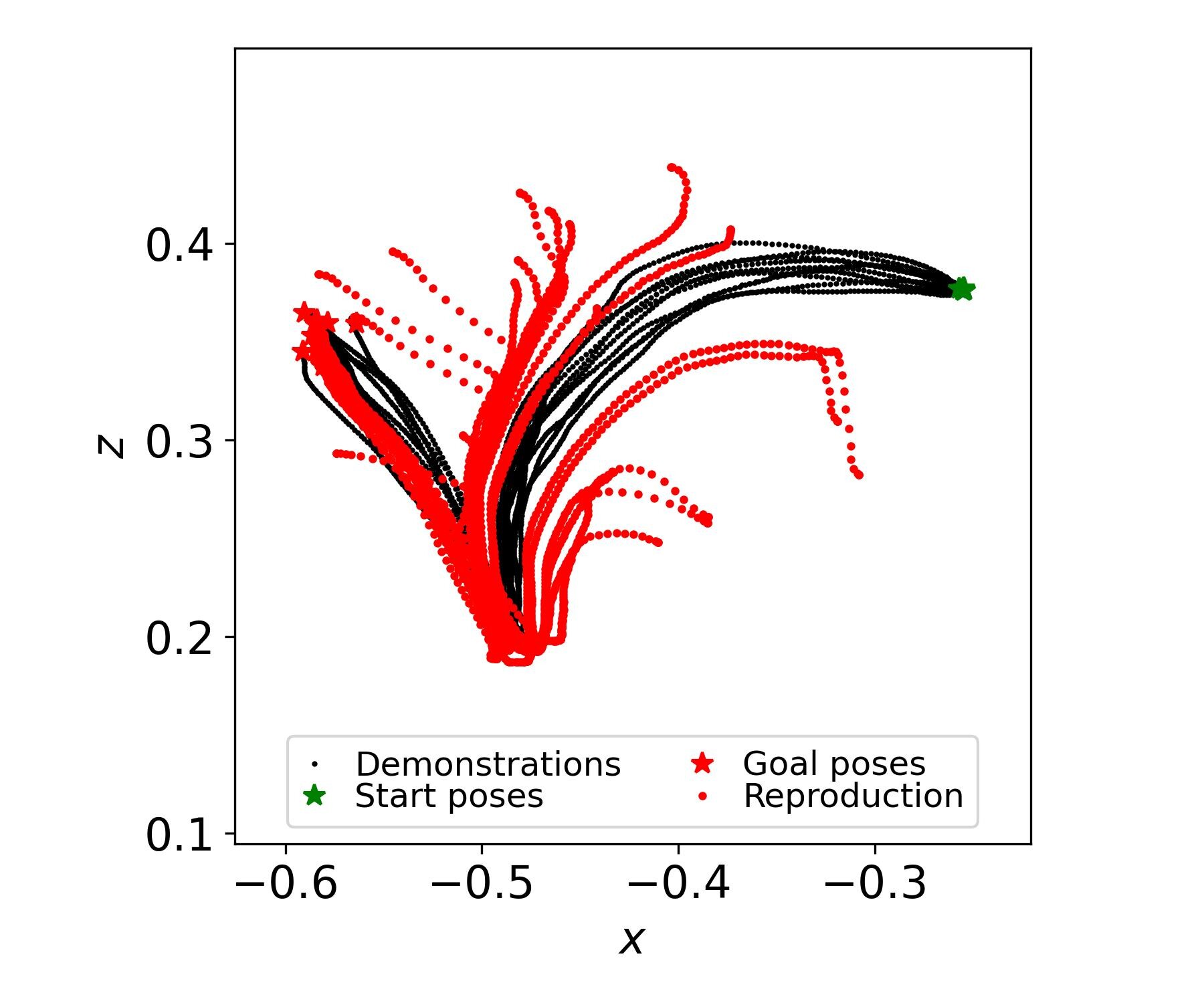}
		\caption{Reproductions}
		\label{fig:regrasp_reproduction}
	\end{subfigure}
	\begin{subfigure}[b]{0.2455\textwidth}
		\centering
		\includegraphics[trim={0 0.5cm 0 0.25cm},clip, width=\linewidth]{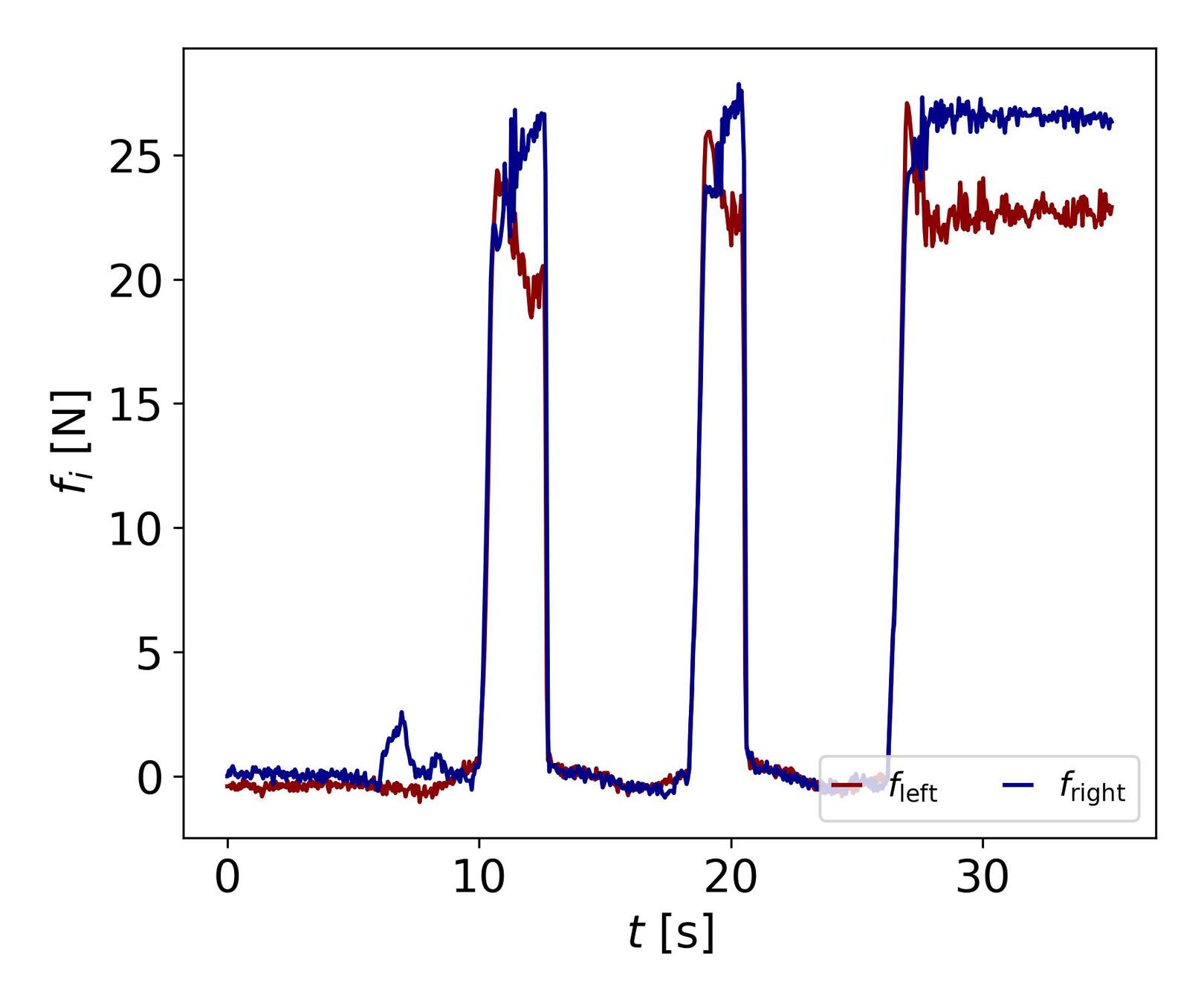}
		\caption{Re-grasp behavior}
		\label{fig:regrasp_forces_time}
	\end{subfigure}
	\begin{subfigure}[b]{0.2455\textwidth}
		\centering
		\includegraphics[trim={0 0.5cm 0 0.25cm},clip, width=\linewidth]{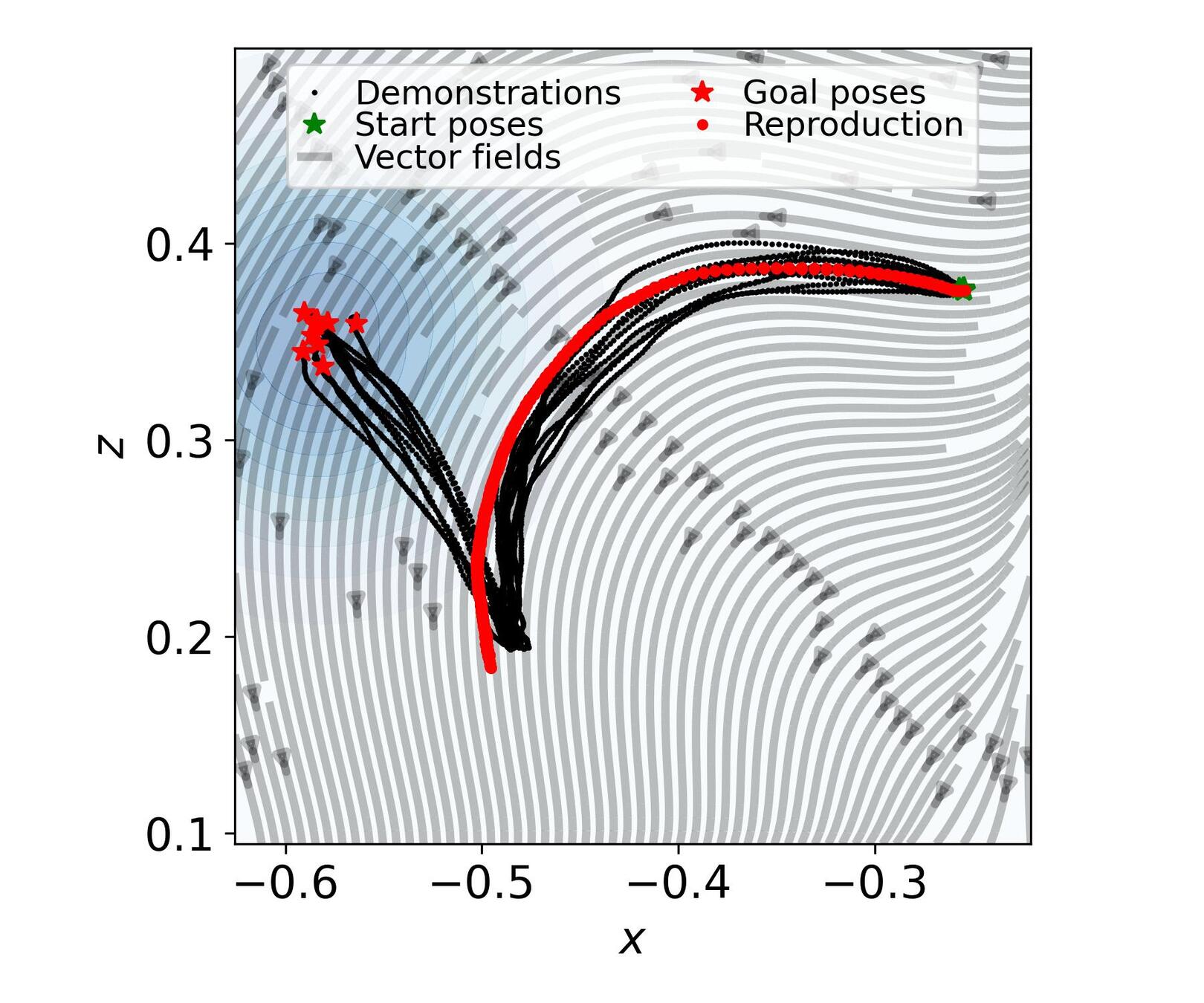}
		\caption{Without stabilizing policy}
		\label{fig:regrasp_wo_stab}
	\end{subfigure}
	\begin{subfigure}[b]{0.2455\textwidth}
		\centering
		\includegraphics[trim={0 0.5cm 0 0.25cm},clip, width=\linewidth]{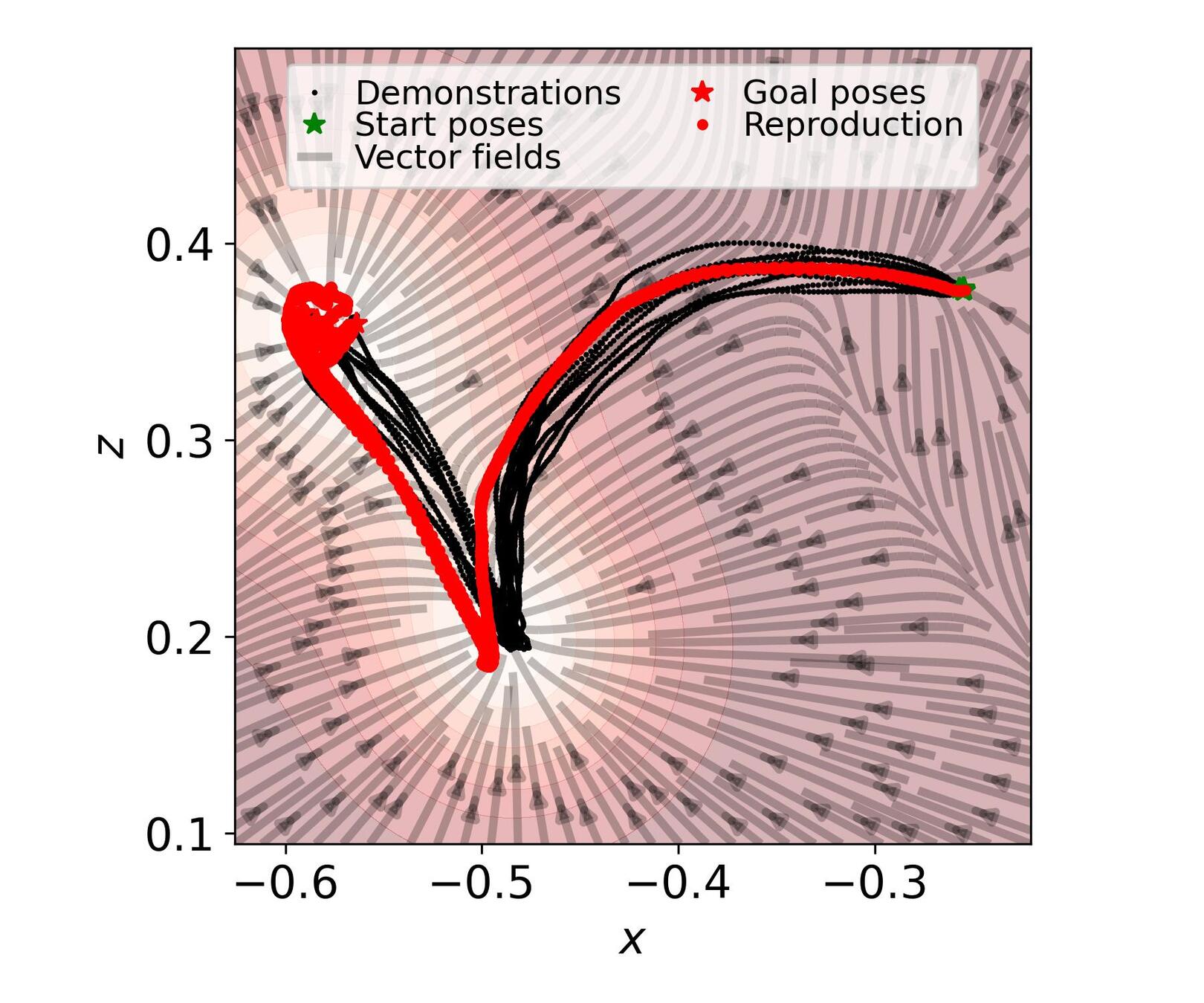}
		\caption{Without goal attractor policy}
		\label{fig:regrasp_wo_goal}
	\end{subfigure}
	\caption{\label{fig:regrasp}Demonstrations and reproductions of the force-conditioned grasping experiment:
		(a) illustrates the demonstrated grasp and goal pose,
		(b) shows the reproduced trajectories from the quantitative evaluation (upper part of \Cref{table:real_robot_experiments}) in the x-z-plane,
		(c) illustrates the demonstrated and reproduced task-parameters from the quantitative evaluation,
		(d) presents the reproduced task-parameter evolution for a trial with multiple re-grasps where the gripper is manually opened by the user,
		(e), (f) display the pre- and post-grasp MoE vector fields at $\bm{s}_{0}$ and $\bm{s}_{I}$ for this trial and
		(g), (h) display the vector fields and observed robot trajectories when removing either the stabilizing or the goal attractor policy within the context-adaptive policy framework.}
\end{figure*}

In the second experiment, we demonstrate our approach's robust transferability to real robots and its ability to react to changing task-parameter values at runtime.
For this, we record ten trajectories via kinesthetic teaching using a 7-DoF KUKA LWR robotic arm equipped with a variable stiffness parallel gripper,
where the robot picks up a 3D-printed sphere from a table and transports it to a demonstrated goal pose (\Cref{fig:regrasp_goal_grasp_pose}).
We make use of the incorporated tactile sensors in both fingertips to derive the forces exerted by the object
and define $\bm{c} = \left[f_{\mathrm{left}} \> f_{\mathrm{right}} \right]^\top$ as task-parameter.
In addition, the input includes the robot position and orientation as Euler vector, i.e.,~$\mathcal{I} = 8$.
The demonstrated and reproduced task-parameters can be seen in \Cref{fig:regrasp_forces}.

Similar to the experiments in \Cref{sec:lasa}, we quantitatively evaluate our approach and different combinations of the individual policies on this task and compare it with Diffusion Policy~\cite{Chi2024}.\footnote{A comparison with SEDS, as in \Cref{sec:lasa}, is not feasible in this and the subsequent experiments, since SEDS is restricted to specific input and output variables due to the incorporation of stability constraints.}
For this, we perform 20 trials, where the initial poses $\{\bm{x}_{0,k}, \bm{\phi}_{0,k}\}$ are drawn uniformly within the demonstrated input ranges, as shown in \Cref{fig:regrasp_reproduction}.
Our context-adaptive policy, whose vector fields are exemplarily displayed for a single trial at $\bm{s}_{0}$ (\Cref{fig:regrasp_reproduced_grasp}) and $\bm{s}_{I}$ (\Cref{fig:regrasp_reproduced_goal}), is used to compute the end-effector velocity $\dot{\bm{x}}_d$, $\dot{\bm{\phi}}_d$ and the gripper joint velocity $\dot{q}_g$ in a 20~Hz loop, hence $\mathcal{O} = 7$.
Due to the longer task horizon compared to \Cref{sec:lasa}, we define $I_{\mathrm{max}} = 1000$, and add a 1~mm or 1$^{\circ}$ offset to $d_r$, depending on whether it represents position or orientation, to account for measurement inaccuracies.
Since this is a real-world task, in addition to $S$, $I$ and $D$, we use the number of collisions (\textit{Coll.}) as comparison metric,
which measures all contact events including unintentional collisions with the target object, collisions with the surrounding environment, and self-collisions of the robot.
Collisions are automatically detected based on joint configurations (self-collisions) and torque measurements; however, to prevent damage to the robot and gripper, the authors proactively stopped the robot if hard collisions were to be expected.
Note that such events would have inevitably led to collision events detected via torque readings.
The results are displayed in the upper part of \Cref{table:real_robot_experiments} and \Cref{fig:regrasp_reproduction}.

Furthermore, we demonstrate the approach's reactivity by opening the gripper manually such that the object drops from the gripper.
Due to the task-parameter conditioning, the robot attempts to re-grasp the sphere at its original place where it is pulled back via a cable.
The resulting task-parameters are exemplified in \Cref{fig:regrasp_forces_time} for the case where the gripper is opened twice.
The corresponding trajectory is displayed in addition to the vector fields in \Cref{fig:regrasp_reproduced_goal}.
Moreover, we show that the proposed combination of policies is essential for accurate reproduction by removing either the stabilizing (\Cref{fig:regrasp_wo_stab}) or the goal attractor policy (\Cref{fig:regrasp_wo_goal}).

\begin{table*}
	\caption{\label{table:real_robot_experiments}Quantitative evaluation (average success $S$, iterations $I$, distance $D$, collisions \textit{Coll.}) on the real-robot experiments described in \Cref{sec:regrasp,sec:fish}.
		The results are reported as the mean and standard deviation computed over 20 trials.}
	\begin{center}
		\begin{tabular}{clcccc}
			\toprule
			Exp.
			 & Method                                                      & $S$ [\%]           & $I$                  & $D$                  & \textit{Coll.} \\
			\midrule
			\multirow{5}{*}{\rotatebox{90}{\Cref{sec:regrasp}}}
			 & GPR (LfD policy \ref{sec:lfd_policy}, \cite{Rasmussen2006}) & $0.0\pm0.0$        & $1000.0\pm0.0$       & $0.096\pm0.037$      & $16$           \\
			 & GPR \& stab. (\ref{sec:stab_policy})                        & $55.0\pm49.7$      & $694.1\pm357.5$      & $0.022\pm0.004$      & $\bm{0}$       \\
			 & GPR \& goal attr. (\ref{sec:goal_policy})                   & $5.0\pm21.8$       & $851.5\pm257.2$      & $0.088\pm0.043$      & $16$           \\
			 & \textbf{Proposed framework}                                 & $\bm{100.0\pm0.0}$ & $\bm{264.3\pm68.1}$  & $\bm{0.017\pm0.006}$ & $\bm{0}$       \\[1pt]
			\cdashline{2-6}\noalign{\vskip 2pt}
			 & Diffusion Policy (\hspace{1sp}\cite{Chi2024})               & $0.0\pm0.0$        & --                   & $0.076\pm0.030$      & $20$           \\
			\midrule
			\multirow{5}{*}{\rotatebox{90}{\Cref{sec:fish}}}
			 & GPR (LfD policy \ref{sec:lfd_policy}, \cite{Rasmussen2006}) & $0.0\pm0.0$        & $1000.0\pm0.0$       & $0.148\pm0.038$      & $7$            \\
			 & GPR \& stab. (\ref{sec:stab_policy})                        & $0.0\pm0.0$        & $1000.0\pm0.0$       & $0.024\pm0.002$      & $\bm{0}$       \\
			 & GPR \& goal attr. (\ref{sec:goal_policy})                   & $20.0\pm40.0$      & $839.3\pm291.3$      & $0.097\pm0.074$      & $3$            \\
			 & \textbf{Proposed framework}                                 & $\bm{95.0\pm21.8}$ & $\bm{370.1\pm165.0}$ & $\bm{0.019\pm0.004}$ & $\bm{0}$       \\[1pt]
			\cdashline{2-6}\noalign{\vskip 2pt}
			 & Diffusion Policy (\hspace{1sp}\cite{Chi2024})               & $0.0\pm0.0$        & $1000.0\pm0.0$       & $0.211\pm0.226$      & $19$           \\
			\bottomrule
		\end{tabular}
	\end{center}
\end{table*}

\subsection{Manipulation of deformable food items}\label{sec:fish}
\begin{figure}
	\centering
	\begin{subfigure}[b]{0.2355\textwidth}
		\centering
		\includegraphics[width=0.75\linewidth]{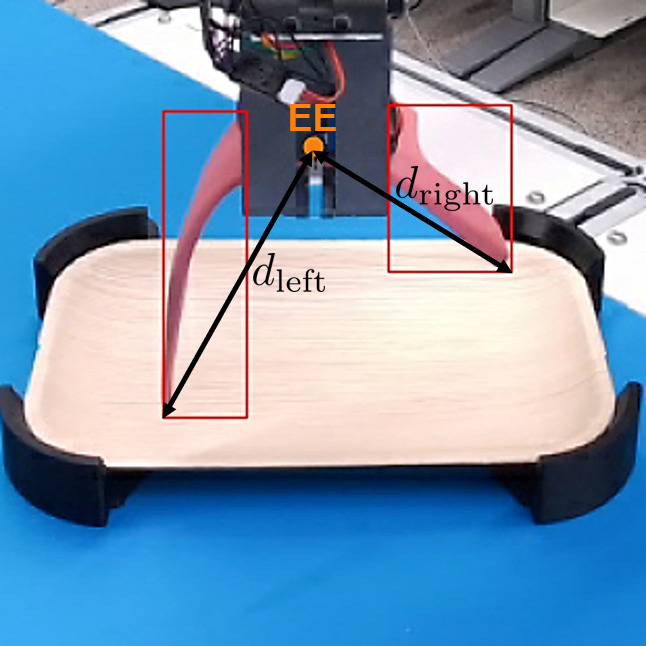}
		\caption{Definition $d_\mathrm{left}$, $d_\mathrm{right}$}
		\label{fig:task-param-def}
	\end{subfigure}
	\begin{subfigure}[b]{0.2355\textwidth}
		\centering
		\includegraphics[trim={0 0.45cm 0 0.25cm},clip, width=\linewidth]{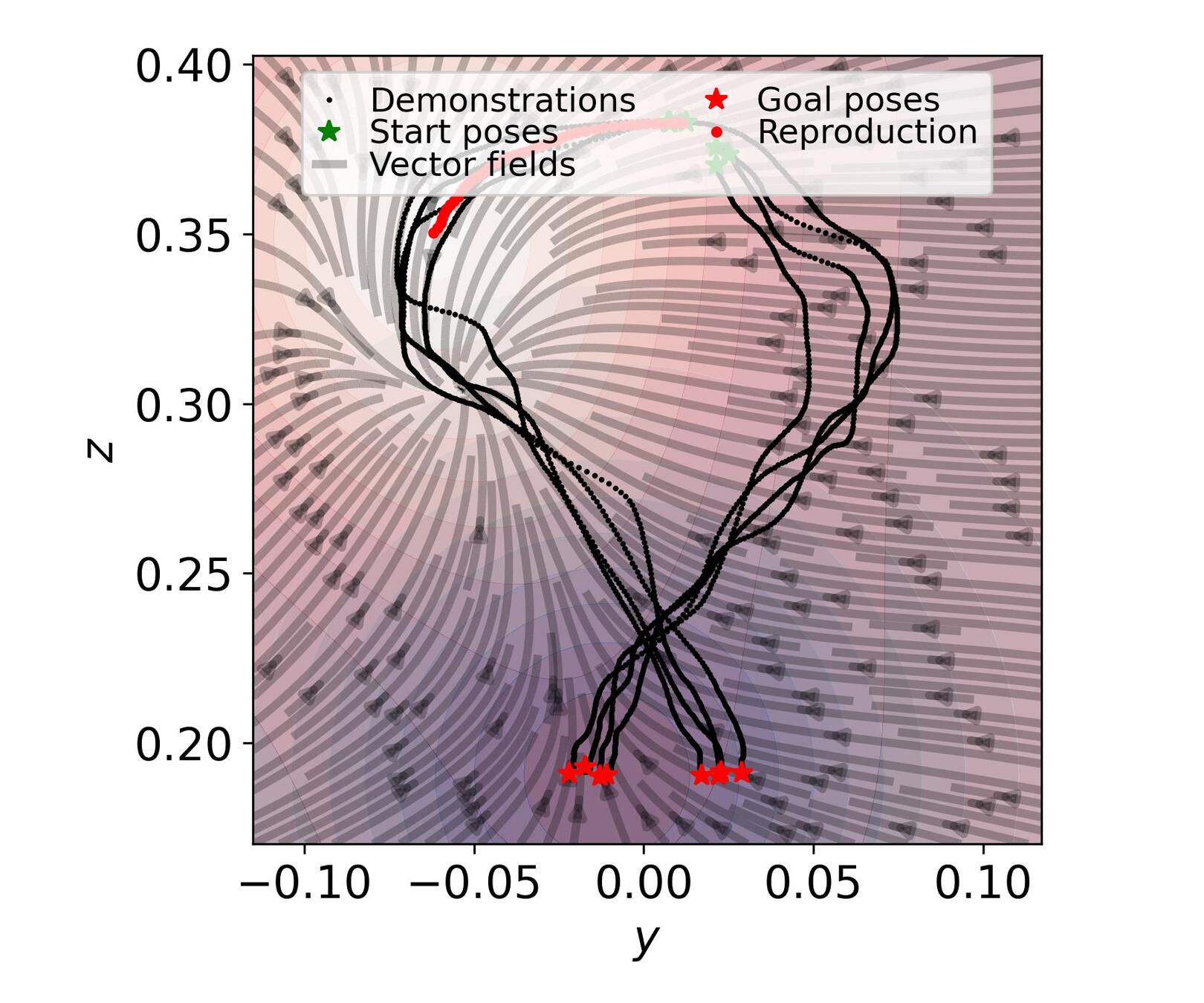}
		\caption{Vector fields at $t=2.5s$}
		\label{fig:fish_reproduced_50}
	\end{subfigure}
	\begin{subfigure}[b]{0.2355\textwidth}
		\centering
		\includegraphics[trim={0 0.45cm 0 0.25cm},clip, width=\linewidth]{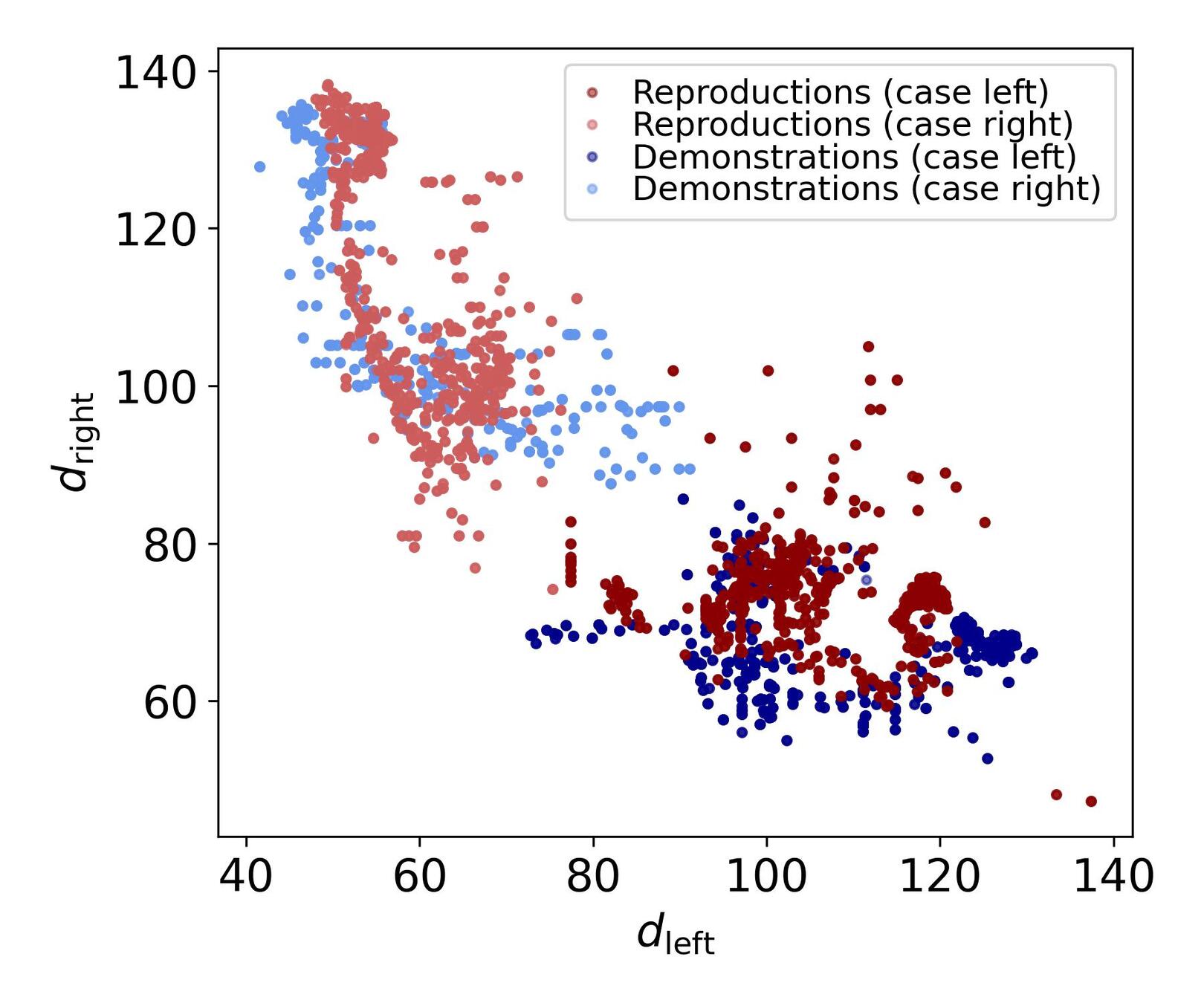}
		\caption{Task-parameters}
		\label{fig:fish_dist}
	\end{subfigure}
	\begin{subfigure}[b]{0.2355\textwidth}
		\centering
		\includegraphics[trim={0 0.45cm 0 0.25cm},clip, width=\linewidth]{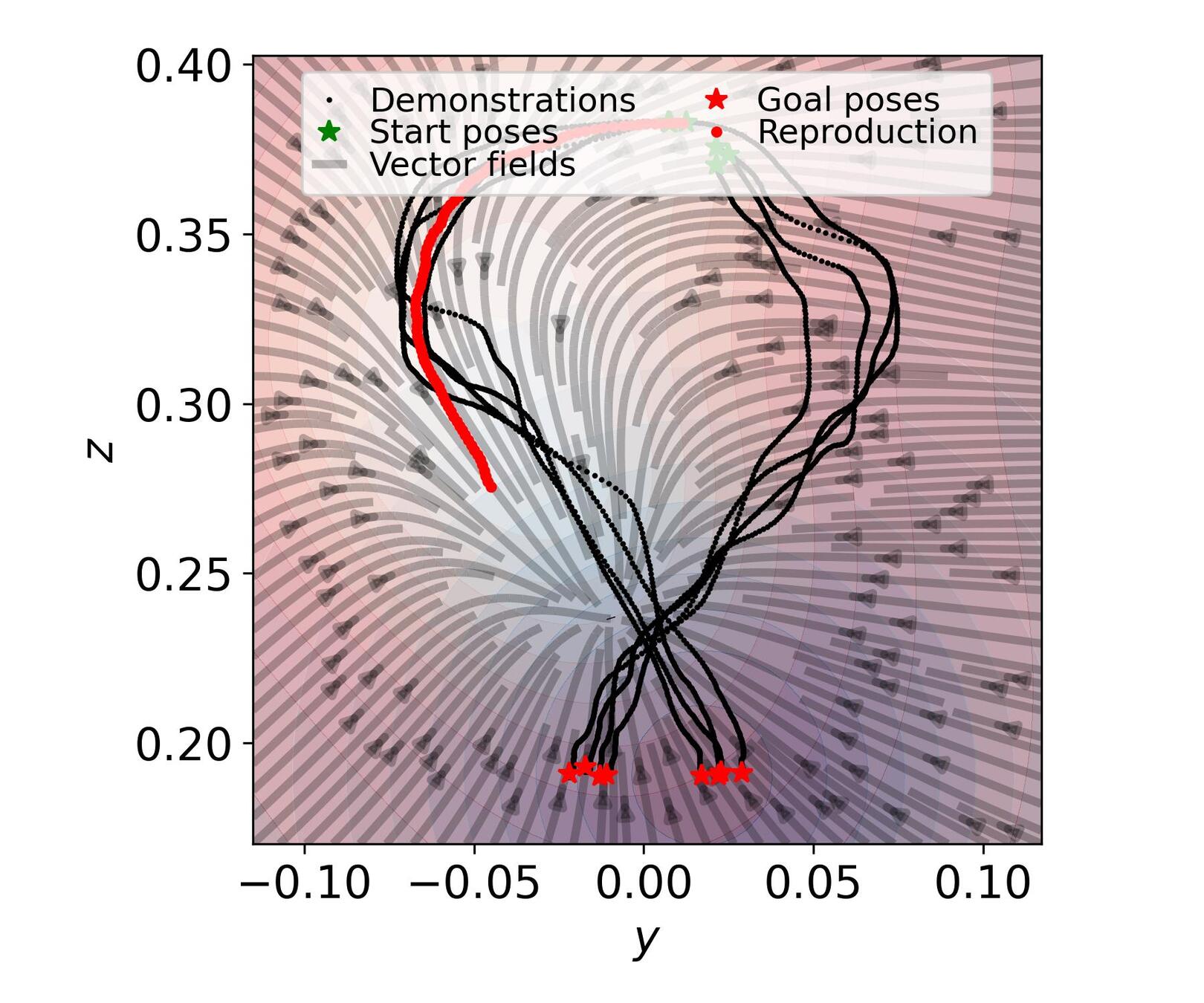}
		\caption{Vector fields at $t=5s$}
		\label{fig:fish_reproduced_100}
	\end{subfigure}
	\begin{subfigure}[b]{0.2355\textwidth}
		\centering
		\includegraphics[trim={0 0 0 0.7cm},clip, width=\linewidth]{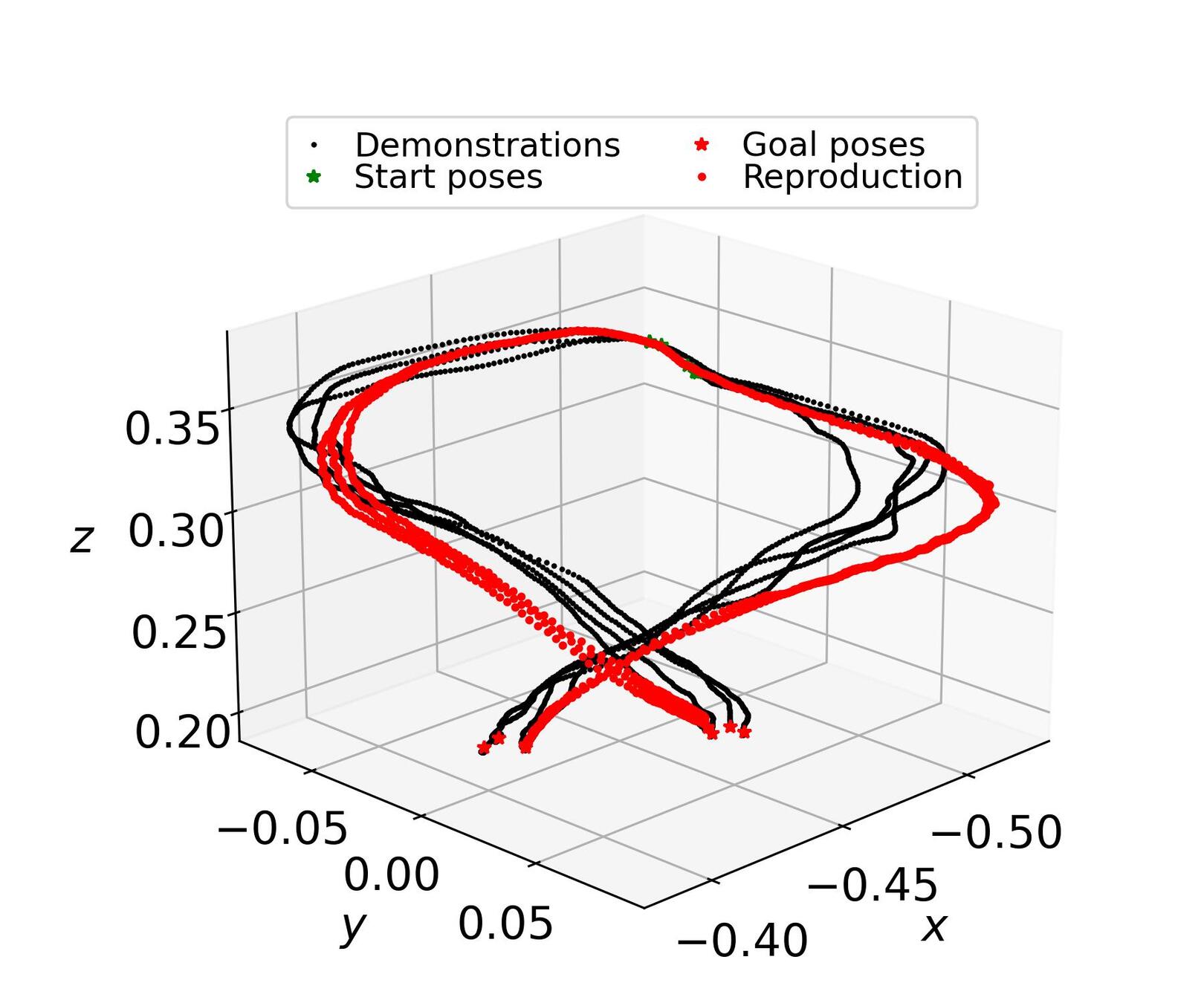}
		\caption{Reproduced trajectories}
		\label{fig:fish_reproductions}
	\end{subfigure}
	\begin{subfigure}[b]{0.2355\textwidth}
		\centering
		\includegraphics[trim={0 0.45cm 0 0.25cm},clip, width=\linewidth]{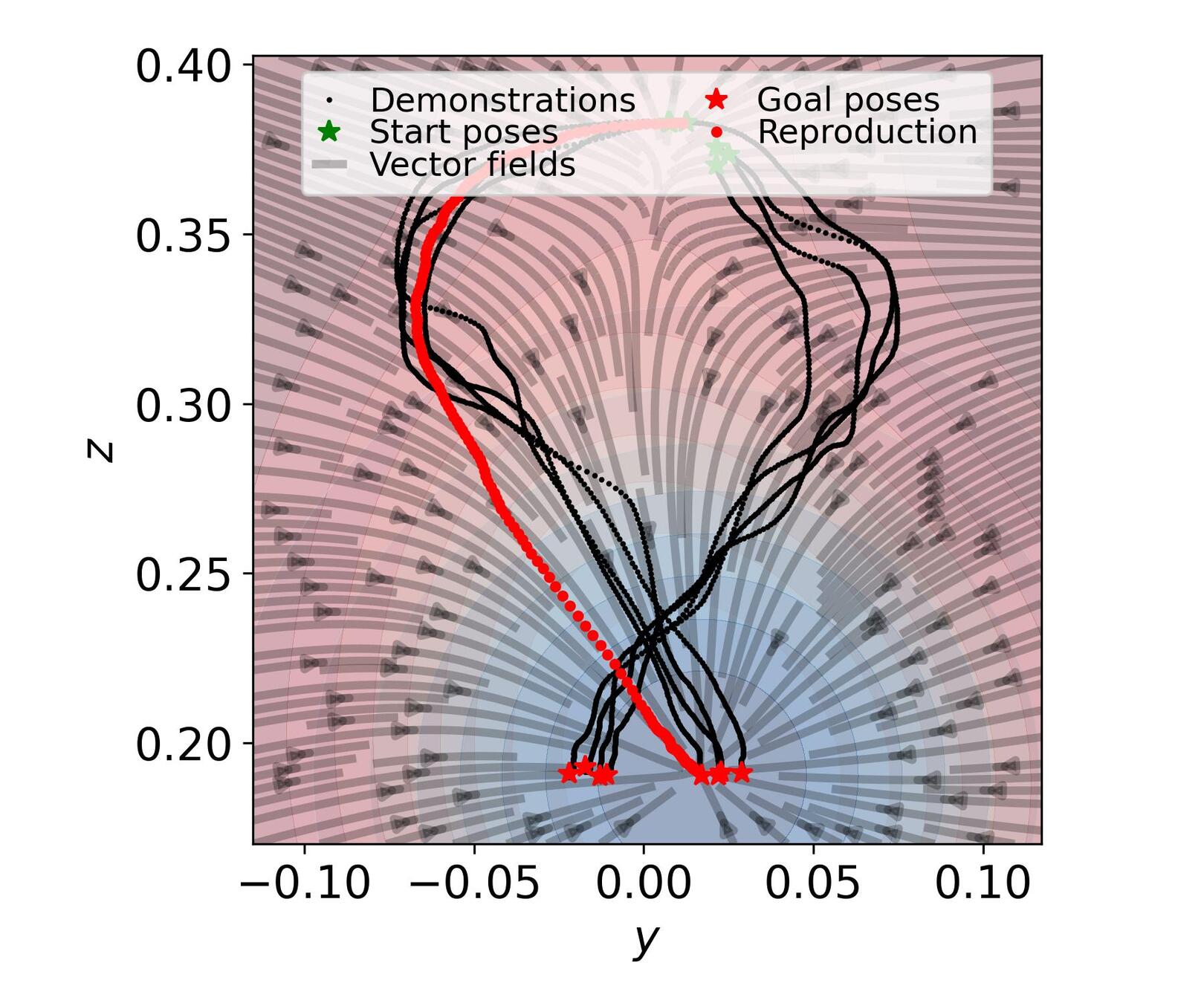}
		\caption{Vector fields at $t=12.5s$}
		\label{fig:fish_reproduced_250}
	\end{subfigure}
	\caption{\label{fig:fish}Demonstrations and reproductions of the manipulation of deformable food items experiment:
		(a) clarifies the definition of the task-parameters,
		(c) displays the reproduced trajectories of each four trials for both fish fillet configurations when starting at the demonstrated start poses,
		(e) shows the demonstrated and reproduced task-parameters from the quantitative evaluation (lower part of \Cref{table:real_robot_experiments}) and
		(b), (d), (f) exemplify the vector fields and the reproduced trajectory of a single trial at $t=2.5s$, $t=5s$ and $t=12.5s$ respectively when the fish fillet hangs on the right.}
\end{figure}

In the third experiment, we demonstrate our approach's adaptation to different placing strategies of a deformable object, in this case a fish fillet.
Using a similar robot and gripper as in \Cref{sec:regrasp}, but with different fingertips in order to facilitate the grasp,
we record the placing of the fish on a tray via kinesthetic teaching, as displayed in \Cref{fig:kinesthetic_teaching}.
Depending on the grasp configuration, the placing is performed differently, by approaching the tray from the side where the fillet hangs.
We provide four demonstrations for each side.
To capture the in-gripper fish geometry, we fine-tune YOLOv7~\cite{Wang2023} on the described use-case to obtain two bounding boxes, one for each side of the gripper.
We define the task-parameters as the fish's overlap on each side of the gripper, i.e.~$\bm{c} = \left[d_{\mathrm{left}} \> d_{\mathrm{right}}\right]^\top$, where $d_i = \max\limits_{k}{\left\|  \left[ u_{k,i} \> v_{k,i} \right]^\top - \left[ u_{ee} \> v_{ee} \right]^\top \right\| }$
is the maximum distance between the end-effector pose $\left[ u_{ee} \> v_{ee} \right]^\top$, transformed to image coordinates, and the $k \in [0,4]$ corners of the respective bounding box $\left[ u_{k,i} \> v_{k,i} \right]^\top$.\footnote{As this work focuses on the context-adaptive policy framework, the deformable object detection is implemented in this simplified, but sufficient form.
	However, more complex or 3D-based detectors would also be conceivable if the task-parameter dimension remains treatable by GPR (see \ref{sec:discussion}).}
\Cref{fig:task-param-def} clarifies the definition of $\bm{c}$ and \Cref{fig:fish_dist} shows the demonstrated task-parameters together with the reproduced ones.
Including robot position and orientation, the input dimension is $\mathcal{I} = 8$.

We perform an identical quantitative evaluation as in \Cref{sec:regrasp}, where the context-adaptive policy is used to compute the end-effector velocity $\{ \dot{\bm{x}}_d, \dot{\bm{\phi}}_d \}$ in a 20~Hz loop, hence $\mathcal{O} = 6$.
The results are shown in \Cref{table:real_robot_experiments} and the reproduced task-parameters are illustrated in \Cref{fig:fish_dist}.
Moreover, \Cref{fig:fish_reproductions} displays reproduced trajectories for both fish fillet configurations when starting at the demonstrated start poses
and \Cref{fig:fish_reproduced_50}, \Cref{fig:fish_reproduced_100} and \Cref{fig:fish_reproduced_250} exemplify the vector fields and the reproduced trajectory of a single trial at different times in case the fish fillet hangs on the right.

\subsection{Ablation analysis of the stabilizing policy's orientation extension}\label{sec:orientation_ablation_analysis}
The previous sections aimed to evaluate the effect of the proposed framework compared to other non-MoE baselines and different combinations of the individual policies.
To isolate the contribution of the stabilizing policy's orientation component (Contribution 3, see \Cref{sec:introduction}),
we perform an ablation where we remove it from \Cref{eq:stabilizing_policy}.
Specifically, we discard the orientation term and compute the gradient with respect to the current position $\bm{x}_*$ using only $v_{\bm{c}, \bm{x}}$, i.e.,
\begin{equation}
	\label{eq:orientation_ablation}
	\bm{\mu}_{\mathrm{sp}} = - K_{{\mathrm{sp}, \bm{x}}} \frac{\nabla_{\!\bm{x}_*} v_{\bm{c}, \bm{x}}}{\max{(\|  \nabla_{\!\bm{x}_*} v_{\bm{c}, \bm{x}} \| , \varepsilon)}} v_{\bm{c}, \bm{x}}.
\end{equation}

\begin{figure}
	\centering
	\begin{subfigure}[b]{1.0\columnwidth}
		\centering
		\includegraphics[trim={0 21cm 0 9cm},clip,width=0.32\columnwidth]{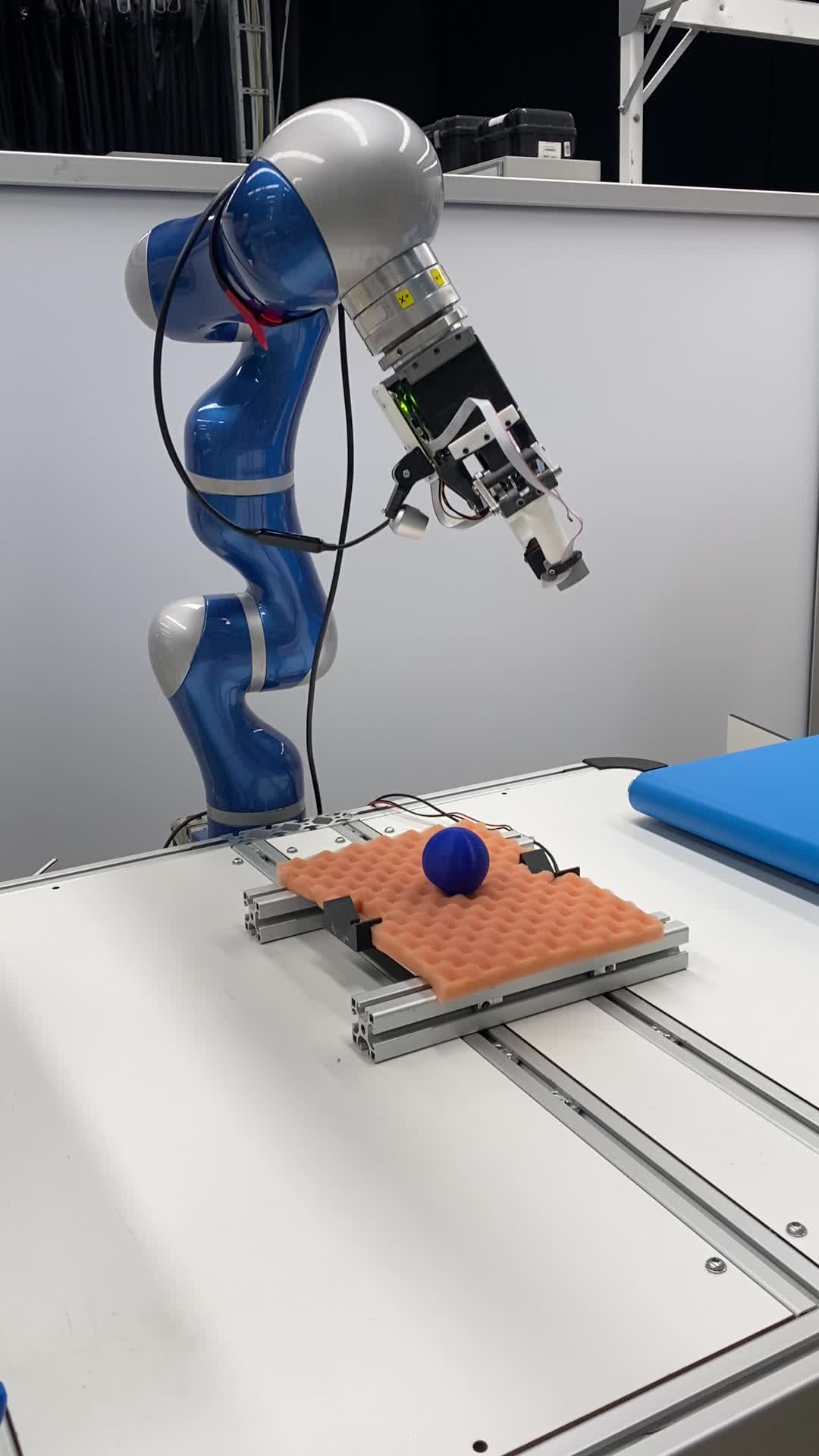}
		\includegraphics[trim={0 21cm 0 9cm},clip,width=0.32\columnwidth]{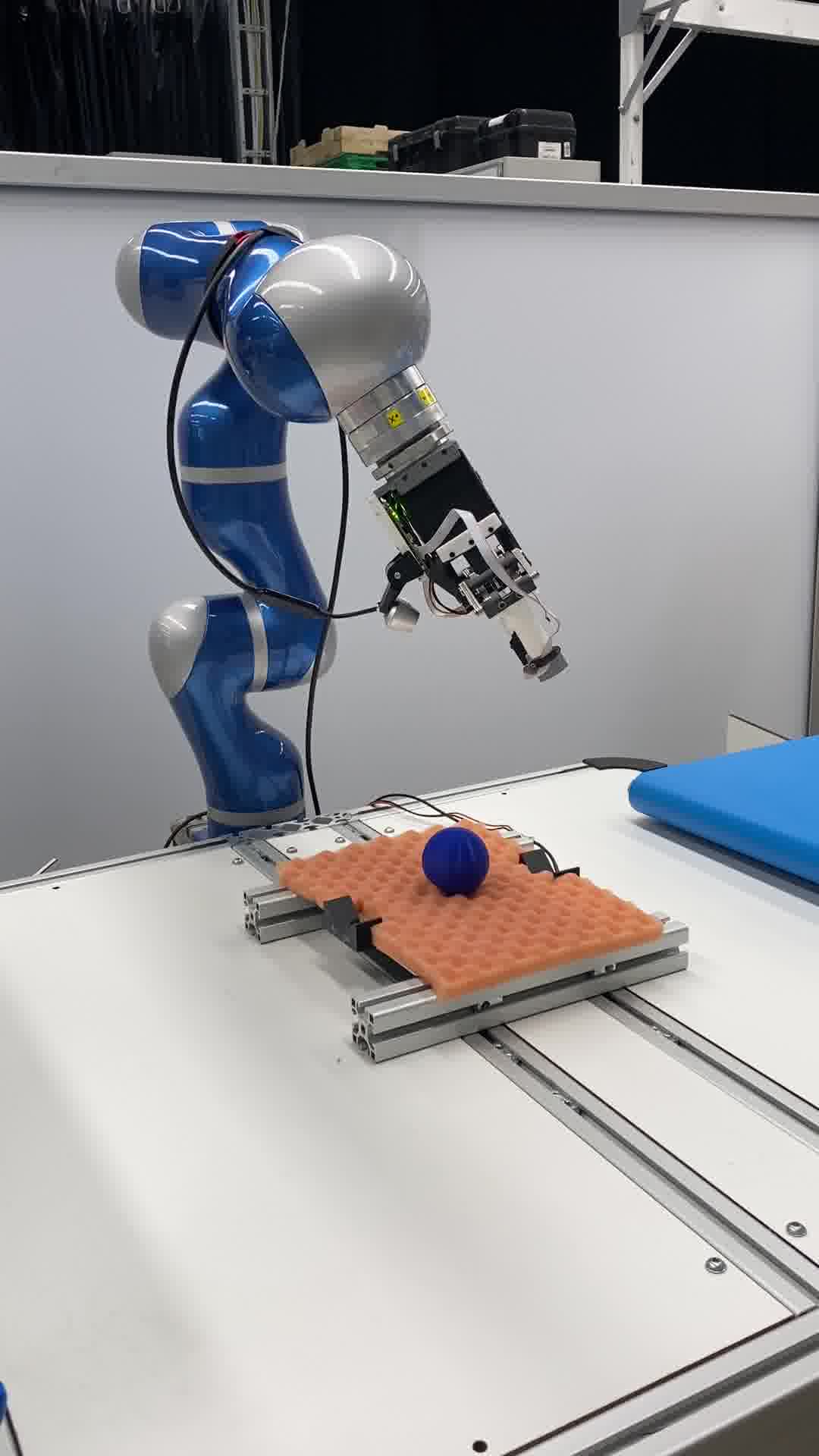}
		\includegraphics[trim={0 21cm 0 9cm},clip,width=0.32\columnwidth]{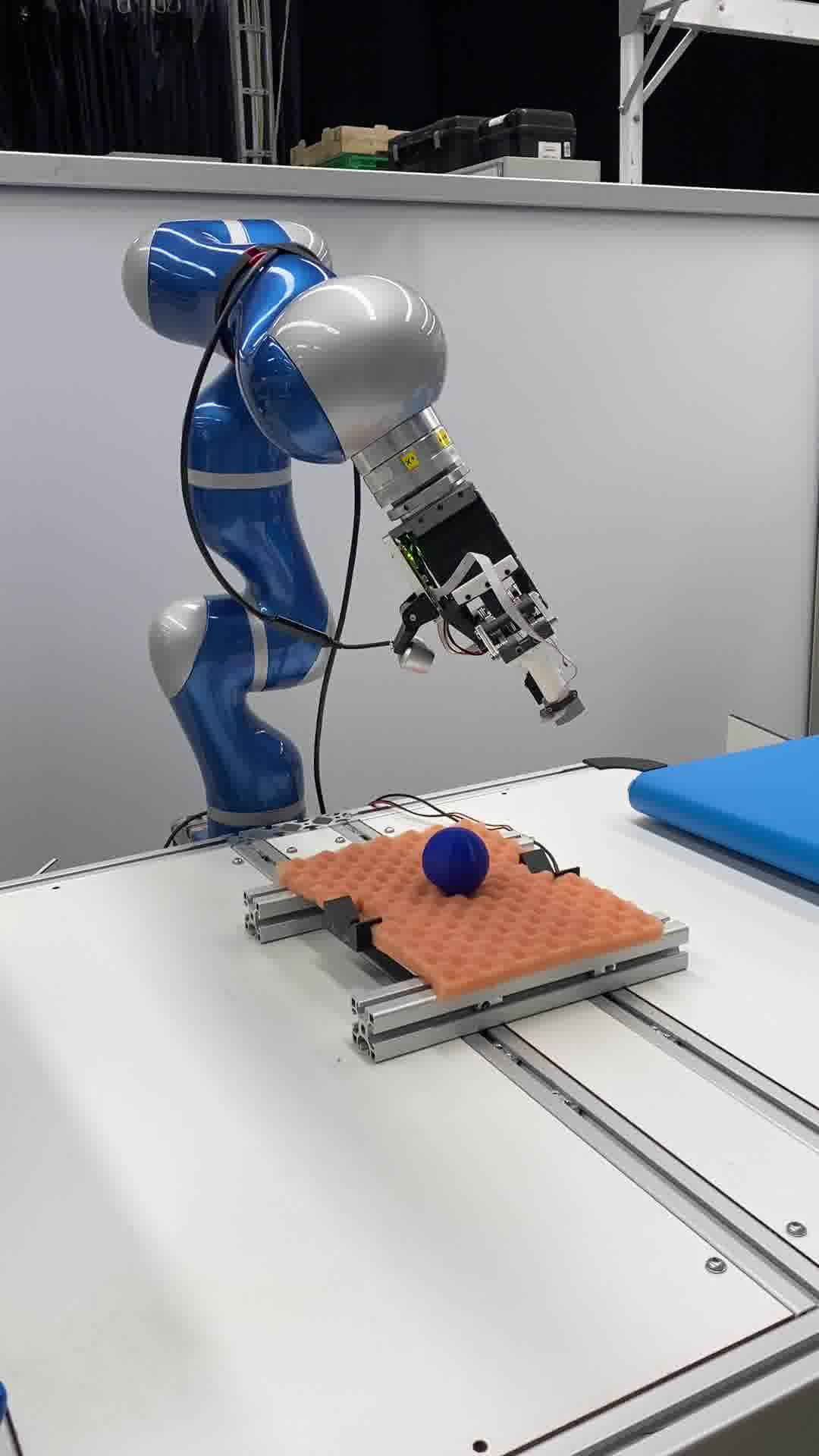}
		\includegraphics[trim={0 14cm 0 6cm},clip,width=0.32\columnwidth]{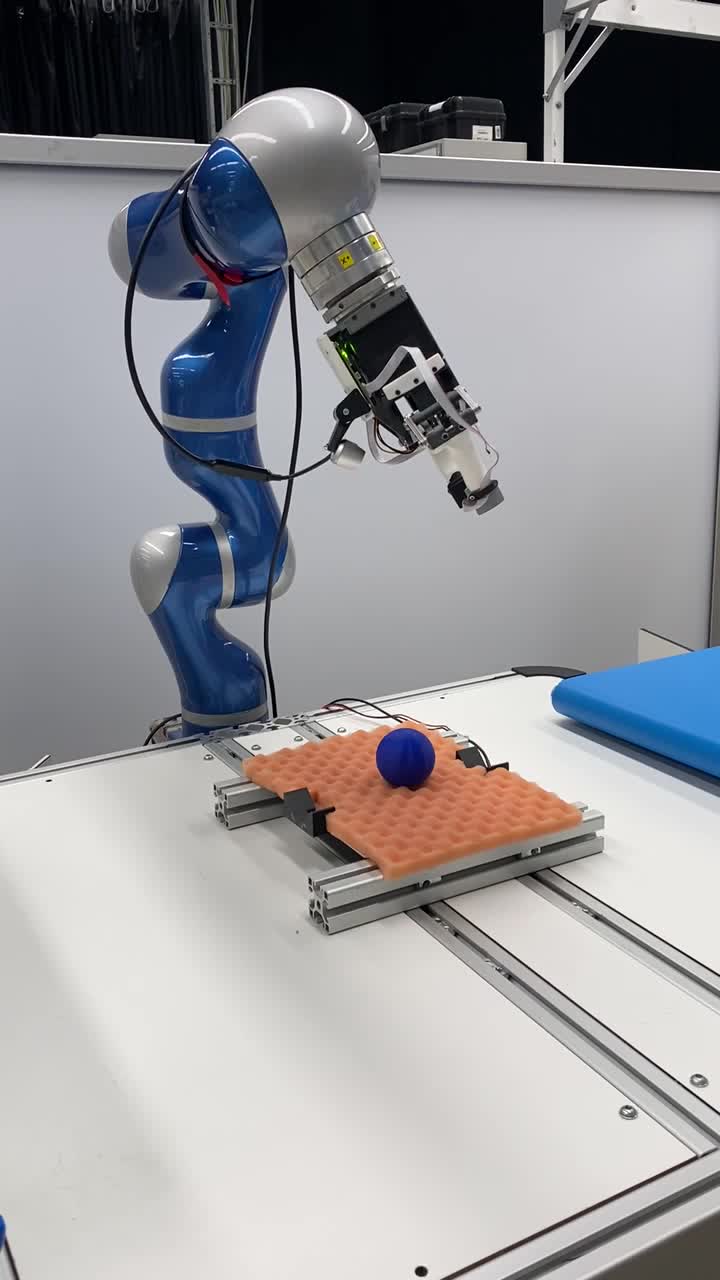}
		\includegraphics[trim={0 14cm 0 6cm},clip,width=0.32\columnwidth]{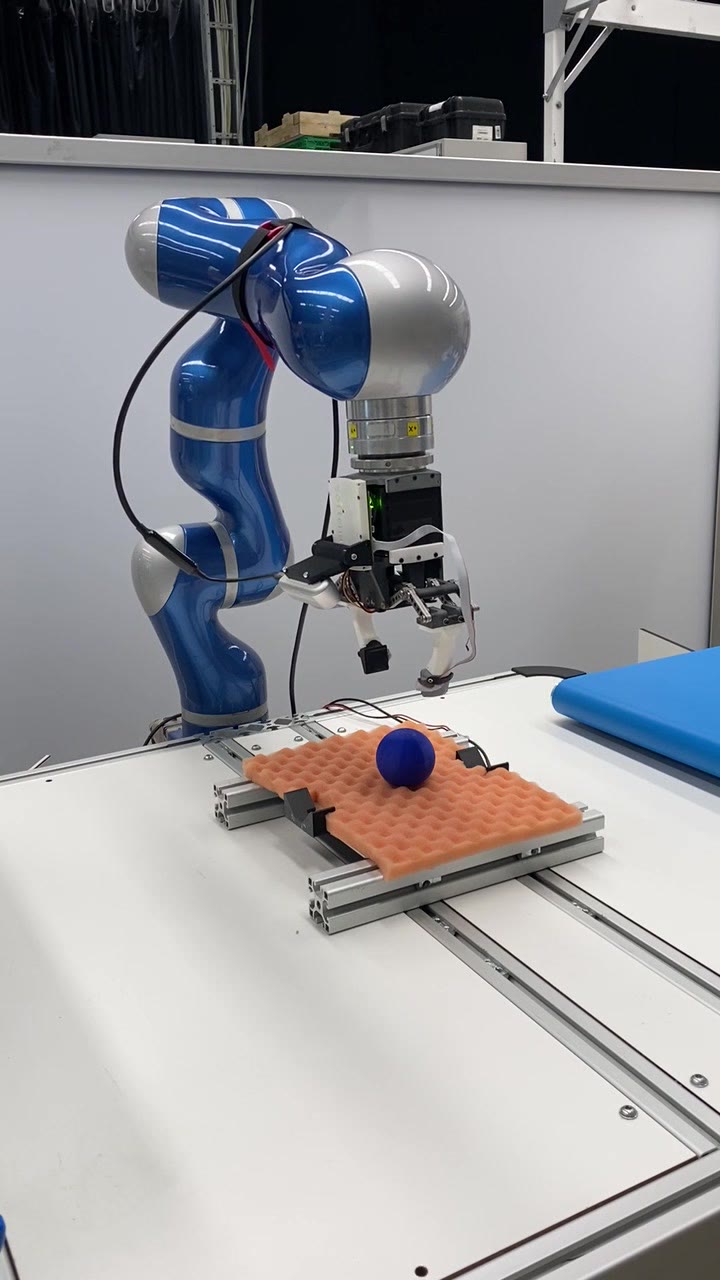}
		\includegraphics[trim={0 14cm 0 6cm},clip,width=0.32\columnwidth]{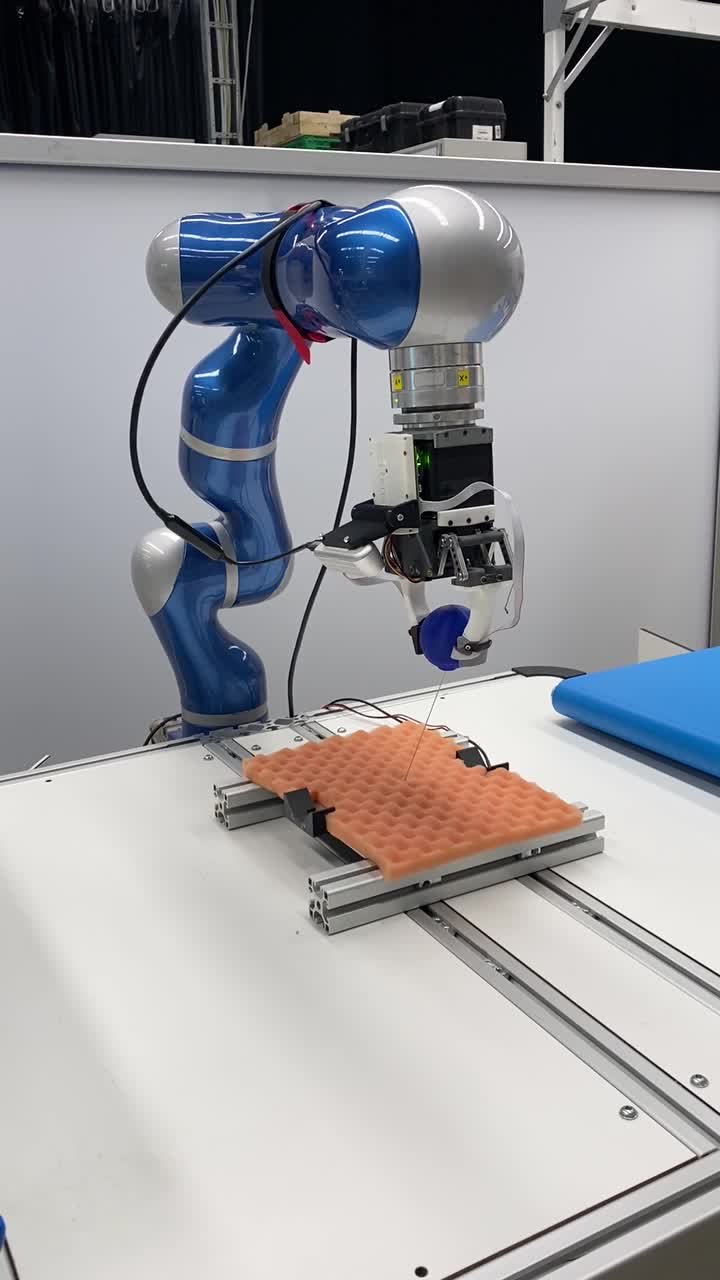}
	\end{subfigure}
	\caption{\label{fig:orientation_ablation}Behavioral differences between the ablated and full stabilizing policy across a representative execution sequence.
		The figure presents a time progression (from left to right) of the reproduction, with three snapshots captured at sequential time steps.
		The upper and lower rows correspond to the ablated stabilizing policy (\Cref{eq:orientation_ablation}) and the full framework (\Cref{eq:stabilizing_policy}), respectively.
		Quantitative performance metrics for this sequence are reported in \Cref{table:orientation_ablation_analysis}.}
\end{figure}

For this ablation, we use the experimental setup of \Cref{sec:regrasp},
but sample the starting poses from an extended range to explicitly examine the effects of the stabilizing policy's orientation component,
which has an impact only in out-of-distribution orientations.
We sample the starting poses around the demonstrated goal poses by perturbing translations uniformly within a sphere of radius 0.15~m
and rotations uniformly about arbitrary axes with a maximum angle of 45$^{\circ}$.
The results are reported in \Cref{table:orientation_ablation_analysis}.
The rows in this table denote the stabilizing policy used:
\Cref{eq:orientation_ablation} corresponds to the ablated policy, which omits the orientation component,
while \Cref{eq:stabilizing_policy} corresponds to our full framework, which includes it.
The behavioral differences arising from this ablation are illustrated in \Cref{fig:orientation_ablation},
where the upper and lower rows correspond to the ablated and full policy, respectively, consistent with \Cref{table:orientation_ablation_analysis}.
As shown, our full framework recovers from out-of-distribution orientations, whereas the ablated policy fails to do so.
The same execution sequence is presented in the accompanying video.

\begin{table}
	\caption{\label{table:orientation_ablation_analysis}Ablation analysis of the stabilizing policy's orientation extension on the experimental setup described in \Cref{sec:regrasp}.
		The rows denote the equation number used for the stabilizing policy:
		(\ref{eq:orientation_ablation}) corresponds to the ablated stabilizing policy without the orientation component,
		and (\ref{eq:stabilizing_policy}) corresponds to our framework with the orientation component.
		The evaluated metrics (average success $S$, iterations $I$, distance $D$, collisions \textit{Coll.}) are reported as the mean and standard deviation computed over 20 trials.}
	\begin{center}
		\begin{tabular}{lcccc}
			\toprule
			Method                         & $S$ [\%]           & $I$                 & $D$                & \textit{Coll.} \\
			\midrule
			\Cref{eq:orientation_ablation} & $5.0\pm21.8$       & $969.5\pm125.3$     & $0.42\pm0.22$      & $1$            \\
			\Cref{eq:stabilizing_policy}   & $\bm{100.0\pm0.0}$ & $\bm{320.1\pm29.6}$ & $\bm{0.03\pm0.02}$ & $\bm{0}$       \\[1pt]
			\bottomrule
		\end{tabular}
	\end{center}
\end{table}

\subsection{System extension --- Object-centric grasping}\label{sec:dynamic_grasping}
\begin{figure}
	\centering
	\begin{subfigure}[b]{1.0\columnwidth}
		\centering
		\includegraphics[trim={0 6cm 20cm 5cm},clip,width=0.49\columnwidth]{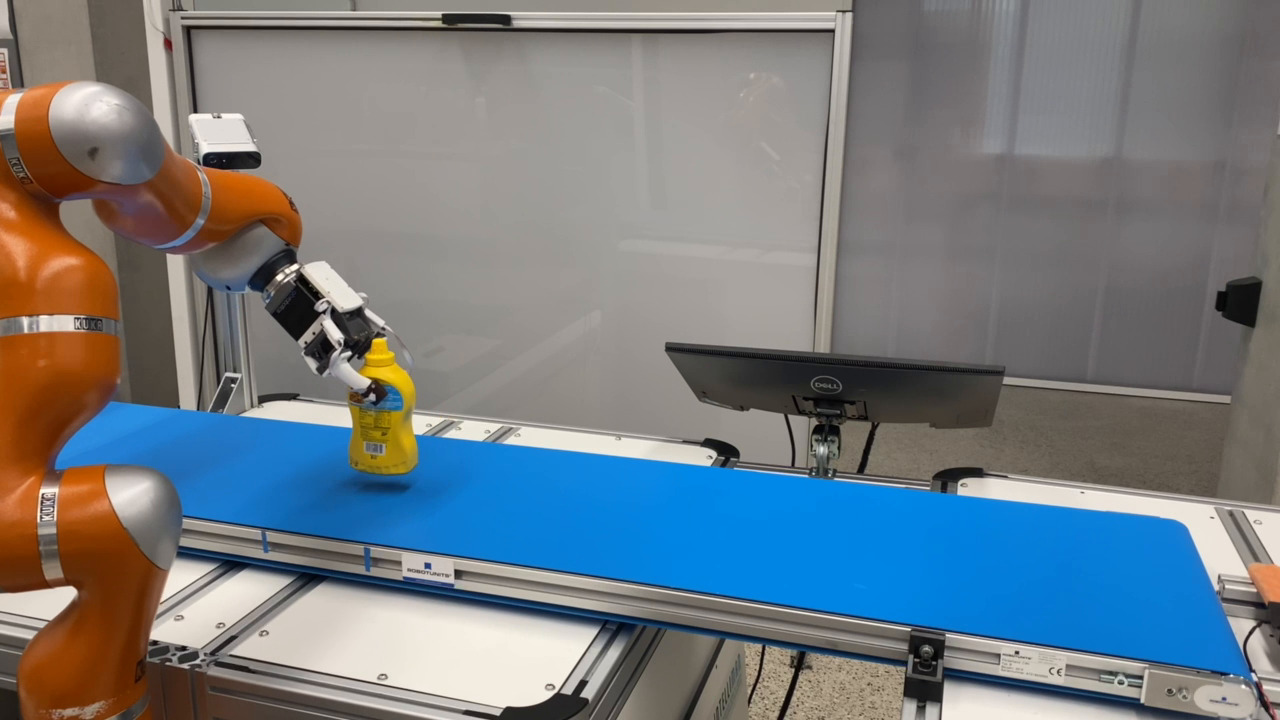}
		\includegraphics[trim={0 6cm 20cm 5cm},clip,width=0.49\columnwidth]{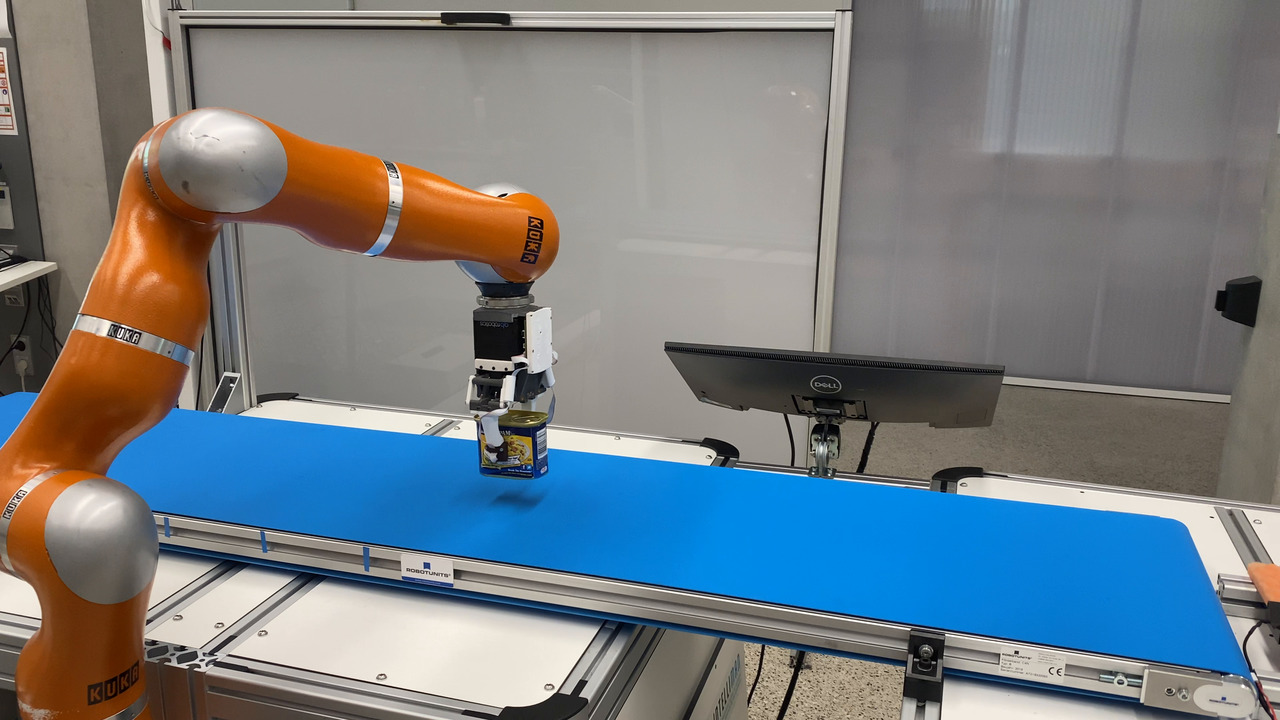}
		\includegraphics[trim={0 6cm 20cm 5cm},clip,width=0.49\columnwidth]{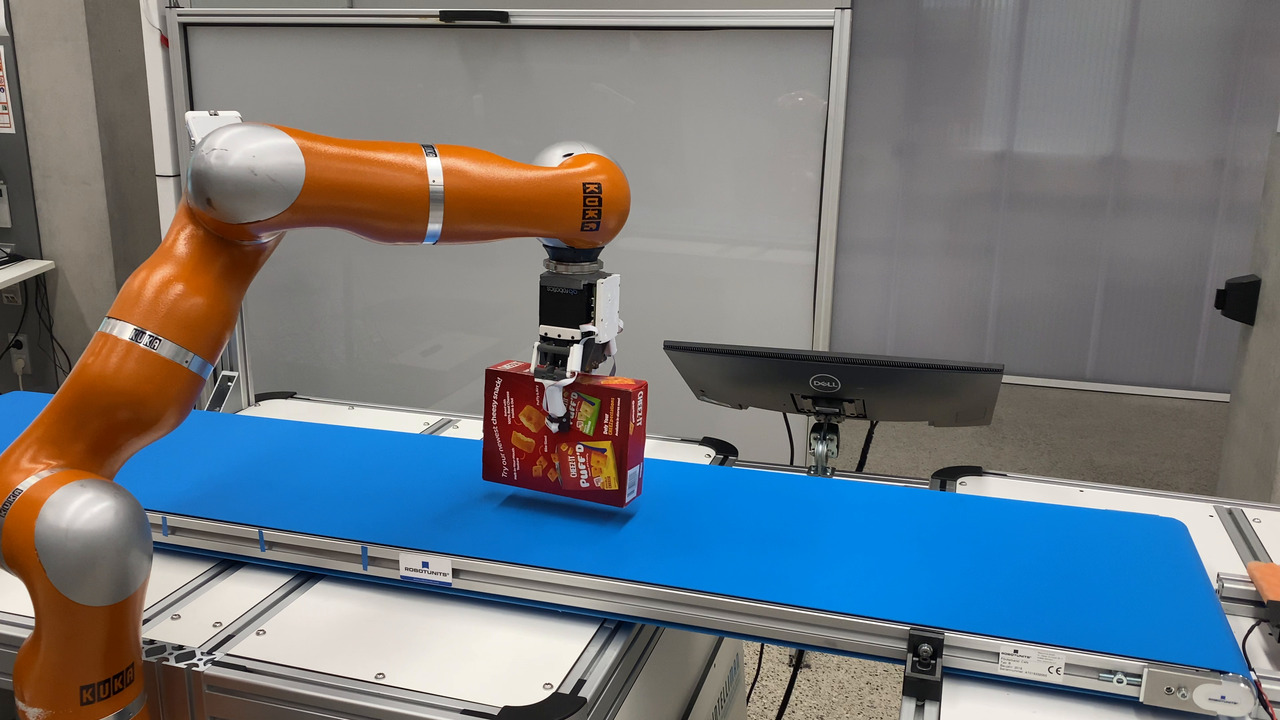}
	\end{subfigure}
	\caption{\label{fig:dynamic_grasping_objects}\textit{Placing} phase of the object-centric grasping experiment: The robot places the grasped objects (``Mustard Bottle'', ``Potted Meat Can'' and ``Cracker Box'') outside the workspace on the conveyor belt.}
\end{figure}

In the final experiment, we apply our approach to grasp three selected YCBV objects\footnote{Available at \url{https://rse-lab.cs.washington.edu/projects/posecnn/}.}  from a conveyor belt.
We record three demonstrations per object via kinesthetic teaching with static conveyor belt consisting of a \textit{grasping}, \textit{placing} and \textit{resting} phase.
During the \textit{grasping} phase the object is approached and grasped within a pre-defined region of the workspace,
before being placed again on the conveyor belt outside the pre-defined region during the \textit{placing} phase.
After placing the object, the robot moves to a defined position in which it waits (\textit{resting} phase) until the object moves back into the pre-defined region.
The \textit{Placing} phase, is exemplarily illustrated in \Cref{fig:dynamic_grasping_objects} for each object.

To robustly accomplish the task, three extensions to the previously introduced framework were made.
First, the continuous demonstrations were divided by phase into three datasets.
Two separate policies were trained: one on the \textit{grasping} phase dataset transformed to object coordinate system, and another on the remaining two datasets in world coordinate system.
Depending on the current phase, the respective policy is selected,
allowing the object to be grasped at any location within the workspace while the \textit{placing} and \textit{resting} poses remain static.
Moreover, a non-zero mean is used for the gripper joint velocity within the \textit{grasping} phase so that the gripper re-opens while approaching the object.
Finally, a Mat\'ern kernel~\cite{Rasmussen2006} is used for $k(\bm{x}_*,\bm{x}_{g})$ and $k(\bm{\phi}_*,\bm{\phi}_{g})$ in \Cref{eq:goal_attractor},
producing higher end-effector velocities near the goal poses and thus allowing higher object velocities.

The object position is tracked at 20~Hz using a combination of learning and model-based computer vision methods, including YOLOv7~\cite{Wang2023}, a 6D object pose estimation~\cite{Ulmer2023} and a 3D Object
Tracking~\cite{Stoiber2023}.
We define the task-parameters as $\bm{c} = \left[k_{\mathrm{obj}} \> j_{\mathrm{phase}} \right]^\top$, where $k_{\mathrm{obj}} \in \{2, 5, 9\}$ represents ``Cracker Box'', ``Mustard Bottle'', and ``Potted Meat Can''.
Further, $j_{\mathrm{phase}} \in \{0, 1, 2\}$ represents the \textit{grasping}, \textit{placing} and \textit{resting} phase, that are active when the object is located within the pre-defined region, grasped or located outside the pre-defined region, respectively.
Thus, the input and output dimensions are $\mathcal{I} = 8$ and $\mathcal{O} = 7$.

\begin{figure}
	\centering
	\begin{subfigure}[b]{1.0\columnwidth}
		\centering
		\includegraphics[trim={0 7cm 20cm 4cm},clip,width=0.49\columnwidth]{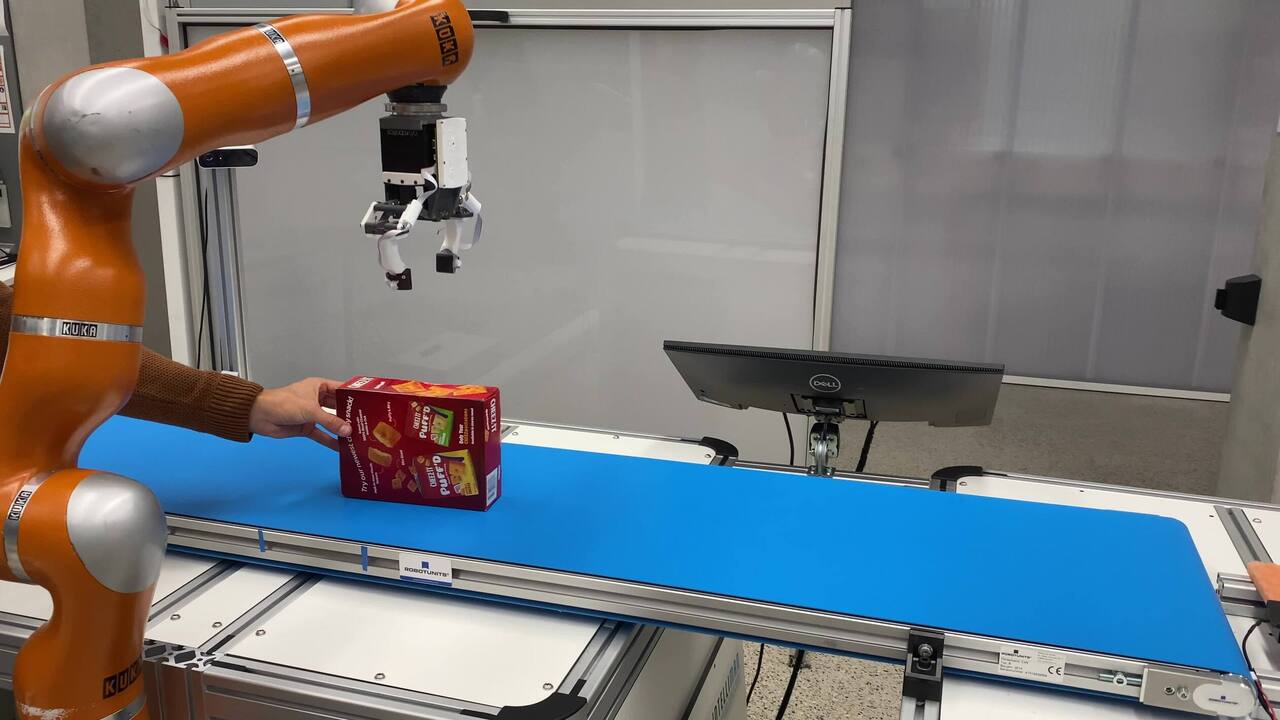}
		\includegraphics[trim={0 7cm 20cm 4cm},clip,width=0.49\columnwidth]{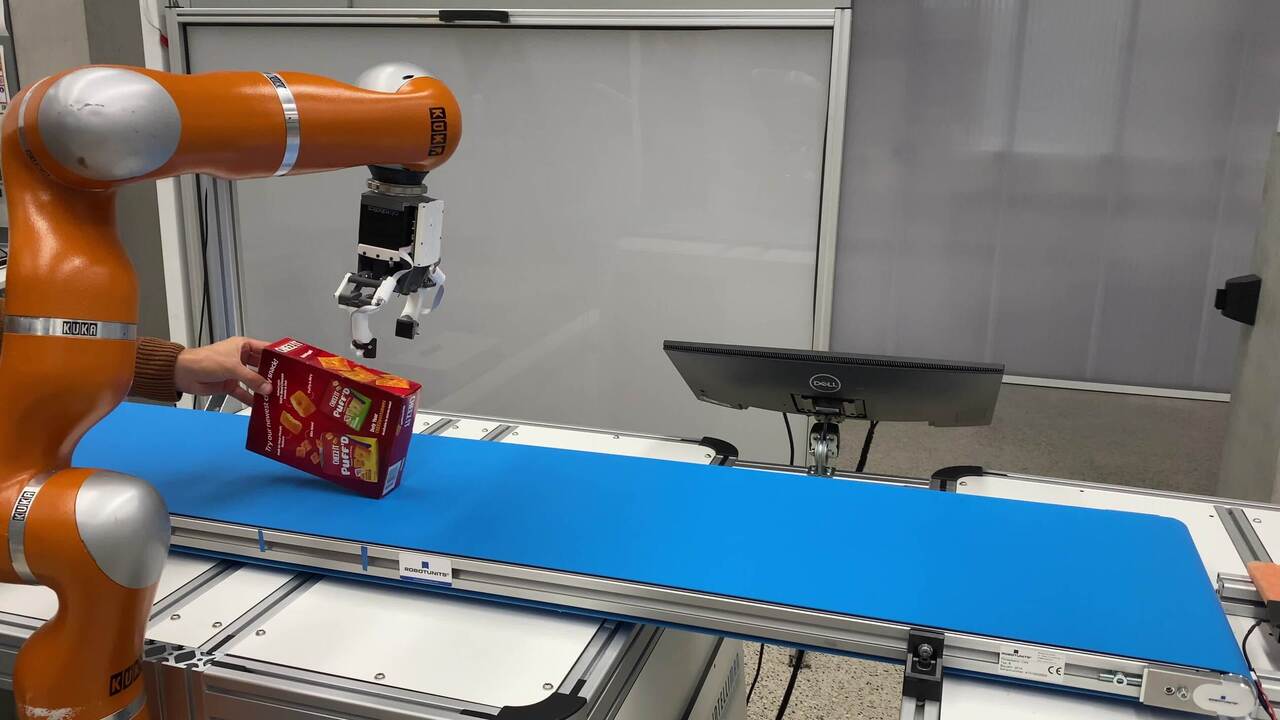}
		\includegraphics[trim={0 7cm 20cm 4cm},clip,width=0.49\columnwidth]{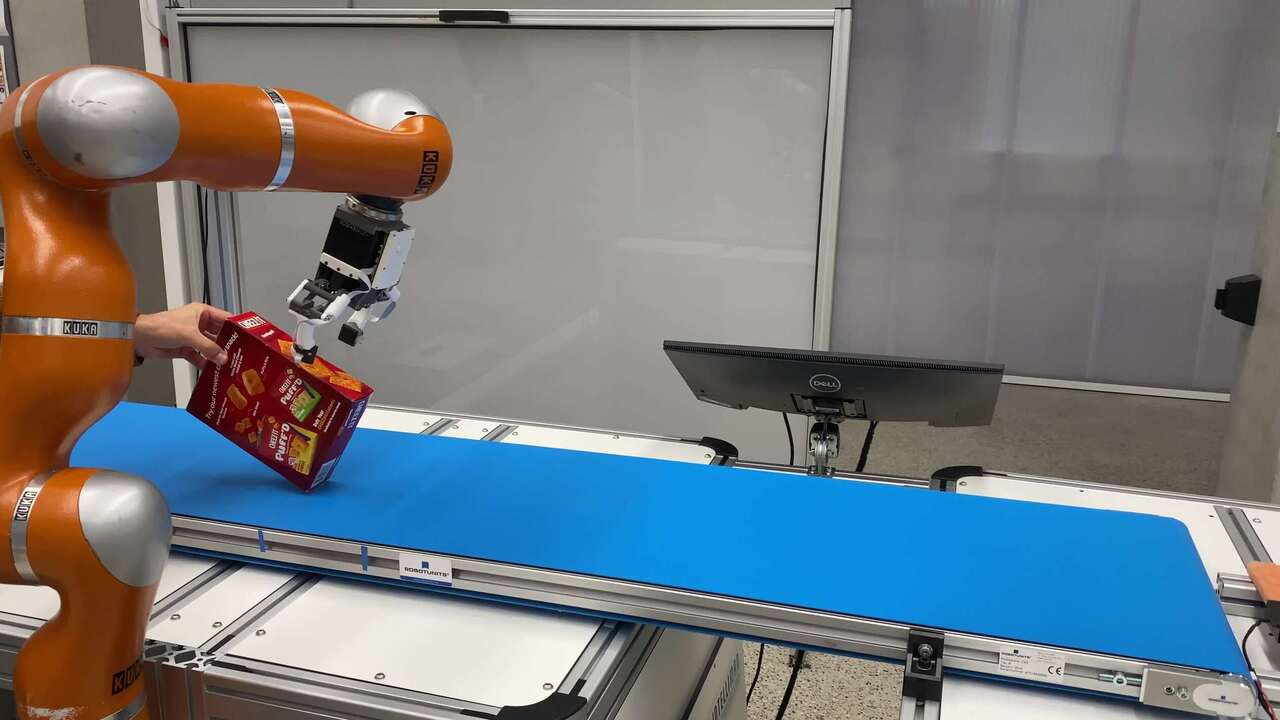}
		\includegraphics[trim={0 7cm 20cm 4cm},clip,width=0.49\columnwidth]{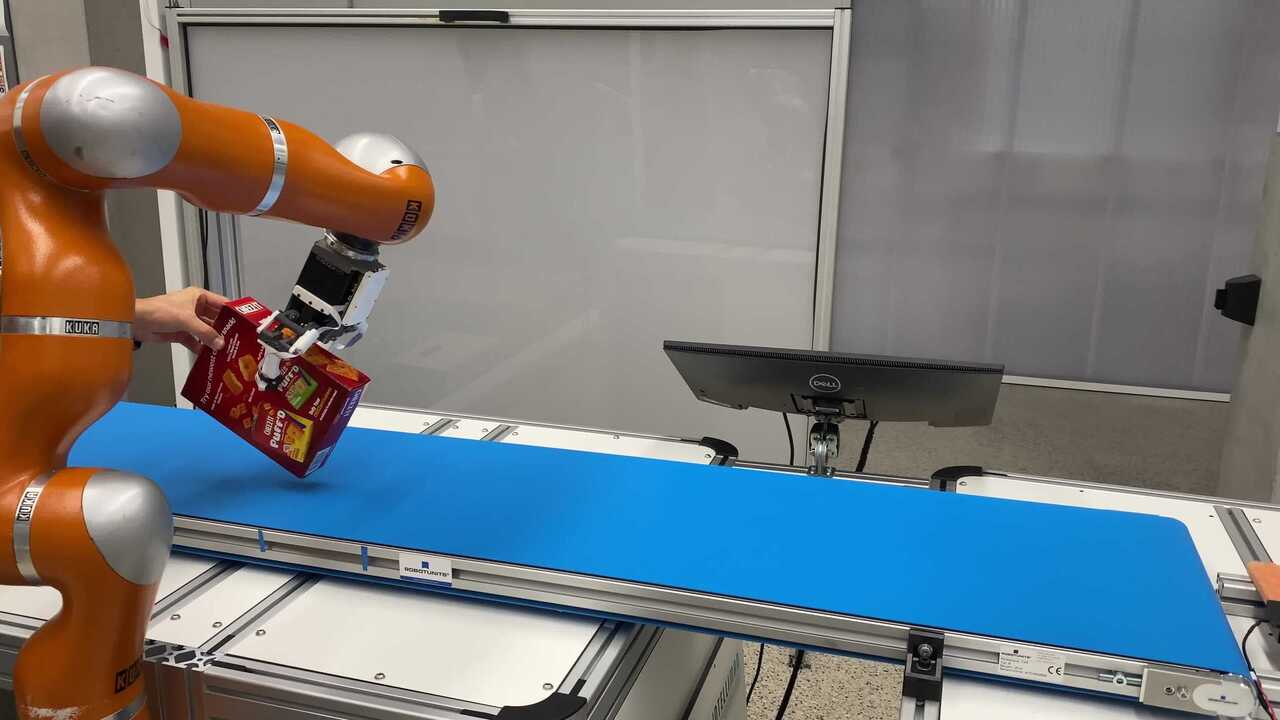}
	\end{subfigure}
	\caption{\label{fig:dynamic_grasping_handover}Possible extensions of our approach include online adaptation of robot behavior to objects poses, e.g.,~during handovers:
		The robot tracks the dynamically changing position of the object (``Cracker Box'') in the human hand, continuously adjusting its trajectory in real time to successfully complete the grasp.}
\end{figure}

We demonstrate our extended framework's capability to grasp the three objects on both static and moving conveyor belt, with varying initial objects and robot poses.
The experimental setup and evaluated metrics match those of the previous experiments.
The results of this experiment are presented in \Cref{table:dynamic_grasping_experiments}, split into three sub-experiments:
\begin{itemize}
	\item Object pose variation: We evaluate our approach's generalization to unseen object configurations by varying both object position and orientation
	      while keeping the robot's starting pose fixed at the resting configuration.
	      The object is placed at the six positions defined by the intersections of $x \in \{-0.15, 0.0, 0.15\}$ m and $y \in \{0.45, 0.60\}$ m,
	      combined with three orientations around the $z$-axis relative to the demonstrated configuration ($\{-45^\circ, 0^\circ, 45^\circ\}$),
	      resulting in 18 distinct object configurations per object.
	\item Robot pose variation: We assess the robustness of our approach to varying initial robot configurations while the object remains static at a single pose ($[x, y] = [0.15, 0.525]$ m with the demonstrated orientation).
	      The robot starting pose is uniformly sampled 15 times per object from the range of demonstrated poses.
	\item Dynamic grasping: We test our approach's capability to grasp objects on a moving conveyor belt under varying motion conditions.
	      The conveyor operates at three different speeds (0.01, 0.02, and 0.03~${\rm m/s}$),
	      with the object placed at two positions ($y \in \{0.45, 0.60\}$ m)
	      and three orientations around the $z$-axis relative to the demonstrated configuration ($\{-45^\circ, 0^\circ, 45^\circ\}$),
	      yielding 18 dynamic test conditions per object.
\end{itemize}
For visual reference, refer to the accompanying video,
where we also showcase how this extension can be used in the context of human-robot handovers (screenshots in \Cref{fig:dynamic_grasping_handover}).

\begin{table*}
	\caption{\label{table:dynamic_grasping_experiments}Quantitative evaluation (average success $S$, iterations $I$, distance $D$, collisions \textit{Coll.}) on the experiment described in \ref{sec:dynamic_grasping}}
	\begin{center}
		\begin{tabular}{lllcccc}
			\toprule
			\multicolumn{2}{L}{Exp.}                                   & Obj.                                       & $S$ [\%]        & $I$               & $D$               & \textit{Coll.}          \\
			\midrule
			\multicolumn{2}{L}{\multirow{3}{*}{Object pose variation}} & Cracker Box                                & $100.0 \pm 0.0$ & $387.4 \pm 75.6$  & $0.059 \pm 0.015$ & $0$                     \\
			                                                           &                                            & Mustard Bottle  & $94.4 \pm 22.9$   & $376.8 \pm 70.8$  & $0.159 \pm 0.077$ & $1$ \\
			                                                           &                                            & Potted Meat Can & $100.0 \pm 0.0$   & $386.3 \pm 98.2$  & $0.113 \pm 0.069$ & $0$ \\
			\midrule
			\multicolumn{2}{L}{\multirow{3}{*}{Robot pose variation}}  & Cracker Box                                & $100.0 \pm 0.0$ & $309.9 \pm 54.8$  & $0.077 \pm 0.023$ & $0$                     \\
			                                                           &                                            & Mustard Bottle  & $93.3 \pm 24.9$   & $348.2 \pm 33.7$  & $0.101 \pm 0.014$ & $1$ \\
			                                                           &                                            & Potted Meat Can & $93.3 \pm 24.9$   & $311.3 \pm 32.5$  & $0.121 \pm 0.076$ & $1$ \\
			\midrule
			\multirow{10}{*}{\rotatebox{90}{Dynamic grasping}}         &
			\multirow{3}{*}{Speed: $0.01 \frac{m}{s}$}
			                                                           & Cracker Box                                & $100.0 \pm 0.0$ & $392.7 \pm 103.3$ & $0.072 \pm 0.020$ & $0$                     \\
			                                                           &                                            & Mustard Bottle  & $83.3 \pm 37.3$   & $353.0 \pm 63.6$  & $0.220 \pm 0.162$ & $1$ \\
			                                                           &                                            & Potted Meat Can & $100.0 \pm 0.0$   & $333.2 \pm 63.3$  & $0.108 \pm 0.041$ & $0$ \\[1pt]
			\cdashline{2-7}\noalign{\vskip 2pt}
			                                                           & \multirow{3}{*}{Speed: $0.02 \frac{m}{s}$}
			                                                           & Cracker Box                                & $83.3 \pm 37.3$ & $461.4 \pm 163.8$ & $0.093 \pm 0.091$ & $1$                     \\
			                                                           &                                            & Mustard Bottle  & $83.3 \pm 37.3$   & $387.4 \pm 134.4$ & $0.202 \pm 0.128$ & $1$ \\
			                                                           &                                            & Potted Meat Can & $100.0 \pm 0.0$   & $422.0 \pm 84.6$  & $0.089 \pm 0.022$ & $0$ \\[1pt]
			\cdashline{2-7}\noalign{\vskip 2pt}
			                                                           & \multirow{3}{*}{Speed: $0.03 \frac{m}{s}$}
			                                                           & Cracker Box                                & $100.0 \pm 0.0$ & $401.2 \pm 121.2$ & $0.062 \pm 0.018$ & $0$                     \\
			                                                           &                                            & Mustard Bottle  & $50.0 \pm 50.0$   & $397.0 \pm 40.0$  & $0.249 \pm 0.150$ & $3$ \\
			                                                           &                                            & Potted Meat Can & $66.7 \pm 47.1$   & $559.5 \pm 44.4$  & $0.138 \pm 0.058$ & $2$ \\
			\bottomrule
		\end{tabular}
	\end{center}
\end{table*}

\section{Discussion}\label{sec:discussion}
In this section, we discuss the results of the conducted experiments, analyze failure cases, and outline the limitations of our approach.

\subsection{Baseline performance}

The low success rates (S) of both GPR (standalone LfD policy, \Cref{sec:lfd_policy}) and the considered Diffusion Policy (DP) baseline
(\Cref{table:baseline_experiments,table:real_robot_experiments}) underscore the challenge of learning reliable and robust behavior from limited demonstrations.
The results demonstrate that these policies, when deployed without additional tuning or architectural adaptations, often lack the robustness required in small-data regimes.\footnote{This observation specifically pertains to the considered, unmodified baseline configurations with limited data, directly motivating the structural enhancements introduced in our approach.}
In our experiments, the observed low success rates can be primarily attributed to either (1) insufficient state-space coverage, due to the small number of demonstrations provided,\footnote{Note that a small-data regime was deliberately chosen to assess the robustness of the proposed method under limited data availability.
	However, the large parameter count in DP contrasts sharply with the limited number of demonstrations, making effective training particularly challenging.}  or (2) a lack of empirical convergence within the tested horizon.

The insufficient state-space coverage causes both baselines, to frequently operate outside the support of the training distribution when exposed to novel initial states or deviations from the demonstrated trajectory.
This often results in unpredictable, potentially safety-critical, behavior,
as reflected in the metrics capturing safety and trajectory fidelity, namely the number of collisions (C) and the distance to demonstrated trajectories (D).
Furthermore, as visible in the accompanying video, both baselines fail to halt upon reaching the goal state, instead continuing to drift slowly without achieving zero velocity.
This lack of empirical convergence within the tested horizon is reflected in the number of iterations (I) and the timeouts (reaching the iteration limit) indirectly depicted therein.
Both of these failure cases are highly evident in the real-world experiments (see \Cref{table:real_robot_experiments}),
where DP resulted in 39 collisions and 1 timeout, while GPR caused 23 collisions and 17 timeouts, across 40 trials per policy.
Since a zero-mean GP is used, causing GPR to predict near-zero velocities in regions far from the training data,
a higher number of timeouts and a lower number of collisions are observed for GPR compared to DP.

\subsection{Proposed framework performance}

In contrast, the combination of multiple \textit{expert} policies informed by the epistemic uncertainty (proposed framework, \Cref{sec:policy_fusion}) is able to achieve high success and low collision rates
while maintaining proximity to the demonstrated data distribution.
Accordingly, \Cref{table:baseline_experiments} shows that, in scenarios without external task-parameter inputs, our approach performs comparably to the established DS-based method SEDS in terms of success rate,\footnote{The few failures of SEDS are due to reaching the iteration limit.}
but superior in terms of adhering to demonstrations.
However, as discussed in \Cref{sec:related_work,sec:policy_fusion}, SEDS does not support modulation by external task parameters, a capability offered by our method,
making a direct comparison against SEDS on the real-robot experiments not possible.
We successfully show this modulation capability in three real-robot experiments (\Cref{sec:regrasp,sec:fish,sec:dynamic_grasping}),
where the context-adaptive policy dynamically adjusts its behavior in response to variations in both continuous and discrete task-dependent parameters during execution.
The results in \Cref{table:real_robot_experiments}, particularly the combined success rate of 97.5\% (the few failures are due to reaching the iteration limit) and 0 collisions,
suggest that our approach can be robustly deployed on real robots, providing important properties like intuitive and data-efficient skill transfer,
as well as improved adaptability to changing environments.

The experimental results also show that both supplementary policies, naturally emerging from the properties of the Gaussian Process framework with RBF kernels (as in \Cref{sec:policy_fusion}),
fulfill their intended purpose in the considered scenarios and that both are required to achieve robust reproduction behaviors,
as presented quantitatively in \Cref{table:baseline_experiments,table:real_robot_experiments} and visually in \Cref{fig:regrasp_wo_stab,fig:regrasp_wo_goal}.
Thus, removing the stabilizing or the goal attractor policy either causes the robot to drift away from the demonstrations or reduced empirical convergence within the tested horizon, respectively.
Complementing these findings, the ablation analysis, reported in \Cref{table:orientation_ablation_analysis}, demonstrates the benefit of incorporating orientation into the stabilizing policy.
The results show that the proposed formulation recovers from out-of-distribution orientations without the need for an additional recovery mechanism,
whereas removing the orientation component prevents such recovery.

Figures~\ref{fig:regrasp_forces} and~\ref{fig:fish_dist} further demonstrate that our approach robustly performs the tasks even when the task-dependent parameters encountered during reproduction (red) differ, in some cases substantially, from those observed in the demonstrations (blue).
In addition, we show in \Cref{table:dynamic_grasping_experiments} that our task-space formulation allows the context-adaptive policy to be represented locally with respect to the objects,
which, in combination with the real-time reactivity of our approach, enables the robot to dynamically adjust its behavior to continuously changing object poses while maintaining robustness.
This reactivity is demonstrated in all experiments and is a direct benefit of the DS-based formulation.
Moreover, we exemplify that our formulation yields hyperparameters applicable across multiple skills within the same domain, reducing computational effort.

\subsection{Failure cases and limitations}

Although our method demonstrates robust performance in both simulation and real-robot evaluations, we identify a few possible failure regimes.

False attractors generated by the LfD policy near the reference trajectory can persist in regions where $\pi_{\mathrm{lfd}} \approx 1$, causing the robot to stall.
The emergence of these false attractors hinges on the extrapolation capabilities of the LfD policy,
yet, in our framework, this limitation can be mitigated by choosing a small kernel length.
This results in a widespread activation area of the stabilizing policy, suppressing the false attractors.
However, it introduces a trade-off, as shorter kernel lengths enforce more restrictive behavior, thereby limiting the system's extrapolation range.
Moreover, spurious equilibria can emerge in complex scenarios when competing policies produce matching velocity magnitudes in opposing directions,
similarly resulting in stalled motion.
In addition, the variance gradient can saturate far from the training distribution, making recovery with the proposed stabilizing policy formulation infeasible for such states.
Increasing the kernel length expands the recovery region, but simultaneously broadens the LfD activation area, which may introduce the false attractors described above.
Nonetheless, the aforementioned failure modes are largely resolvable through the proposed hyperparameter optimization, evidenced by the 100\% success rates achieved in our simulated experiments (\Cref{sec:lasa}),
in which the hyperparameters were optimized as described in \Cref{sec:hyperparameter_opt}.

Further, initial starting positions near the goal may cause the robot to shortcut directly to the target without replicating the expert trajectory.
Incorporating task-completion states (e.g.,~phase indicators) into task-dependent parameters, as demonstrated in \Cref{sec:dynamic_grasping}, effectively mitigates the issue.

Additionally, the $\argmax$ selection creates Voronoi boundaries, at which the policy switches abruptly between goal points, creating discontinuities in the vector field.
In practice, however, the underlying Cartesian impedance controller mitigates these discontinuities by inherently imposing velocity limits and filtering rapid command transitions.
A common strategy employed in SEDS-like approaches to address this limitation is to transform all demonstration data into a goal-centric reference frame,
ensuring that all demonstrations share the same attractor point at the origin.
While this could in principle be applied to our method, it assumes that a single goal-centered dynamical system can explain all demonstrations
--- a premise that does not hold in all our experimental settings (e.g.,~\Cref{sec:fish,sec:dynamic_grasping}).

In the real-robot experiments that incorporate continuous task-dependent parameters (\Cref{sec:regrasp,sec:fish}),
failures predominantly arise from OOD $\bm{c}$ values where the LfD policy provides inadequate guidance.
Under these conditions, the stabilizing policy merely constrains the robot to the vicinity of the demonstrations, without contributing to task completion,
often causing the robot to stall or oscillate within a limited range.
Addressing this limitation requires a stabilizing policy that not only steers the robot back toward the training distribution
but also actively manipulates the environment to bring back task-relevant parameters into the known regime --- an extension we leave for future work.

Furthermore, self-collisions and collisions with the environment and target objects remain practical constraints,
as evidenced in \Cref{table:dynamic_grasping_experiments}.
While self-collisions occur only rarely and can be prevented through joint limit checks,
the majority of failure cases stem from unintended collisions with target objects resulting from misalignment between the object and robot during the grasp attempt.
These collisions typically trigger a safety stop, but may also result in objects being displaced or knocked over.
In such cases, the altered relative grasping trajectories can intersect with environmental obstacles, leading to collisions with the environment.
Further, collisions with objects increase with belt velocity due to an increased misalignment between the object and gripper during the grasp attempt.
This behavior stems from the reactive nature of our learned policy, which responds to changes but lacks feedforward or anticipatory planning capabilities.
Pose measurement errors also contribute to these failures, though their individual impact remains unquantified.
We plan to mitigate these issues by integrating an explicit obstacle-avoidance policy into the MoE framework.

In addition, the Euler-vector RBF kernel does not strictly respect SO(3) topology, resulting in wrap-around effects near $\pm \pi$, introducing discontinuities.
In practice, this limitation was mitigated in our experiments by expressing all orientations relative to the first demonstrated configuration,
as empirically shown in real-robot experiment results, where we did not observe this behavior.
To address this limitation, manifold-aware kernels will receive attention in future work.

Moreover, current limitations include the individualized representation of task-parameters, the limited scalability arising from the kernel-based formulation, and the GPR-induced restricted input vector dimensionality.
Accordingly, the proposed method is specifically designed for and most effective in low-dimensional, small-data scenarios --- a common setting in many real-world tasks.
For high-dimensional sensor inputs, by contrast, feature extraction needs to be performed decoupled from the policy, as implemented in \Cref{sec:fish,sec:dynamic_grasping} or in a more generic way to enable zero-shot transfer.
However, the modularity of our formulation not only permits the extension of the framework with application-dependent policies, but also, in principle, the incorporation of alternative LfD policy representations, subject to their ability to provide uncertainty quantification.

\section{Conclusion}\label{sec:conclusion}
We introduced a DS-based Mixture-of-Experts framework that can be modulated by generic low-dimensional task-parameters,
making it well-suited for reactive object manipulation including soft, deformable objects.
We showed the applicability of our solution in both simulation and three manipulation tasks on a real 7-DoF robotic system.
In future work, we will investigate the use of alternative LfD policy representations, such as diffusion or flow-based policies,
that can handle more diverse inputs while simultaneously leveraging the robustness provided by our framework.

\printcredits

\section*{Declaration of generative AI and AI-assisted technologies in the manuscript preparation process}
During the preparation of this work the authors used Qwen3.6 and similar models
in order to improve language. After using this tool,
the authors reviewed and edited the content as needed and take full
responsibility for the content of the published article.

\section*{Funding}
This work has received funding from the European Union's Horizon Europe research and innovation program under grant agreement No.101070600, project SoftEnable.

\appendix

\renewcommand{\thefigure}{\thesection.\arabic{figure}}
\setcounter{figure}{0}

\section{Hyperparameters}\label{app:hyperparams}

\Cref{table:hyperparams} summarizes all experiment-dependent parameters and hyperparameters used throughout the experiments described in
\Cref{sec:lasa,sec:regrasp,sec:fish,sec:dynamic_grasping}, for both the full framework and combinations of individual policies.
All hyperparameters (columns $\sigma^2$ - $K_{{\mathrm{gap},\bm{\phi}}}$) are obtained via CMA-ES optimization as described in \Cref{sec:hyperparameter_opt}.
Vector-valued hyperparameters are split into individual columns, and missing values are indicated by ``--''.

\def\thetable{A.\arabic{table}}
\setcounter{table}{0}
\begin{table*}
	\centering
	\caption{\label{table:hyperparams}Parameters and hyperparameters used in each experiment. Vector-valued hyperparameters are split into individual columns. Values apply to both the full framework and combinations of individual policies.}
	\begin{tabular}{lccccccccccccc}
		\toprule
		Exp.                        & $\mathcal{I}$ & $\mathcal{O}$ & $N$ & $I_{\mathrm{max}}$ & Seed & $\sigma^2$ & $l_{\bm{c}}$ & $l_{\bm{x}}$ & $l_{\bm{\phi}}$ & $K_{\mathrm{sp},\bm{x}}$ & $K_{\mathrm{sp},\bm{\phi}}$ & $K_{\mathrm{gap},\bm{x}}$ & $K_{\mathrm{gap},\bm{\phi}}$ \\
		\midrule
		\Cref{sec:lasa}             & 2             & 2             & 500 & 500                & 0    & 1.471      & --           & 3.800        & --              & 49.955                   & --                          & 84.870                    & --                           \\

		\Cref{sec:regrasp}          & 8             & 7             & 500 & 1000               & 42   & 0.432      & 8.095        & 0.066        & 0.074           & 0.280                    & 0.627                       & 0.281                     & 0.782                        \\

		\Cref{sec:fish}             & 8             & 6             & 500 & 1000               & 42   & 0.432      & 20.095       & 0.066        & 0.074           & 0.280                    & 0.627                       & 0.281                     & 0.782                        \\

		\Cref{sec:dynamic_grasping} & 8             & 7             & 500 & 1000               & 42   & 0.432      & 0.095        & 0.066        & 0.074           & 0.280                    & 0.627                       & 0.281                     & 0.782                        \\
		\bottomrule
	\end{tabular}
\end{table*}

\Cref{table:diffusion_params} lists the hyperparameters used for the Diffusion Policy across all experiments.
These values were the same for both the LASA and real-robot tasks.
The notation for the hyperparameters is adopted from~\cite{Chi2024}.

\begin{table}
	\centering
	\caption{\label{table:diffusion_params}Diffusion Policy hyperparameters used across all experiments.}
	\begin{tabular}{lr}
		\toprule
		Hyperparameter                         & Value         \\
		\midrule
		Observation horizon, $T_o$             & 2             \\
		Prediction horizon, $T_p$              & 16            \\
		Action horizon, $T_a$                  & 8             \\
		Diffusion steps (train), D-Iters Train & 100           \\
		Diffusion steps (eval), D-Iters Eval   & 10            \\
		Number of training epochs              & 100           \\
		Total number of NN parameters          & $\sim$ 65.3 M \\
		\bottomrule
	\end{tabular}
\end{table}

\begin{figure}
	\centering
	\begin{subfigure}[b]{0.49\columnwidth}
		\centering
		\includegraphics[trim={0 0 0 1cm},clip,width=\linewidth]{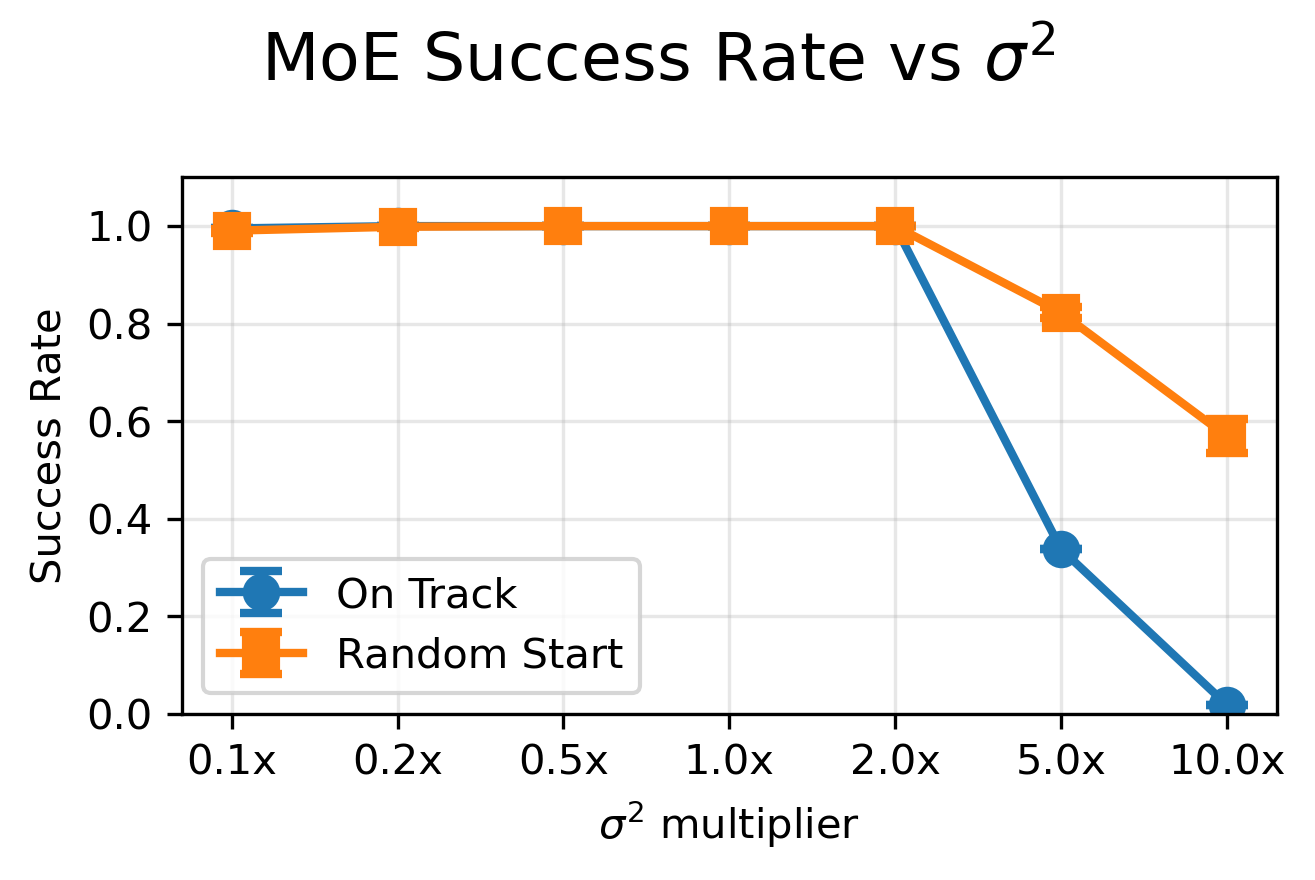}
		\caption{MoE success rate vs $\sigma^2$}
	\end{subfigure}
	\begin{subfigure}[b]{0.49\columnwidth}
		\centering
		\includegraphics[trim={0 0 0 1cm},clip,width=\linewidth]{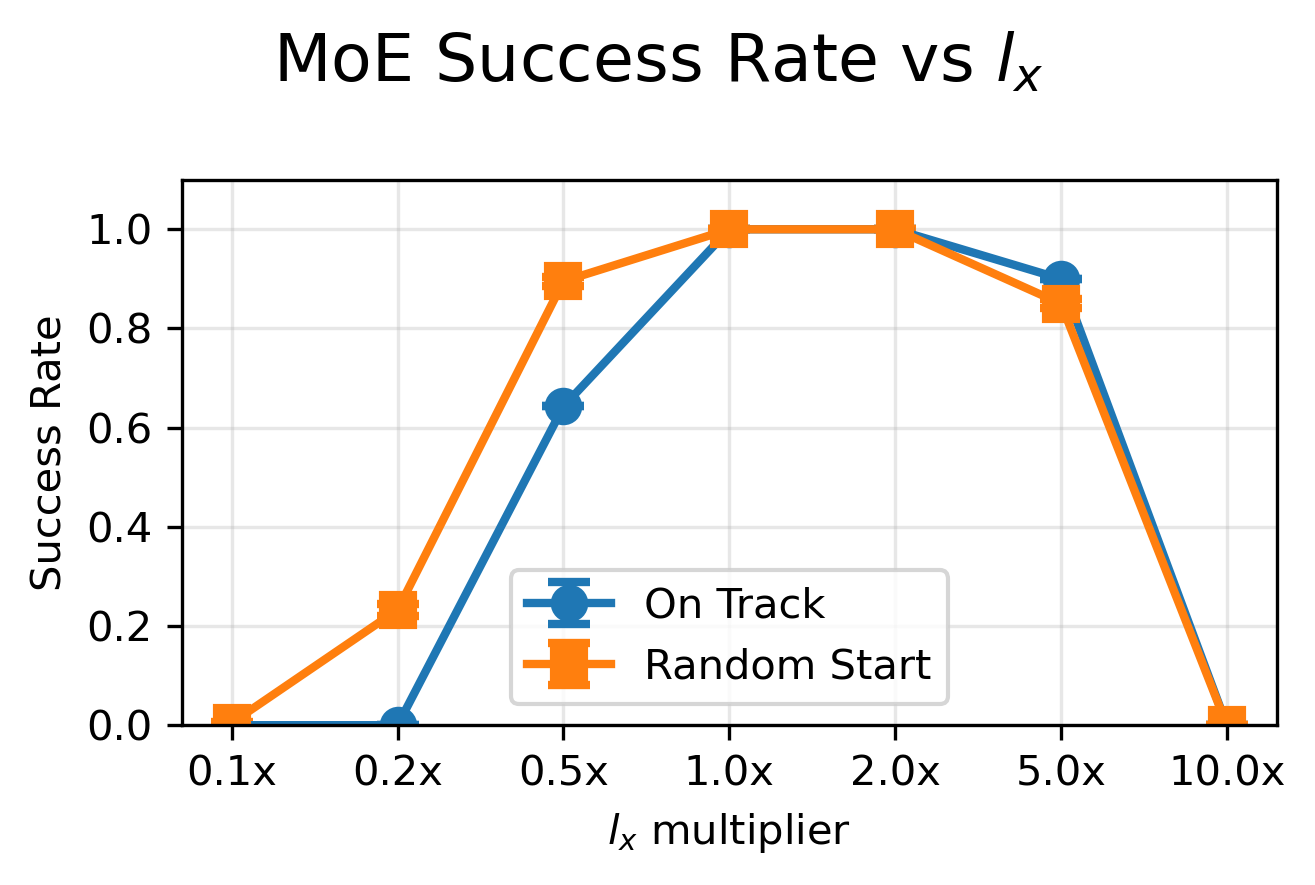}
		\caption{MoE success rate vs $l_{\bm{x}}$}
	\end{subfigure}
	\begin{subfigure}[b]{0.49\columnwidth}
		\centering
		\includegraphics[trim={0 0 0 1cm},clip,width=\linewidth]{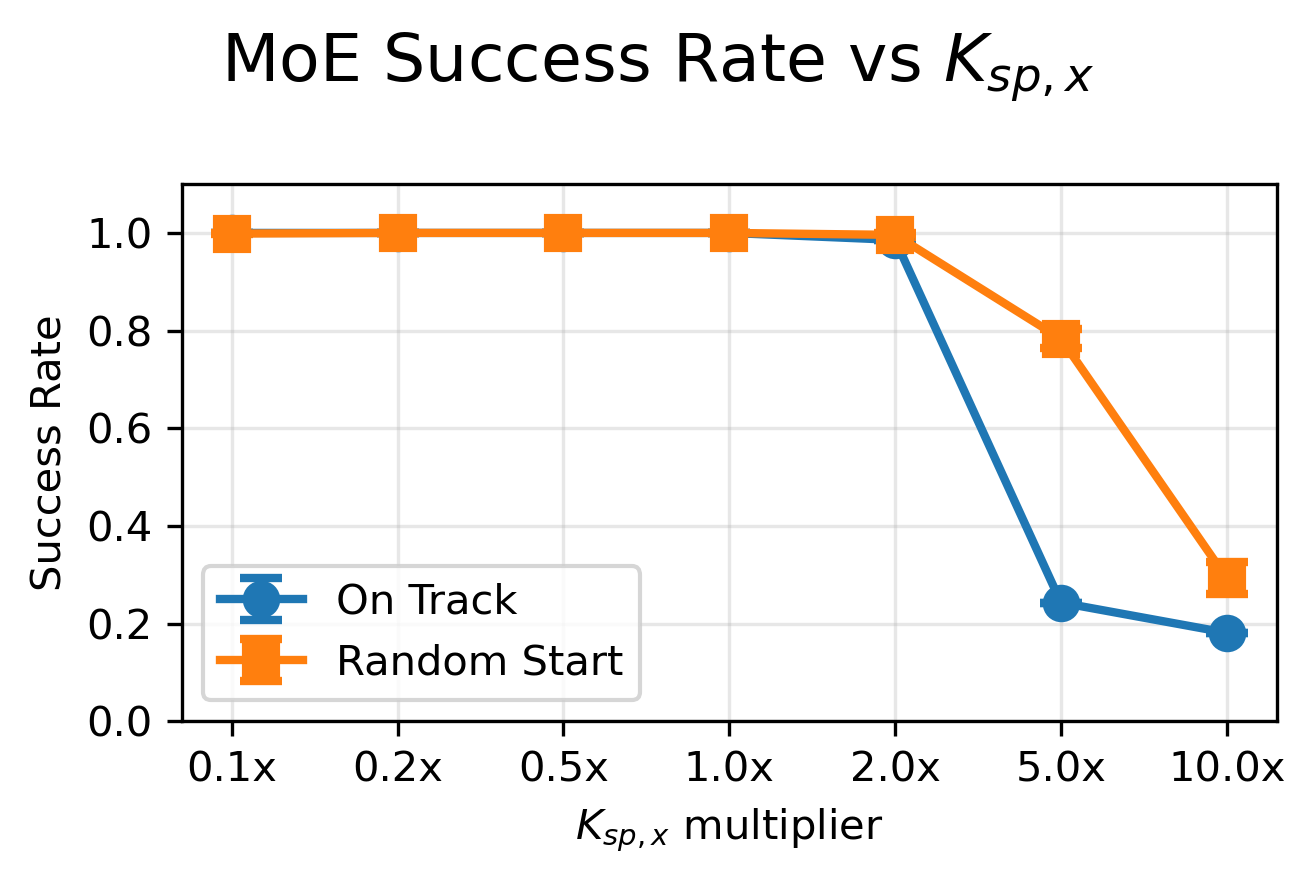}
		\caption{MoE success rate vs $K_{{\mathrm{sp},{\bm{x}}}}$}
	\end{subfigure}
	\begin{subfigure}[b]{0.49\columnwidth}
		\centering
		\includegraphics[trim={0 0 0 1cm},clip,width=\linewidth]{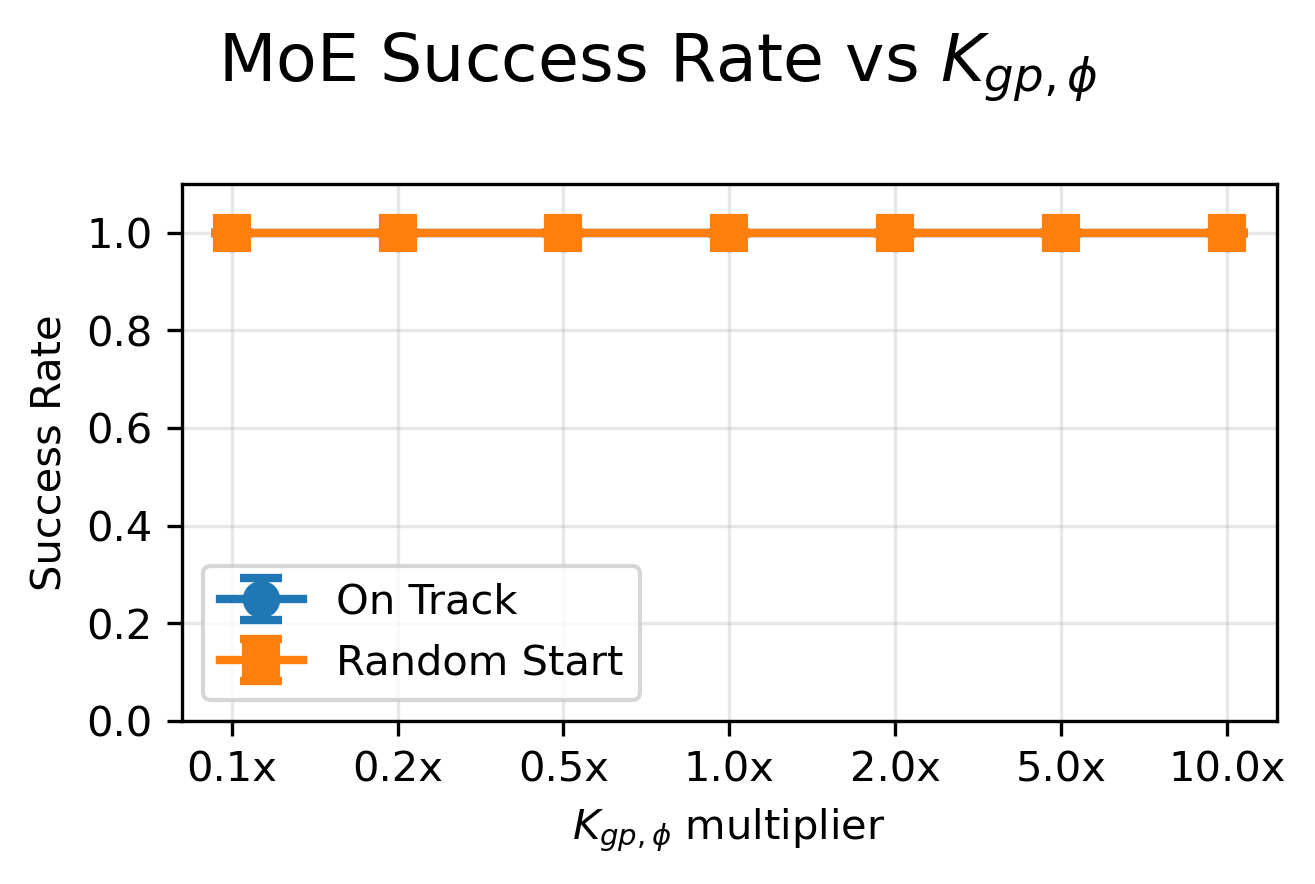}
		\caption{MoE success rate vs $K_{{\mathrm{gap},{\bm{x}}}}$}
	\end{subfigure}
	\caption{\label{fig:sensitivity}Hyperparameter sensitivity analysis:
		Success rate of our MoE approach measured with different multipliers applied to the CMA-ES optimized base value for (a) $\sigma^2$, (b) $l_{\bm{x}}$, (c) $K_{{\mathrm{sp},{\bm{x}}}}$, and (d) $K_{{\mathrm{gap},{\bm{x}}}}$.
		Results are shown for on-demonstration (blue) and random starting poses (orange), with error bars indicating the standard deviation across 50 trials.
		The multiplier 1.0 corresponds to the optimized value.}
\end{figure}

A sensitivity analysis on the LASA handwriting dataset (\Cref{sec:lasa}), which shows how variations in the hyperparameters affect the success rate of the proposed framework,
is displayed in \Cref{fig:sensitivity}, demonstrating robust performance across a reasonable range of hyperparameter values around the optimized solution.

\begin{figure*}
	\centering
	\includegraphics[width=1.0\textwidth]{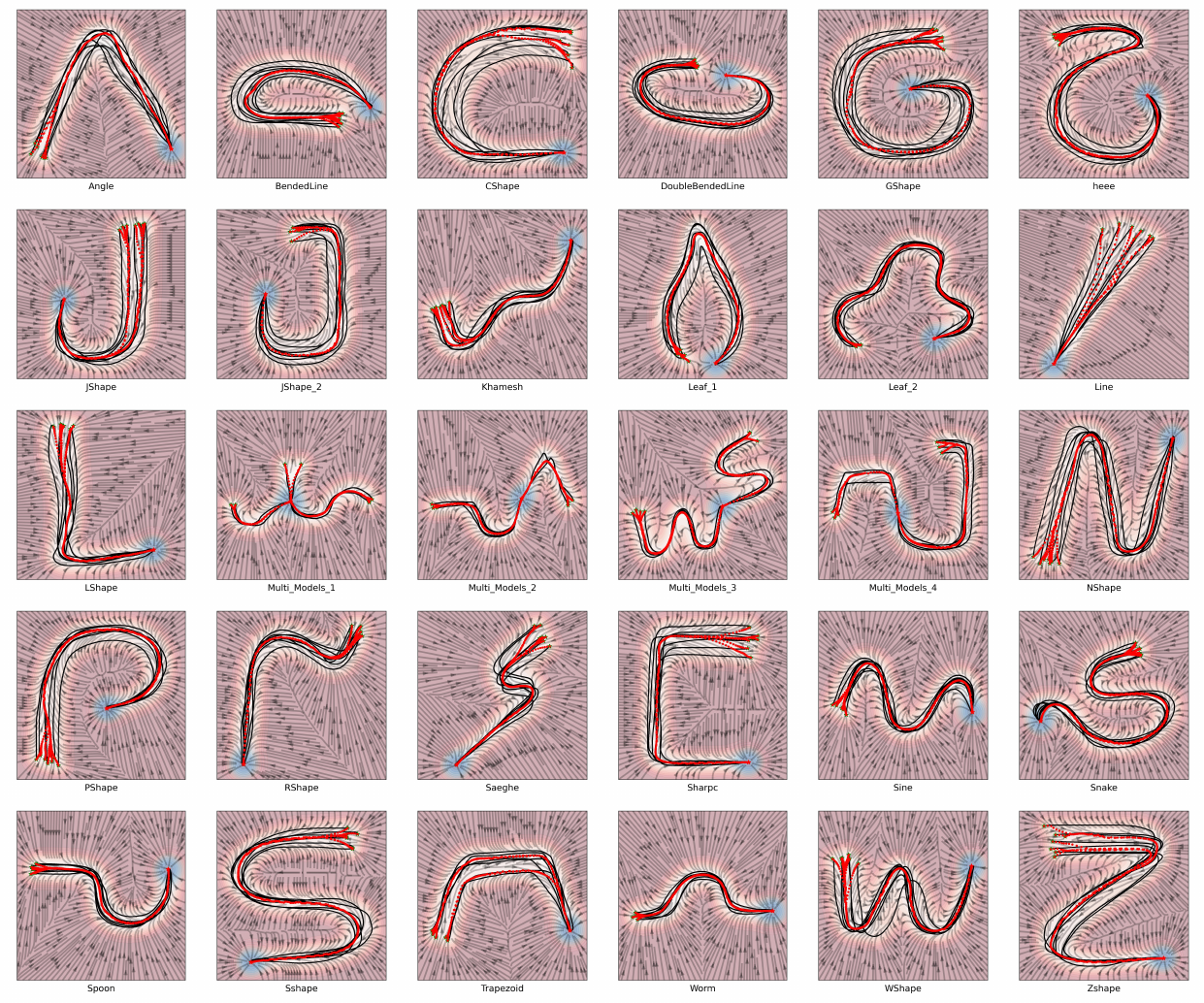}
	\caption{\label{fig:LASA_on_demo}Reproductions for all letters of the LASA handwriting dataset when $\bm{x}_{0,k}$ is placed at the initial pose of each of the seven demonstrations
		showing the executed paths, vector fields and mixing coefficients of the MoE (refer to \Cref{fig:MoE} for clarification).}
\end{figure*}

\section{Runtime performance}

To evaluate the practical viability of the proposed framework for reactive manipulation,
we report inference times, hardware specifications, and a scaling analysis with respect to the number of demonstrations $N$.

We conducted an inference time comparison on the experimental setup described in \Cref{sec:lasa}, where we measure the inference time over 200 trials with multiple inferences per trial.
The results are presented in \Cref{table:inference_time}.

\def\thetable{B.\arabic{table}}
\begin{table}
	\caption{\label{table:inference_time}Inference time comparison on the experimental setup described in \Cref{sec:lasa}}
	\begin{center}
		\begin{tabular}{lc}
			\toprule
			Method                          & Inf time [ms]     \\
			\midrule
			Our approach (MoE)              & $0.402\pm0.002$   \\
			SEDS \cite{KhansariZadeh2011}   & $0.373\pm0.001$   \\
			Diffusion Policy \cite{Chi2024} & $1035.64\pm16.33$ \\
			\bottomrule
		\end{tabular}
	\end{center}
\end{table}

Simulation evaluations were conducted on a workstation equipped with a 13th Gen Intel Core i5-13500 CPU (20 logical cores), 16~GB RAM, and an NVIDIA GeForce GTX 1080 Ti GPU (11~GB VRAM).
Our method is implemented in standard Python without speed optimizations, parallelization, or GPU acceleration.
For direct comparison we also report inference times for SEDS and Diffusion Policy,
with SEDS being trained using the provided MATLAB code, and inference being executed using a standard Python GMR implementation,\footnote{Available at \url{https://github.com/AlexanderFabisch/gmr}.}
where the GMM is initialized with the trained parameters, running entirely on CPU.
Diffusion Policy was evaluated using the state-based environment,\footnote{Adapted from \url{https://github.com/real-stanford/diffusion\_policy} to match our dataset and environment.}
which leverages GPU acceleration.
The table confirms that our framework is computationally lightweight and does not require specialized hardware.

Real-robot experiments were conducted on a separate workstation with an Intel Core i9-14900K CPU (32 logical cores), 64~GB RAM, and an NVIDIA RTX 4500 Ada Generation (24~GB VRAM).
While formal runtime benchmarking was not conducted on the physical platform,
our implementation, despite the unoptimized implementation, comfortably satisfies the real-time control frequencies required for reactive manipulation.

\begin{figure}
	\centering
	\begin{subfigure}[t]{0.49\columnwidth}
		\centering
		\includegraphics[ width=\linewidth]{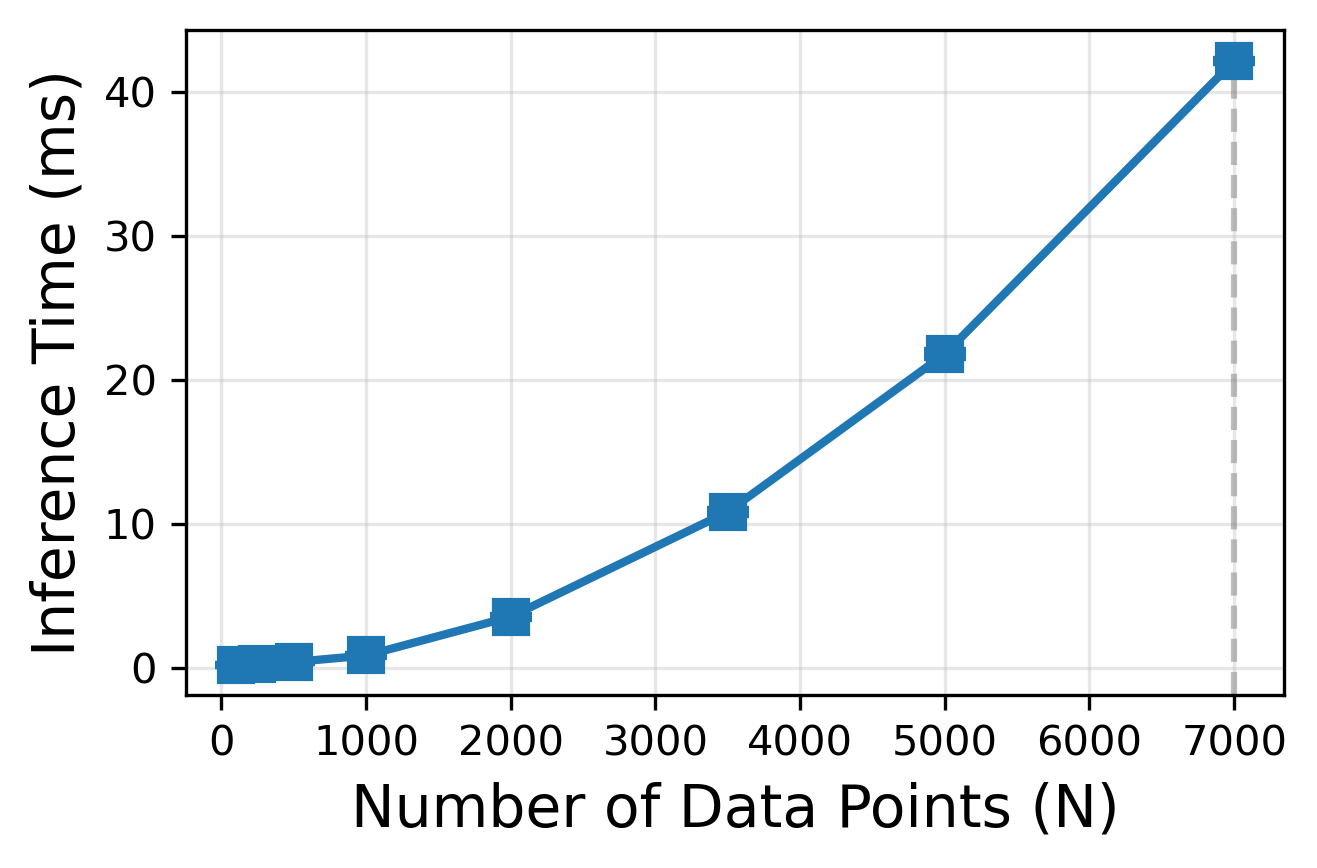}
		\caption{Inference time over $N$}
		\label{fig:inference_scaling}
	\end{subfigure}
	\begin{subfigure}[t]{0.49\columnwidth}
		\centering
		\includegraphics[ width=\linewidth]{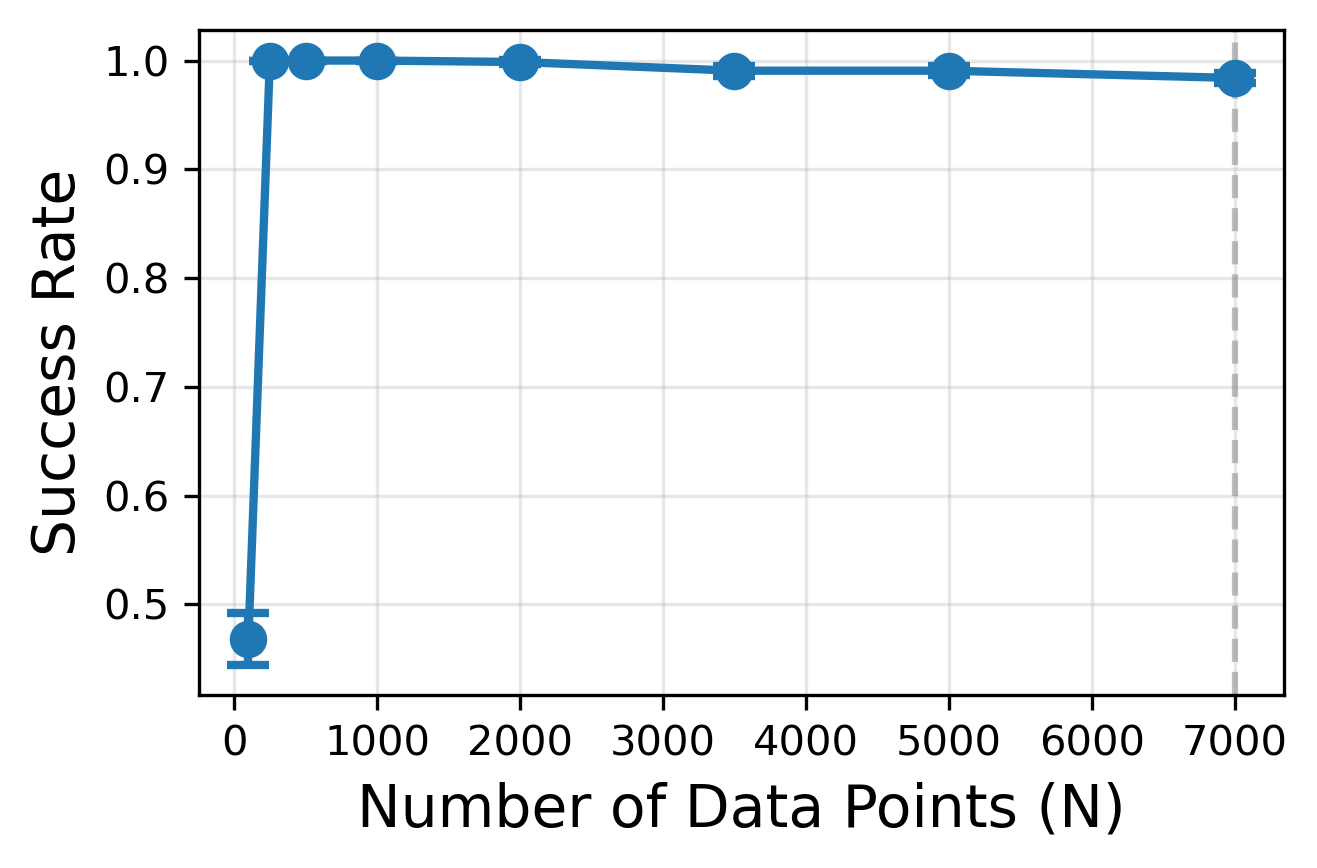}
		\caption{Success rate over $N$}
		\label{fig:success_scaling}
	\end{subfigure}
	\caption{\label{fig:scaling_overhead}Scaling behavior with respect to the number of demonstration samples $N$:
		(a) shows that the inference time per query increases quadratically with $N$, aligning with the theoretical $O(N^2)$ cost of Gaussian process inference.
		(b) demonstrates that the success rate rapidly saturates, forming a plateau for $N \geq 250$.
		Vertical bars denote $\pm 1$ standard deviation across five seeds, each consisting of 10 trials per letter.
		The vertical dashed line marks the full LASA handwriting dataset size ($N=7000$).}
\end{figure}

We further investigate how our approach scales with the number of demonstration samples $N$ using the experimental setup described in \Cref{sec:lasa}.
All experiments employ the hyperparameters optimized for $N=500$ (\Cref{table:hyperparams}).
As shown in \Cref{fig:inference_scaling}, the inference time grows quadratically with $N$, consistent with the $O(N^2)$ complexity of Gaussian process predictive evaluation.
In contrast, the success rate quickly saturates, reaching a plateau for $N \geq 250$ (\Cref{fig:success_scaling}).
This behavior indicates that a modest number of demonstrations is sufficient to adequately cover the underlying task distribution.
Consequently, our chosen configuration $N=500$ resides comfortably on the performance plateau while maintaining computationally efficient inference times,
and the outer control loop frequency is set with a buffer large enough to ensure that it is met even in the event of significant fluctuations of the inference times.

\section{Empirical robustness to $\bm{c}$-perturbations}\label{sec:c-perturbations}

\begin{figure}
	\centering
	\includegraphics[width=0.45\textwidth]{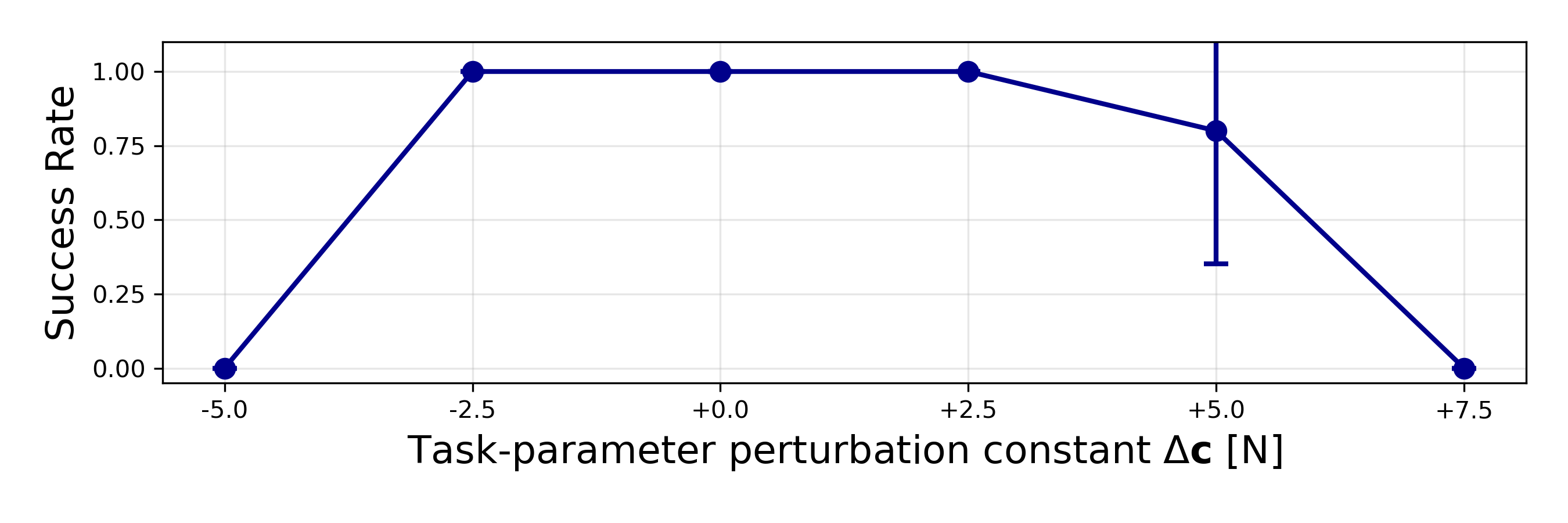}
	\caption{\label{fig:c-perturbations}Empirical robustness to $\bm{c}$ perturbations.
		Constant biases $\Delta \bm{c}$ (in N) are added to $\bm{c}$ during reproduction to simulate systematic force or tracking offsets.
		The $y$-axis reports task success rate (\Cref{sec:regrasp}); error bars represent $\pm 1$ standard deviation over trials.
		With $l_{\bm{c}} \approx 8$ (\Cref{table:hyperparams}), the system maintains full success for deviations up to $\pm 2.5\,\text{N}$,
		confirming practical tolerance to realistic $\bm{c}$ fluctuations.}
\end{figure}

To characterize the practical robustness and tolerance bounds with respect to $\bm{c}$-perturbations,
we conduct an ablation study on the experimental setup described in \Cref{sec:regrasp}.
In this analysis, $\bm{c}$ is artificially perturbed by introducing constant biases $\Delta \bm{c}$ of varying sign
during reproduction to simulate systematic force offsets or drift.
As shown in \Cref{fig:c-perturbations}, the proposed framework achieves full success rates under perturbations up to $\pm 2.5\,\text{N}$,
demonstrating effective tolerance to realistic $\bm{c}$-fluctuations.
We note that the tolerable deviation envelope depends strongly on the task-parameter kernel length $l_{\bm{c}}$ (set to $l_{\bm{c}} \approx 8$, as detailed in \Cref{table:hyperparams}).

\section[cpt,idstr=appS,pfx=Appendix\space]{Supplementary data}

A video showing all experiments, including the LASA benchmarks (\Cref{sec:lasa}) and real-robot tasks (\Cref{sec:regrasp,sec:fish,sec:orientation_ablation_analysis,sec:dynamic_grasping}), is provided as supplementary material.
The video is available at: \href{https://www.youtube.com/watch?v=vxQcqhXqHA8}{Supplementary video}.
It includes both demonstrations and executions, illustrating the performance of the proposed framework.

\section*{Data availability}
Data will be made available on request.

\bibliographystyle{./style/elsarticle-num}

\bibliography{bibliography}


\bio{}
\endbio

\bio{}
\endbio

\bio{}
\endbio

\end{document}